\documentclass{article}
\usepackage{arxiv}

\usepackage{ bbold }
\usepackage{multirow}
\usepackage{amssymb}
\usepackage{amsmath}
\usepackage[all]{genealogytree}
\usepackage{ dsfont }
\usepackage{ mathrsfs }
\usepackage{xcolor}
\usepackage[normalem]{ulem}
\usepackage{collab_tex}
\usepackage{algorithm}
\usepackage{algorithm,algpseudocode}
\usepackage{lineno}
\usepackage{ifthen}

\DeclareMathOperator*{\snr}{SNR}
\DeclareMathOperator*{\CQ}{CQ}
\DeclareMathOperator*{\CI}{CI}
\DeclareMathOperator*{\argmin}{arg\,min}
\DeclareMathOperator*{\argmax}{arg\,max}
\DeclareMathOperator*{\KL}{kl}
\DeclareMathOperator*{\pb}{pb}
\DeclareMathOperator*{\start}{start}
\DeclareMathOperator*{\It}{it}
\DeclareMathOperator*{\IG}{IG}
\DeclareMathOperator*{\diag}{diag}

\DeclareMathOperator*{\test}{test}

\usepackage{xpatch}
\usetikzlibrary{positioning}
\usepackage{neuralnetwork}
\usepackage{ifthen}
\usepackage{algorithm}
\usepackage{algorithm,algpseudocode}
\usetikzlibrary{arrows.meta}
\algnewcommand{\Inputs}[1]{%
	\State \textbf{Inputs:}
	\Statex \hspace*{\algorithmicindent}\parbox[t]{.8\linewidth}{\raggedright #1}
}
\algnewcommand{\Initialize}[1]{%
	\State \textbf{Initialize:}
	\Statex \hspace*{\algorithmicindent}\parbox[t]{.8\linewidth}{\raggedright #1}
}
\makeatletter
\xpatchcmd{\linklayers}{\nn@lastnode}{\lastnode}{}{}
\xpatchcmd{\linklayers}{\nn@thisnode}{\thisnode}{}{}
\makeatother
\renewcommand{\paragraph}[1]{\textit{#1} }
\title{A Review on Quantile Regression for Stochastic Computer Experiments}

\author{
	L\'{e}onard Torossian \\
	MIAT, Universit\'{e} de Toulouse, INRA\\
	and Institut de Math\'{e}matiques de Toulouse\\
	\texttt{leonard.torossian@inra.fr} \\
	\And
	Victor Picheny \\
	PROWLER.io, 72 Hills Road, Cambridge\\
	\texttt{victor@prowler.io} \\
	 \AND
	Robert Faivre \\
	MIAT, Universit\'{e} de Toulouse, INRA \\
    \texttt{robert.faivre@inra.fr} \\
	 \And
	 Aur\'{e}lien Garivier\\
	Univ. Lyon, ENS de Lyon \\
	\texttt{aurelien.garivier@ens-lyon.fr} \\
}

\begin{document}
		\maketitle
		
		\begin{abstract}
			We report on an empirical study of the main strategies for quantile regression in the context of stochastic computer experiments. To ensure adequate diversity, six metamodels are presented, divided into three categories based on order statistics, functional approaches, and those of Bayesian inspiration. The metamodels are tested on several problems characterized by the size of the training set, the input dimension, the signal-to-noise ratio and the value of the probability density function at the targeted quantile. The metamodels studied reveal good contrasts in our set of experiments, enabling several patterns to be extracted. Based on our results, guidelines are proposed to allow users to select the best method for a given problem.
		\end{abstract}

	\section{Introduction}
	Computer simulation models are now essential for performance evaluation, quality control and uncertainty quantification to assess decisions in complex systems.
	These computer simulators generally model systems depending on multiple input variables that can be divided into two categories: the controllable variables and the uncontrollable variables. 
	
For example, in  pharmacology, the optimal drug dosage
	depends on the drug formulation but also on the targeted individual (genetics, age, sex) and environmental interactions. The shelf life and performance of a manufacturing device depend on its design, but also on its environment and on some uncertainties due to the manufacturing process. The plant growth and yield depend on the genes of the plant and on the gardening techniques but also on the weather and potential diseases.
	
	Evaluating the influence of the controllable and uncontrollable variables directly on the real-life problems can be costly and tedious. 
One solution is to encapsulate the systems into a computer simulation model which would reduce the cost and the time required for each test (see \cite{herwig2014computational} and reference therein for computer experiments applied to clinical trials, see \cite{gijo2012product} for a computer simulation model applied to industrial design and \cite{van2003modelling,casadebaig2011sunflo} for computer experiments applied to crop production).

	In such computer simulation models, the links between the inputs and outputs may be too complex to be fully understood or to be formulated in a closed form. In this case, the system can be considered as a black box and formalized by an unknown function: $\Psi:\mathcal{X}\times\Omega\rightarrow \mathds{R}$, where $\mathcal{X}\subset\mathds{R}^d$ denotes the compact space of controllable variables, and $\Omega$ denotes a probability space representing the uncontrollable variables. Note that some black boxes have their outputs in a multidimensional space but this aspect is beyond the scope of the present paper. 
	
	Based on the standard constraints encountered in computer experiments, throughout this paper, we assume that the function is only accessible through pointwise evaluations $\Psi(x,\omega)$; no structural information is available regarding $\Psi$; in addition, we take into account the fact that evaluations may be expensive, which drastically limits the number of possible calls to $\Psi$.
	
If the space $\Omega$ is small enough (or highly structured), it may be possible to work directly on the space $\mathcal{X}\times\Omega$ (see \cite{janusevskis2013simultaneous} for example). 
Often, however, $\Omega$ is too complex and working on the joint space is intractable. 
This is the case for intrinsic stochastic simulators (see \cite{lei2012stochastic,ludkovski2010optimal} for examples of biological systems, where the stochasticity is driven by stochastic equations such as the Fokker-Planck or the
chemical Langevin equations), or for simulators associated with a very large space $\Omega$ (see \cite{casadebaig2011sunflo} for the crop model SUNFLO that considers $5$ weather indicators on $130$ days, $i.e$ $\Omega$ is a space of dimension $650$). 

In this paper we consider the case where $\Omega$ is too complex. We assume that $\Omega$ has any structure and its contribution is considered as random. In contrast to deterministic systems, for any fixed $x$, $\Psi(x,\cdot)$ is considered as a random variable of distribution $\mathds{P}_x$; hence, such systems are often referred to as stochastic black boxes.
In order to understand the behavior of the system of interest or to take optimal decisions, information is needed about the output distribution. An intuitive approach is to use a simple Monte-Carlo technique and to evaluate $\Psi(x,\omega_1), \ldots, \Psi(x,\omega_n)$ to extract statistical moments, the empirical cumulative distribution function, etc. Unfortunately, such a stratified approach is not efficient when evaluating $\Psi$ is expensive.
	
	Instead, we focus on \textit{surrogate models} (also referred to as \textit{metamodels} or \textit{statistical emulators}), which are appropriate approaches in small data settings \cite{stohel08,vilfol12}. 
	Among the vast choice of surrogate models, the most popular ones include regression trees, Gaussian processes, support vector machines and neural networks. In the framework of stochastic black boxes, the standard approach consists in estimating the conditional expectation of $\Psi$. This case has been extensively treated in the literature and many applications, including Bayesian optimization \cite{shahriari2016taking}, have been developed. However the conditional expectation is \textit{risk-neutral}, as it does not provide information about the variability of the output, whereas pharmacologists, manufacturers, asset managers, data scientists and agronomists may need to quantify potential losses associated with their decisions.
	
	Risk information can be introduced via heteroscedastic Gaussian processes \cite{kersting2007most,lazaro2011variational} in which the unknown distribution is estimated by a surrogate expectation-variance model. However, such approaches usually imply that the shape of the distribution (e.g. normal) is the same for all $x\in\mathcal{X}$.
	Another possible approach would be to learn the unknown distribution with no strong structural hypotheses \cite{moutoussamy2015emulators, hall2004cross, efromovich2010dimension}, but this requires a large number of evaluations of $\Psi$. Here, we focus on the conditional quantile estimation of order $\tau$, a flexible way to tackle cases in which the distribution of $\Psi(x,.)$ varies markedly in spread and shape with respect to $x\in\mathcal{X}$, and a classical risk-aware tool in decision theory~\cite{rostek2010quantile}.
	\subsection{Paper Overview}
	Many metamodels originally designed to estimate conditional expectations have been adapted to estimate the conditional quantile. However, despite extensive literature on estimating the quantile in the presence of spatial structure, few studies have reported on the constraints associated with stochastic black boxes. The performance of a metamodel with high dimension input is treated in insufficient details, performance based on the number of points has rarely been tackled and, to our knowledge, dependence on specific aspects of the quantile functions has never been studied. The aim of the present paper is to review quantile regression methods under standard constraints related to the stochastic black box framework, so as to provide information on the performance of the selected methods, and to recommend which metamodel to use depending on the characteristics of the computer simulation model and the data.
	
	A comprehensive review of quantile regression is of course beyond the scope of the present work. We limit our review to the approaches that are best suited for our framework, while ensuring the necessary diversity of metamodels. In particular,  we have chosen six  metamodels that are representative of three main categories: approaches based on statistical order (K-nearest neighbors [KN] regression \cite{bhattacharya1990kernel} and random forest [RF]  regression~\cite{meinshausen2006quantile}), functional or frequentist approaches (neural networks [NN] regression~\cite{cannon2011quantile} and regression in reproducing kernel Hilbert space [RK]~\cite{takeuchi2006nonparametric}), and Bayesian approaches based on Gaussian processes (Quantile Kriging [QK]~\cite{plumlee2014building} and the variational Bayesian [VB] regression \cite{abeywardana2015variational}). Each category has some specificities in terms of theoretical basis, implementation and complexity. We begin this presentation by describing the methods in full in sections~\ref{sec:order}, \ref{sec:functional} and \ref{sec:bayesian}.
	
	In order to identify the relevant areas of expertise of the different metamodels, an original benchmark system is designed based on four toy functions and an agronomical model~\cite{casadebaig2011sunflo}. The dimension of the problems ranges from $1$ to $9$ and the number of observations from $40$ to $2000$. Particular attention is paid to the performance of each metamodel according to the size of the learning set, the value of the probability density function at the targeted quantile $f(.,q_\tau)$ and the dimension of the problem. Sections \ref{sec:implementation} and \ref{sec:benchmarkdesign} describe the benchmark system and detail its implementation, with particular focus on the tuning of the hyperparameters of each method.
	Full results and discussion are to be found in Sections~\ref{sec:results} and \ref{sec:discussion}, respectively.

		\section{Quantile emulators and design of experiments}\label{sec:basics}
\begin{figure}[!ht]
			\begin{center}
				\includegraphics[trim=0 5mm 0 20mm, clip, width=.5\textwidth]{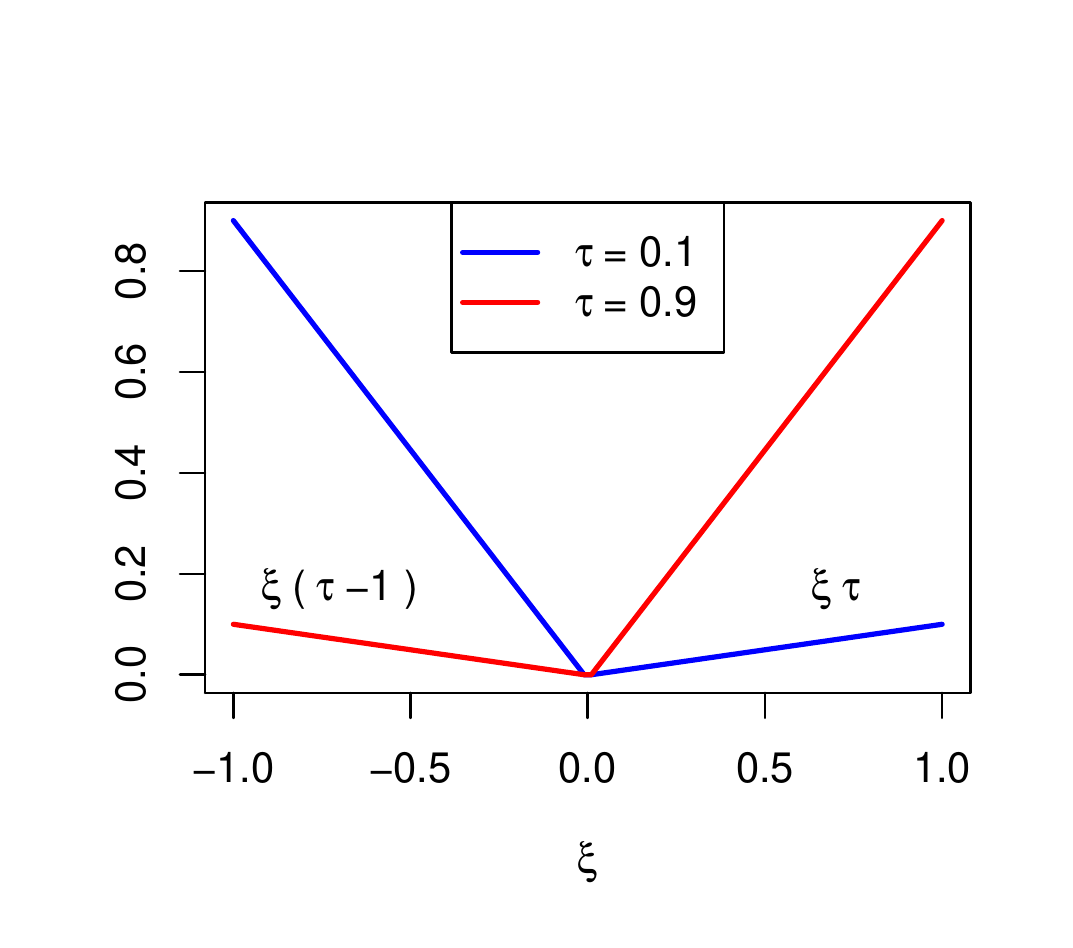}
			\end{center}
			\caption{Pinball loss function with $\tau=0.1$ and $\tau=0.9$.}\label{fig:pinball}
			
		\end{figure}

		We first provide the necessary definitions, objects and properties related to the quantile.
		The quantile of order $\tau\in(0,1)$ of a random variable $Y$ can be defined either as the (generalized) inverse of a cumulative distribution function (CDF), or as the solution to an optimization problem:
		\begin{equation}
		q_\tau= \min\big\{ q \in \mathds{R}~: F(q)\geq \tau  \big\} = \arg\min_{q\in\mathds{R}}\mathds{E}\big[l_{\tau}(Y-q)\big],\label{estimateur}
		\end{equation} 
		 $F(\cdot)$ is the CDF of $Y$ and
		\begin{equation}
		l_{\tau}(\xi)= (\tau-\mathds{1}_{(\xi<0)})\xi, \quad \xi\in\mathds{R}\label{pin}
		\end{equation} is the so-called pinball loss \cite{koenker1978regression} (Figure \ref{fig:pinball}). In the following, we only consider situations in which $F$ is strictly increasing and continuous.
		
		Given a finite observation set $\mathcal{Y}_{n}=(y_1,...,y_n)$ composed of i.i.d samples of $Y$, the empirical estimator of $q_\tau$ can thus be introduced in two different ways:
		\begin{equation}
		\hat{q}_\tau=\min\left\lbrace y_i\in \mathcal{Y}_n: \hat{F}(y_i)\geq \tau \right\rbrace \label{q11} 
			\end{equation}
			or
				\begin{equation}
	\hat{q}_\tau	= \arg\min_{q\in\mathds{R}}\dfrac{1}{n}\sum_{i=1}^{n}l_{\tau}(y_i-q),\label{q22}
		\end{equation}
		where $\hat{F}$ denotes an estimator of the CDF function.
	In (\ref{q11}) $\hat{q}_\tau$ coincides with an order statistic. For example, if $\hat{F}$ is the empirical CDF function then $\hat{q}_\tau=y_{([n\tau])}$, where $[n\tau]$ represents the smallest integer greater than or equal to $n\tau$ and $y_{(k)}=\mathcal{Y}_n(k)$ is the $k$-th smallest value in the sample $\{y_1, \ldots, y_n\}$. The estimators (\ref{q22}) and (\ref{q11}) may coincide, but are in general not equivalent.
	
		Similarly to (\ref{estimateur}), the conditional quantile of order $\tau\in(0,1)$ can be defined in two equivalent ways:
		\begin{equation}
		q_\tau(x)= \min \big\{q : F(q|X = x)\geq \tau  \big\} = \arg\min_{q\in\mathds{R}}\mathds{E}\big[l_{\tau}(Y_x-q)\big],\label{estimateurcondi}
		\end{equation}
		where $Y_x$ is a random variable of distribution $\mathds{P}_{x}$ and $F(.|X = x)$ is the CDF of $Y_x$. 
		
		In a quantile regression context, one only has access to a finite observation set $\mathcal{D}_{n} = \big\{(x_1,y_1), \allowbreak \ldots, \allowbreak (x_n,y_n)\big\}=(\mathcal{X}_n,\mathcal{Y}_n)$ with $\mathcal{X}_n$ a $n\times d$ matrix. Estimators for (\ref{estimateurcondi}) are either based on the order statistic as in (\ref{q11}) (section~\ref{sec:order}), or on a minimizer of the pinball loss as in (\ref{q22})  (sections~\ref{sec:functional} and \ref{sec:bayesian}). Throughout this work, the observation set $\mathcal{D}_{n}$ is fixed (we do not consider a dynamic or sequential framework). Following the standard approach used in computer experiments, the training points $x_i$ are  chosen according to a space-filling design \cite{cavazzuti2013design} over a hyperrectangle. In particular, we assume that there are no repeated experiments: $x_i \neq x_j$, $\forall i \neq j$; most of the methods chosen in this survey (KN, RF, RK, NN, VB) work under that setting.
		
		However, as a baseline approach, one may decide to use a stratified experimental design $\mathcal{D}_{n',r}$ with $r$ i.i.d samples for a given $x_i$, $i=1,..,n'$, extract pointwise quantile estimates using (\ref{q11}) and fit a standard metamodel to these estimates. The immediate drawback is that for the same budget ($n'\times r = n$) such experimental designs cover much less of the design space than a design with no repetition. The QK method is based on this approach.
		\section{Methods based on order statistics}\label{sec:order}
		A simple way to compute a quantile estimate is to take an order statistic of an i.i.d. sample. 
		A possible approach is to emulate such a sample by selecting all the data points in the neighborhood of the query point $x$, and then by taking the order statistic of this subsample as an estimator for the conditional quantile.
		One may simply choose a subsample of $\mathcal{D}_n$ based on a distance defined on $\mathcal{X}$: this is what the $K$-nearest neighbors approach does.
		It is common to use KN based on the Euclidean distance but of course any other distance can be used, such as Mahalanobis \cite{verdier2011adaptive} or weighted Euclidean distance \cite{dudani1976distance}.
		Alternatively, one may define a notion of neighborhood using some space partitioning of $\mathcal{X}$. 
		That includes all the decision tree methods \cite{breiman2017classification}, in particular regression trees, bagging or 
		random forest \cite{meinshausen2006quantile}. 
		
		
		\subsection{$K$-nearest neighbors}
		The $K$-nearest neighbors method was first proposed for the estimation of conditional expectations \cite{stone1975nearest,stone1977consistent}. 
		Its extension to the conditional quantile estimation can be found in \cite{bhattacharya1990kernel}. 
		\subsubsection{Quantile regression implementation}
		Define $\mathcal{X}_{\test}$ as the set of query points.
		KN works as follows: for each $x^*\in\mathcal{X}_{\test}$ define $\mathcal{X}^K(x^*)$ the subset of $\mathcal{X}_n$ containing the $K$ points that are the closest to the query point $x^*$. Define $\mathcal{Y}_K^{x^*}$ the associated outputs, and define $\hat{F}^K(y|X=x^*)$ as the associated empirical CDF. Following (\ref{q11}), the conditional quantile of order $\tau$ can be defined as the statistical order
		\begin{equation}\label{eq:knn}
		\hat{q}_{\tau}(x^*) =  \mathcal{Y}_K^{x^*}([K\tau]).
		\end{equation}
		Algorithm \ref{knear} details the implementation of the KN method.
		\begin{algorithm}
			\caption{K-nearest neighbors}
			\begin{algorithmic}[1]
				\Inputs{$\mathcal{D}_n$, $\tau$, $K$, $\mathcal{X}_{\test}$}
				\For{each point in $x^*\in\mathcal{X}_{\test}$}
				\State\text{Compute all the distances between $x^*$ and $\mathcal{X}_n$}
				\State \text{Sort the computed distances} 
				\State \text{Select the K-nearest points from $x^*$} 
				\State $\hat{q}_{\tau}(x^*)=\mathcal{Y}_K^{x^*}([K\tau])$
				\EndFor
				
			\end{algorithmic}\label{knear}
		\end{algorithm} 
		\subsubsection{Computational complexity}
		For a naive implementation of such an estimator, one needs to compute $n\times N_{new}$ distances, where $N_{new}$ is the number of query points, hence for a cost in $O(nN_{new}d)$.
		Moreover, sorting $n$ distances in order to extract the $K$ nearest points has a cost in $O(nN_{new}\log n)$. Combining the two operations implies a complexity of order $$O(nN_{new}d)+O(nN_{new}\log n).$$ 
		Note that some algorithms have been proposed in order to reduce the computational time, for example by using GPUs \cite{garcia2008fast} or by using tree search algorithms \cite{arya1998optimal}.
		\subsection{Random forests}
		Random forests were introduced by Breiman \cite{breiman2001random} for the estimation of conditional expectations. 
		They have been used successfully for classification and regression, especially with problems where the number of variables is much larger than the number of observations \cite{diaz2006gene}.
		\subsubsection{Overview}
\begin{figure*}
			\begin{center}
				\begin{tabular}{cc}
					\scalebox{0.7}{
						\begin{tikzpicture}
						\genealogytree
						[
						template=formal graph,
						box={code={
								\gtrifroot
								{\tcbset{colback=red!50}}{
									\gtrifleafparent
									{\tcbset{colback=blue!50}}{
										\gtrifleafchild
										{\tcbset{colback=green!50}}{}
									}
								}
							}}
							]{
								child{
									g{\text{ \boldmath{$n_1$} }}
									
									child{g[box={colframe=black!80!white,colback=blue!50}]{\text{ \boldmath{$n_2$} }}c[female,box={colframe=black!80!white,colback=white}]{\text{\boldmath{$A_1$} }}c[box={colframe=black!80!white,colback=white}]{\text{ \boldmath{$A_2$} }}

									}
									child{g[box={colframe=black!80!white,colback=green!50}]{\text{ \boldmath{$n_3$} }}
										child{g[box={colframe=black!80!white,colback=black!25}]{\text{ \boldmath{$n_4$} }}c[female,box={colframe=black!80!white,colback=white}]{\text{\boldmath{$A_3$}  }}c[box={colframe=black!80!white,colback=white}]{\text{\boldmath{$A_4$}}}
											
										}

										child{g[box={colframe=black!80!white,colback=orange!50}]{\text{ \boldmath{$n_5$} }}c[female,box={colframe=black!80!white,colback=white}]{\text{\boldmath{$A_5$}}}c[box={colframe=black!80!white,colback=white}]{\text{ \boldmath{$A_6$} }}
											
										}
									}
									
								}

							}
							\end{tikzpicture}
							
						}
						\includegraphics[width=.33\textwidth]{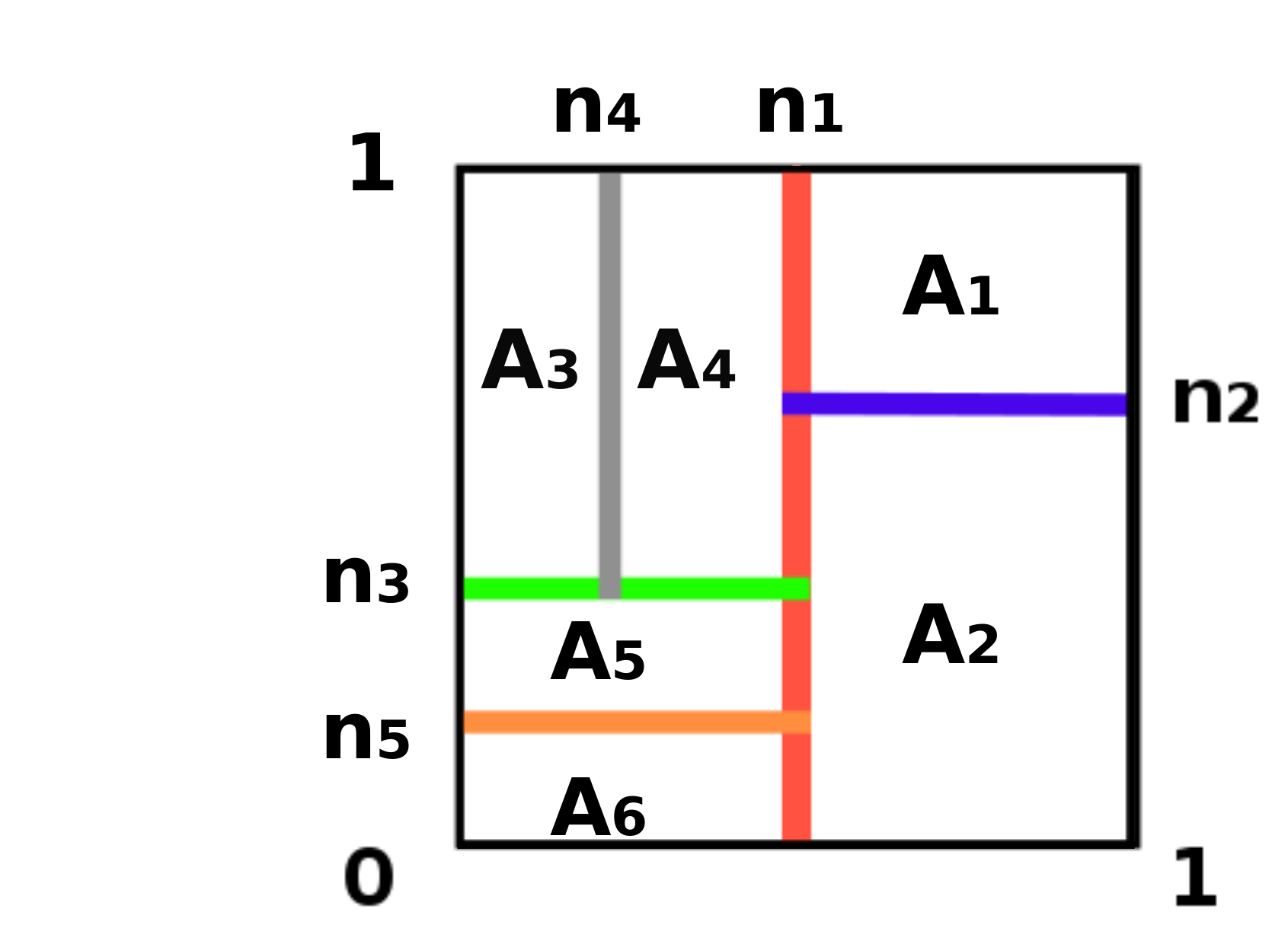}
					\end{tabular}
					\caption{Left: a partitioning tree $T$. The nodes $n_i$ ($1\leq i \leq 5$) represent the splitting points, the $A_i$'s ($1\leq i \leq 6$) represent the leaves. 
						Right: $\mathcal{X}=[0,1]^2$ as partitioned by $T$. The regression tree prediction is constant on each leaf $A_i$.}\label{tree}
				\end{center}
			\end{figure*}

		The basic element of random forests is the \emph{regression tree} $T$, a simple regressor built via a binary recursive partitioning process. Starting with all data in the same partition $i.e$ $\mathcal{X}$, the following sequential process is applied. 
		At each step, the data is split into two, so that $\mathcal{X}$ is partitioned in a way that it can be represented by a tree as it is presented Figure \ref{tree}.
		
			Several splitting criteria can be chosen (see~\cite{ishwaran2015effect}). In~\cite{meinshausen2006quantile}, the splitting point $x_S$ is the data point that minimizes
			\begin{equation}
			C(x_s)=\sum_{x_i\leq x_s}(y_i-\bar{Y}_L)^2+\sum_{x_j>x_s}(y_j-\bar{Y}_R)^2\label{cart},
			\end{equation} 
			where $\bar{Y}_L$ and $ \bar{Y}_R$ are the mean of the left and right sub-populations, respectively. Equation (\ref{cart}) applies when the $x$'s are real-valued. In the multidimensional case, the dimension $d_S$ in which the split is performed has to be selected. The split then goes through $x_S$ and perpendicularly to the direction $d_S$. There are several rules to stop the expansion of $T$. For instance, the process can be stopped when the population of each cell is inferior to a minimal size $nodesize$: then, each node becomes a terminal node or leaf.
			The result of the process is a partition of the input space into hyperrectangles $R(T)$. Like the KN method, the tree-based estimator is constant on each neighborhood. 
			The hope is that the regression trees automatically build neighborhoods from the data that should be adapted to each problem.
			
			Despite their simplicity of construction and interpretation, regression trees are known to suffer from a certain rigidity and a high variance (see \cite{breiman1996bias} for more details). To overcome this drawback, regression trees can be used with ensemble methods like bagging. Instead of using only one tree, bagging creates a set of tree $\mathcal{T}_N =\left\lbrace T^1,..,T^N\right\rbrace $ based on a bootstrap version $\mathcal{D}_{N,n}=\big\{\big((x_{1_t},y_{1_t}),...,(x_{n_t},y_{n_t})\big)\big\}_{t=1}^{N}$ of $\mathcal{D}_n$. 
			Then the final model is created by averaging the results among all the trees. 
			
			Bagging reduces the variance of the predictor, as the splitting criterion has to be optimized over all the input dimensions, but computing (\ref{cart}) for each possible split is costly when the dimension is large. 
			The random forest algorithm, a variant of bagging, constructs an ensemble of weak learners based on $\mathcal{D}_{N,n}$ and aggregates them. Unlike plain bagging, at each node evaluation, the algorithm uses only a subset of  $\tilde{d}$ covariables for the choice of the split dimensions. Because the $\tilde{d}$ covariables are randomly chosen, the result of the process is a random partition $R(t)$ of $\mathcal{X}$ constructed by the random tree $T^t$.
			\subsubsection{Quantile prediction}
			We present the extension proposed in \cite{meinshausen2006quantile} for conditional quantile regression.
			Let us define $\ell(x^*,t)$ the leaf obtained from the tree $t$ containing a query point $x^*$ and
			$$\omega_i(x^*,t)=\dfrac{\mathds{1}_{\{x_i\in \ell(x^*,t)\}}}{\#\{j:x_j \in \ell(x^*,t)\}},~~i=1,...,n  $$
			$$\bar{\omega}_i(x^*)=\dfrac{1}{N}\sum_{t=1}^{N}\omega_i(x^*,t)\;.$$ 
			The $\bar{\omega}_i(x^*)$'s represent the weights illustrating the ``proximity'' between $x^*$ and $x_i$.
			In the classical regression case, the estimator of the expectation is:
			\begin{equation}
			\hat{\mu}(x^*)=\sum_{i=1}^{n}\bar{\omega}_i(x^*)y_i.
			\end{equation}
			In \cite{meinshausen2006quantile} the conditional quantile of order $\tau$ is defined as in (\ref{q11}) with the CDF estimator defined as 
			\begin{equation}
			\hat{F}(y|X=x^*)=\sum_{i=1}^{n}\bar{\omega}_i(x^*)\mathds{1}_{\{y_i\leq y\}}.
			\end{equation}
			Algorithm \ref{forest} details the implementation of the RF method.
			\begin{algorithm}
		\caption{Random forest }
				\begin{algorithmic}[1]
					\State \textbf{Training}
					\Inputs{$\mathcal{D}_n$, $N$, $\tilde{d}$, $m_s$}
					\For{ each of the $N$ trees}
					\State {\small Uniformly sample with replacement $n$ points in $\mathcal{D}_{n}$ to create $\mathcal{D}_{t,n}$. }
					\State {\small Consider the cell $R = \mathcal{X}$.}
					\While{any cell of the tree contains more that $m_s$ observations}
					\For{the cells containing more than $m_s$ observations}
					
					\State {\small Uniformly sample without remplacement $\tilde{d}$ covariables in ${1,...,d}.$}
					\State {\small Compute the cell point among the $\tilde{d}$ covariables that minimizes (\ref{cart}).}
					\State {\small Split the cell at this point perpendicularly to the selected covariable.}
					\EndFor
					\EndWhile
					\EndFor
					\State \textbf{Prediction}
					\Inputs{$\mathcal{X}_{\test}$, $\tau$ }
					\For{each point in $x^*\in\mathcal{X}_{\test}$}
					
					\State Compute $\bar{\omega}_i(x^*),~~i=1\dots,n$

					\State$ \hat{F}(y|X=x^*)=\sum_{i=1}^{n}\bar{\omega}_i(x^*)\mathds{1}_{\{y_i\leq y\}}$
					
					\State $\hat{q}_{\tau}(x^*)=\inf \big\{ y_i : \hat{F}(y_i|X = x^*)\geq \tau  \big\}$
					\EndFor
					
				\end{algorithmic}\label{forest}
			\end{algorithm} 
			\subsubsection{Computational complexity}
			Assuming that the value of (\ref{cart}) can be computed sequentially for consecutive thresholds, the RF computation burden lies in the search of the splitting point that implies sorting the data.
			Sorting $n$ variables has a complexity in $O(n\log n)$. Thus, at each node the algorithm finds the best splitting points considering only $\tilde{d}\leq d$  covariables (classically $\tilde{d}=d/3$). 
			This implies a complexity of $O\big(\tilde{d}n\log n\big)$ per node. In addition, the depth of a tree is generally upper bounded by $\log n$. 
			Then the computational cost of building a forest containing $N$ trees under the criterion (\ref{cart}) is $$O\big(N\tilde{d}n\log^2(n)\big)$$ \cite{louppe2014understanding, witten2016data}. 
			One may observe that RF are easy to parallelize and that contrary to KN the prediction time is very small once the forest is built. 
			
			\section{Approaches based on functional analysis}\label{sec:functional}
			Functional methods search directly for the function mapping the input to the output in a space fixed beforehand by the user.
			With this framework, estimating any functional $S$ of the conditional distribution implies selecting a loss $l$ (associated to $S$) and a function space $\mathcal{H}$.
			Thus, the estimator $\hat{S} \in \mathcal{H}$ is
			obtained as the minimizer of the empirical risk $\mathcal{R}_{e}$ associated to $l$, $i.e.$
			\begin{equation}
			\hat{S} \in \arg\min_{s\in\mathcal{H}}\mathcal{R}_{e}[s]=\arg\min_{s\in\mathcal{H}}\dfrac{1}{n}\sum_{i=1}^{n}l\big(y_i-s(x_i)\big).\label{risk1}
			\end{equation}
			The functional space $\mathcal{H}$ must be chosen flexible enough 
			to extract some signal from the data. In addition, $\mathcal{H}$ needs to have enough structure to make the optimization procedure feasible (at least numerically). 
			In the literature, several formalisms such as linear regression \cite{seber2012linear}, spline regression \cite{marsh2001spline}, support vector machine \cite{vapnik2013nature}, 
			neural networks \cite{bishop1995neural} or deep neural networks \cite{schmidt2017nonparametric} use structured functional spaces with different levels of flexibility.
			
			However, using a too large $\mathcal{H}$  can lead to overfitting, $i.e.$ return predictors that are good only on the training set and generalize poorly.
			Overcoming overfitting requires some \emph{regularization} (see \cite{scholkopf2001kernel, zhao2006model, zou2005regularization} for instance), defining for example the regularized risk 
			\begin{equation}
			\mathcal{R}_{r,e}[s]=\dfrac{1}{n}\sum_{i=1}^{n}l(y_i-s(x_i))+\lambda\left\|s \right\|^\beta,\label{optrisk}
			\end{equation}
			where $\lambda\in\mathds{R}^+$ is a penalization factor, $\beta\in\mathds{R}^+$ and $\left\|. \right\|$ is either a norm 
			for some methods (Section \ref{sec:RK}) or a measure of variability for others (Section \ref{sec:NN}).
			The parameter $\lambda$ plays a major role, as it allows to tune the balance between bias and variance. 
			
			Classically, squared loss is used: it is perfectly suited to the estimation of the conditional expectation. 
			Using the pinball loss (Eq. \ref{pin}) instead allows to estimate quantiles.
			In this section we present two approaches based on Equation~(\ref{optrisk}) with the pinball loss. 
			The first one is regression using artificial neural networks (NN), a rich and versatile class of functions that has shown a high efficiency in several fields. 
			The second approach is the generalized linear regression in reproducing kernel Hilbert spaces (RK). 
			RK is a non-parametric regression method that has been much studied in the last decades (see~\cite{steinwart2008support}) since it appeared in the core of learning theory in the 1990's \cite{scholkopf2001kernel,vapnik2013nature}.
			
			\subsection{Neural Networks}\label{sec:NN}
			Artificial neural networks have been successfully used for a large variety of tasks such as classification, 
			computer vision, music generation, and regression \cite{bishop1995neural}. In the regression setting, 
			feed-forward neural networks have shown outstanding achievements. Here we present quantile regression neural network \cite{cannon2011quantile} which is an adaptation of the traditional feed-forward neural network.
			
			\subsubsection{Overview}
\begin{figure}[!ht]
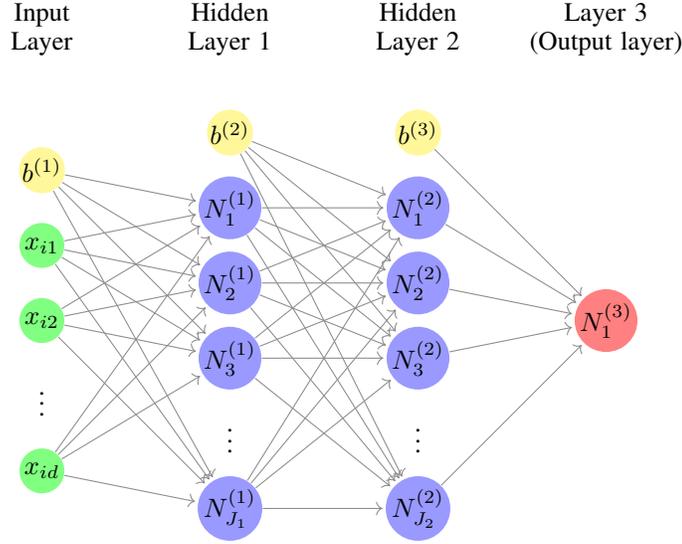

				\begin{center}
				\begin{neuralnetwork}[height=7]
					\newcommand{\nodetextclear}[2]{}
					\newcommand{\nodetextx}[2]{\ifthenelse{\equal{#2}{0}}{$b^{(1)}$}{\ifnum #2=4 $x_{id}$ \else $x_{i#2}$ \fi}}
					\newcommand{\nodetextz}[2]{$\hat{q}_\tau$}
					\newcommand{\nodetexth}[2]{\ifthenelse{\equal{#2}{0}}{$b^{(2)}$}{\ifnum #2=5 $N_{J_1}^{(1)}$  \else $N_{#2}^{(1)}$ \fi}}
					\newcommand{\nodetexthu}[2]{\ifthenelse{\equal{#2}{0}}{$b^{(3)}$}{\ifnum #2=5 $N_{J_2}^{(2)}$  \else $N_{#2}^{(2)}$ \fi}}
					\newcommand{\nodetexthi}[2]{%
						\pgfmathsetmacro\num{int(#2+4)}%
						\ifthenelse{\equal{#2}{0}}{$b_2$}{$N_1^{(3)}$}}
					\inputlayer[count=4, bias=true, exclude={3}, title={Input\\Layer}, text=\nodetextx]
					
					\hiddenlayer[count=5, bias=true, exclude={4},title={Hidden\\Layer 1},text=\nodetexth]
					\link[from layer=0, to layer=1, from node=0, to node=2]
					\link[from layer=0, to layer=1, from node=0, to node=1]
					\link[from layer=0, to layer=1, from node=0, to node=3]
					\link[from layer=0, to layer=1, from node=0, to node=5]
					\link[from layer=0, to layer=1, from node=1, to node=1]
					\link[from layer=0, to layer=1, from node=1, to node=2]
					\link[from layer=0, to layer=1, from node=1, to node=3]
					\link[from layer=0, to layer=1, from node=1, to node=5]
					\link[from layer=0, to layer=1, from node=2, to node=1]
					\link[from layer=0, to layer=1, from node=2, to node=2]
					\link[from layer=0, to layer=1, from node=2, to node=3]
					\link[from layer=0, to layer=1, from node=2, to node=5]
					\link[from layer=0, to layer=1, from node=4, to node=1]
					\link[from layer=0, to layer=1, from node=4, to node=2]
					\link[from layer=0, to layer=1, from node=4, to node=3]
					\link[from layer=0, to layer=1, from node=4, to node=5]
					
					\hiddenlayer[count=5,exclude={4},title={Hidden \\Layer 2}, text=\nodetexthu]
					\link[from layer=1, to layer=2, from node=0, to node=2]
					\link[from layer=1, to layer=2, from node=0, to node=1]
					\link[from layer=1, to layer=2, from node=0, to node=3]
					\link[from layer=1, to layer=2, from node=0, to node=5]
					\link[from layer=1, to layer=2, from node=1, to node=1]
					\link[from layer=1, to layer=2, from node=1, to node=2]
					\link[from layer=1, to layer=2, from node=1, to node=3]
					\link[from layer=1, to layer=2, from node=1, to node=5]
					\link[from layer=1, to layer=2, from node=2, to node=1]
					\link[from layer=1, to layer=2, from node=2, to node=2]
					\link[from layer=1, to layer=2, from node=2, to node=3]
					\link[from layer=1, to layer=2, from node=2, to node=5]
					\link[from layer=1, to layer=2, from node=3, to node=1]
					\link[from layer=1, to layer=2, from node=3, to node=2]
					\link[from layer=1, to layer=2, from node=3, to node=3]
					\link[from layer=1, to layer=2, from node=3, to node=5]
					\link[from layer=1, to layer=2, from node=5, to node=1]
					\link[from layer=1, to layer=2, from node=5, to node=2]
					\link[from layer=1, to layer=2, from node=5, to node=3]
					\link[from layer=1, to layer=2, from node=5, to node=5]
					\outputlayer[count=1, bias=false,title={Layer 3\\(Output layer)}, text=\nodetexthi]
					
					\linklayers[not from={4}]
					\path (L0-2) -- node{$\vdots$} (L0-4);
					\path (L1-3) -- node{$\vdots$} (L1-5);
					\path (L2-3) -- node{$\vdots$} (L2-5);
				\end{neuralnetwork}
					\end{center}
				\caption{Architecture of 3-layer feedforward neural network.}\label{network}
			\end{figure}

			A feed-forward neural network is defined by its number of hidden layers $H$, its numbers of neurons per layer $J_h, 1\leq h\leq H$, and its activation functions $g_{h}$, $h=1,\dots,H$. 
			Given an input vector $x\in\mathds{R}^d$ the information is fed to the hidden layer $1$ composed of a fixed number of neurons $J_1$. 
			For each neuron $N_i^{(1)}$, $i=1,..,J_1$, a scalar product (noted $\left\langle.~,.\right\rangle$) is computed between the input vector $x=(x_1,...,x_d)\in\mathds{R}^d$ and 
			the weights $w_i^{(1)}=(w_{i,1}^{(1)},...,w_{i,d}^{(1)})\in\mathds{R}^d$ of the $N_i^{(1)}$ neurons. Then a bias term $b_i^{(1)}\in\mathds{R}$ is added 
			to the result of the scalar product. The result is composed with the activation function $g_{1}$ 
			(linear or non-linear) which is typically the sigmoid or the ReLu function \cite{schmidt2017nonparametric} and the result is given 
			to the next layer where the same operation is processed until the information comes out from the outpout layer. 
			For example, the output of a $3$-layers NN at $x^*$ is given by
			\begin{equation}
			s(x^*)=g_3\left(  \sum_{j=1}^{J_2}g_{2}\Big( \sum_{i=1}^{J_1}g_{1}\big(\langle w_i^{(1)},x^*\rangle+b_i^{(1)} \big)w_{j,i}^{(2)} +b_j^{(2)}\Big)w_{1,j}^{(3)}+b^{(3)}\right).\label{nnout}
			\end{equation}
			The corresponding architecture can be found in Figure \ref{network}.
			
		
			The architecture of the NN defines $\mathcal{H}$. 
			Finding the right architecture is a very difficult problem which will not be treated in this paper. However a classical implementation procedure consists of creating a network large enough (able to overfit) 
			and then using techniques such as early stopping, dropout, bootstrapping or risk regularization to avoid overfitting \cite{srivastava2014dropout}. 
			In \cite{cannon2011quantile}, the following regularized risk is used:
			\begin{equation}
			\mathcal{R}_{r,e}[s]=\dfrac{1}{n}\sum_{i=1}^{n}l\big(y_i-s(x_i)\big)+\lambda\sum_{j=1}^{H} \sum_{z=1}^{J_j}\left\| w_z^{(j)}\right\| ^2.\label{rqnn}
			\end{equation}
			\subsubsection{Quantile regression}
			
			Minimizing Equation~\eqref{rqnn} (with respect to all the weights and biases) is in general challenging, as $\mathcal{R}_{r,e}$ is a highly multimodal function.
			It is mostly tackled using derivative-based algorithms and multi-starts ($i.e$ launching the optimization procedure $M_s$ times with different starting points).
			In the case of quantile estimation, the loss function is non-differentiable at the origin, which may cause problems to some numerical optimization procedures. 
			To address this issue, \cite{cannon2011quantile} introduced a smooth version of the pinball loss function, defined as:
			$$l_{\tau}^{\eta}(\xi)= h^{\eta}(\xi)(\tau-\mathds{1}_{\xi<0}) ,$$
			where 
			\begin{equation}
			h^{\eta}(\xi)=
			\left\{
			\begin{aligned}
			\dfrac{\xi^2}{2\eta}~~\text{if}~~0\leq |\xi|\leq \eta\\
			|\xi|-\dfrac{\eta}{2}~~\text{if}~~ |\xi|\geq \eta.\\
			\end{aligned}
			\right.
			\end{equation}
			Note that if the optimizer is based on a first order method such as \cite{kingma2014adam}, then the transfer function does not require continuous derivatives. But using a second order method as it is done in the original paper implies the loss function to be twice differentiable with respect to the weights of the neural network. Then transfer functions such as logistic or hyperbolic tangent functions should be used over piecewise linear ones such as the ReLu or the PReLU functions \cite{ramachandran2017searching}.
			
			Let us define $\boldsymbol{w}$ the list containing the weights and bias of the network.
			To find $\boldsymbol{w}^*$, a minimizer of $\mathcal{R}_{r,e}$, the idea is to solve a series of problems using the smoothed loss instead of the pinball one with a sequence $E_K$ corresponding to $K$ decreasing values of $\eta$.
			The process begins with the optimization with the larger value $\eta_1$. Once the optimization converges, the optimal weights are used as the initialization for the optimization with $\eta_2$, and so on. 
			The process stops when the weights based on $l_{\tau}^{\eta_K}$ are obtained. Finally, $\hat{q}_\tau(x^*)$ is given by the evaluation of the optimal network at $x^*$.
			Algorithm \ref{neural} details the implementation of the NN method.
			
			\begin{algorithm}
				\caption{Neural network }
				\begin{algorithmic}[1]
					\State \textbf{Training}
					\Inputs{$\mathcal{D}_n$, $\tau$, $\lambda$, $H$, $(J_1,\dots,J_H)$, $(g_1,\dots,g_H)$, $E_K$}
					\Initialize{Fix $\boldsymbol{w}_0$ as the list containing the initial weights and biases.}
					\For{t = 1 to K}
					\State$\epsilon \gets E_K[t]$
					\State Starting the optimization procedure with $\boldsymbol{w}_0$ and define $$\boldsymbol{w}_{\tau}^*=\argmin_{\boldsymbol{w}} \dfrac{1}{n}\sum_{t=1}^{n}l_{\tau}^{\epsilon}(y_i-\hat{q}^{\boldsymbol{w}}(x_i))+\lambda \sum_{j=1}^{H} \sum_{i=1}^{J}\left\| w_i^{(j)}\right\| ^2$$
					\quad~ with $\hat{q}^{\boldsymbol{w}}(\cdot)$ the output of the network with the weights $\boldsymbol{w}$.
					\State $\boldsymbol{w}_0 \gets \boldsymbol{w}_{\tau}^*$
					\EndFor
					\State \textbf{Prediction}
					\Inputs{$\mathcal{X}_{\test}$, $\boldsymbol{w}_{\tau}^*$, $\lambda$, $H$, $(J_1,\dots,J_H)$, $(g_1,\dots,g_H)$}
					\For{each point in $x^*\in\mathcal{X}_{\test}$}
					\State	Define $\hat{q}_{\tau}(x^*)$ = $\hat{q}^{\boldsymbol{w}_{\tau}^*}(x^*)$.
					\EndFor
				\end{algorithmic}\label{neural}
			\end{algorithm} 
			
			\subsubsection{Computational complexity}\label{compnn}
			In \cite{cannon2011quantile} the optimization is based on a Newton method. Thus, the procedure needs to inverse a Hessian matrix. Without sophistication, 
			its cost is $O(s_{\pb}^3)$ with $s_{\pb}$ the size of the problem $i.e$ the number of parameters to optimize. Note that using a high order method makes sense here because NN has few parameters (in contrast to deep learning methods).  Moreover providing an upper bound on the number of 
			iterations needed to reach an optimal point may be really hard in practice because of the non convexity of (\ref{rqnn}). In the non-convex case, 
			there is no optimality guaranty and the optimization could be stuck in a local minima. However, it can be shown 
			that the convergence near a local optimal point is at least super linear (see \cite{boyd2004convex} Eq.~(9.33)) and may be quadratic (if the gradient is small). It implies, for each $\eta$, the 
			number of iterations until $\mathcal{R}_{r,e}^\eta(\boldsymbol{w})-\mathcal{R}_{r,e}^{\eta}(\boldsymbol{w}^*)\leq\epsilon$ is bounded above by
			$$\dfrac{\mathcal{R}_{r,e}^{\eta}(\boldsymbol{w}_0)-\mathcal{R}_{r,e}^{\eta}(\boldsymbol{w}^*)}{\gamma}+\log_2\log_2(\epsilon_0/\epsilon),$$
			with $\gamma$ the minimal decreasing rate, $\epsilon_0=2M_\eta^3/L_\eta^2$, $M_\eta$ the strong convexity constant of $\mathcal{R}_{r,e}^{\eta}$ near $\boldsymbol{w}^*$
			and $L_\eta$ the Hessian Lipschitz constant (see~\cite{boyd2004convex} page 489). As $\log_2\log_2(\epsilon_0/\epsilon)$ increases very slowly with 
			respect to $\epsilon$, it is possible to bound the number of iterations $N$ typically by 
			$$\dfrac{\mathcal{R}_{r,e}^{\eta}(\boldsymbol{w}_0)-\mathcal{R}_{r,e}^{\eta}(\boldsymbol{w}^*)}{\gamma}+6.$$
			That means, near an optimal point, the complexity is $O(L_{\eta}n(Jd)^3)$, with $J$ the total number of neurons. Then using a multistart procedure implies a complexity of 
			$$O(M_{s}L_{\eta^*}n(Jd)^3),$$ with $L_{\eta^*}=\max_{\eta_1,\dots,\eta_K}L_\eta$  .
			
			\subsection{Regression in RKHS}\label{sec:RK}
			Regression in RKHS was introduced for classification via Support Vector Machine by \cite{cortes1995support,hearst1998support}, 
			and has been naturally extended
			for the estimation of the conditional expectation \cite{drucker1997support,rosipal2001kernel}. Since, many applications have been developed (see \cite{steinwart2008support, scholkopf2001kernel} for some examples), here we present the quantile regression in RKHS \cite{takeuchi2006nonparametric,sangnier2016joint}. 
			
			\subsubsection{RKHS introduction and formalism}
			Under the linear regression framework, $S$ is assumed to be under the form
			$S(x)=x^T\alpha$,
			with $\alpha$ in $\mathds{R}^d$. To stay in the same vein while creating non-linear responses, one can map the input space $\mathcal{X}$ to a space of higher dimension $\mathcal{H}$ (named the feature space), thanks to a feature map $\Phi$. Nevertheless, although large dimensional spaces enable more flexibility, it may introduce difficulties during the estimation procedure.
		A good trade-off can be found		
by combining regularized empirical risk \begin{equation} \label{risk}
		\mathcal{R}_{r,e}[s]=\dfrac{1}{n}\sum_{i=1}^{n}l(y_i-s(x_i))+\dfrac{\lambda}{2}\left\|s \right\|_{\mathcal{H}}^2,
		\end{equation}
		and the RKHS formalism that is based on the \textit{representer theorem} and the so-called \textit{kernel trick}.
			Define $\mathcal{H}$ as a $\mathds{R}$-Hilbert functional space and let us assume there exists a symmetric definite positive function $k : \mathcal{X}\times\mathcal{X} \rightarrow \mathds{R}$ such that:
			\begin{equation}
			k(x,x')=\left\langle \Phi(x'),\Phi(x)\right\rangle _\mathcal{H}.\label{kernel}
			\end{equation}
			Under this setting, $\mathcal{H}$ is a RKHS with the reproducing kernel $k$, that means $\Phi(x)=k(.,x)\in \mathcal{H}$ for all $x\in \mathcal{X}$ and the reproducing property
			\begin{equation}
			s(x)=\left\langle s, k(.,x)\right\rangle _\mathcal{H}\label{reproducing}
			\end{equation}
			holds for all $s\in \mathcal{H}$ and all $x\in \mathcal{X}$.
			It can be shown that working with a fixed kernel $k$ is equivalent to work with its associated functional Hilbert space.
			Note that the kernel choice is based on kernel properties or assumptions made on the functional space. See for instance \cite{steinwart2008support}, chapter 4, for some kernel definitions and properties. 
			In the following, $\mathcal{H}_{\theta}$ and $k_{\theta}$ denote respectively a RKHS and its kernel associated to the hyperparameters vector $\theta$. 
			$K_{x,x}^{\theta}\in \mathcal{R}^{n\times n}$ is the kernel matrix obtained via $K_{x,x}^\theta(i,j)=k_{\theta}(x_i,x_j)$.
			
			From a theoretical point of view, the \textit{representer theorem} implies that the minimizer $\hat{S}$ of (\ref{risk}) lives in 
			$$\mathcal{H}_{|X}^{\theta}=\operatorname{span}\{\Phi(x_i): i=1,...,n\}~\text{with}~\left\|  s\right\|_{\mathcal{H}_{|X} }^2~=~\sum_{i=1}^{n}\sum_{j=1}^{n}\alpha_i \alpha_j k_{\theta}(x_j,x_i).$$ Combining this result to the definition (\ref{kernel}) and the reproducing property (\ref{reproducing}), it is possible to write $\hat{S}$ as:
			$$\hat{S}(x)=\sum_{i=1}^{n}\alpha_i  k_{\theta}(x,x_i).$$
			Hence, the estimation procedure becomes an optimization problem over $n$ coefficients $\boldsymbol{\alpha}=(\alpha_1,\alpha_2,...,\alpha_n)\in\mathds{R}^n$. More precisely, finding
			$\hat{S}$ is equivalent to minimize with respect to $\boldsymbol{\alpha}$ the quantity
			\begin{equation}
			\dfrac{1}{n}\sum_{i=1}^{n}l\Big(y_i-\big(\sum_{j=1}^{n}\alpha_j k_{\theta}(x_i,x_j)\big)\Big)+\dfrac{\lambda}{2}\sum_{i=1}^{n}\sum_{j=1}^{n}\alpha_i \alpha_j k_{\theta}(x_j,x_i).\label{rkhsrisk}
			\end{equation}
			
			\subsubsection{Quantile regression}
			Quantile regression in RKHS was introduced by \cite{takeuchi2006nonparametric}, followed by several authors \cite{li2007quantile,steinwart2011estimating,christmann2008consistency,christmann2008svms,sangnier2016joint}. 
			Quantile regression has two specificities compared to the general case. Firstly the loss $l$ is defined as the pinball. Secondly, to ensure the quantile property, the intercept is not regularized. 
			More precisely, we assume that 
			$$q_{\tau}(x)=g(x)+b~~\text{with}~~g\in\mathcal{H}_\theta~~\text{with}~~b\in\mathds{R},$$
			and we consider the empirical regularized risk 
			\begin{equation}
			\mathcal{R}_{r,e}[q]:=\dfrac{1}{n}\sum_{i=1}^{n}l_{\tau}\big(y_i-q(x_i)\big)+\dfrac{\lambda}{2}\left\|g \right\|_{\mathcal{H}_{|X}}^2.\label{rkhsriskq}
			\end{equation}
			Thus, the \textit{representer theorem} implies that $\hat{q}_{\tau}$ can be written under the form
			$$\hat{q}_{\tau}(x^*)=\sum_{i=1}^{n}\alpha_i  k_{\theta}(x^*,x_i) +b,$$
			for a new query point $x^*$.
			Since (\ref{rkhsriskq}) cannot be minimized analytically, a numerical minimization procedure is used. 
			\cite{cortes1995support} followed by \cite{takeuchi2006nonparametric} introduced non-negative variables $\xi^{(*)}\in\mathds{R}^+$ 
			to transform the original problem into
			$$\mathcal{R}_{r,e}[q]:=\dfrac{1}{n}\sum_{i=1}^{n}\tau \xi_i+(1-\tau)\xi_i^*+\dfrac{\lambda}{2}\sum_{i=1}^{n}\sum_{j=1}^{n}\alpha_i \alpha_j k_{\theta}(x_j,x_i), $$
			subject to $$y_i-\left(\sum_{j=1}^{n}\alpha_j k_{\theta}(x_i,x_j)+b\right)\leq \xi_i$$\text{and} $$\sum_{j=1}^{n}\alpha_j k_{\theta}(x_i,x_j)+b-y_i\leq \xi_i^*,~ \text{where} ~~\xi_i,~\xi_i^*\geq 0.$$
			Using a Lagrangian formulation, it can be shown (see \cite{steinwart2008support} for instance) that minimizing $\mathcal{R}_{r,e}$ is equivalent to the problem:
			\begin{eqnarray}\label{eq:RKquad}
			\min_{\boldsymbol{\alpha}\in\mathds{R}^n}&&
			\dfrac{1}{2}\boldsymbol{\alpha}^T K_{x,x}^\theta\boldsymbol{\alpha}-\boldsymbol{\alpha}^T \boldsymbol{y} \\
			s.t. &&
			\dfrac{1}{\lambda n}(\tau-1)\leq \alpha_i\leq\dfrac{1}{\lambda n}\tau,~\forall ~1\leq i\leq n\nonumber \\
			\text{and}&&
			\sum_{i=1}^{n}\alpha_i=0\nonumber .
			\end{eqnarray}
			It is a quadratic optimization problem under linear constraint, for which many efficient solvers exist.

			The value of $b$ may be obtained from the Karush-Kuhn-Tucker slackness condition or fixed independently of the problem. A simple way to do so is to choose 
			$b$ as the $\tau$-quantile of $\big(y_i-\sum_{j=1}^{n}\alpha_j k_\theta(x_i,x_j)\big)_{1\leq i \leq n}$.
			Algorithm \ref{Fig:rkhsalg} details the implementation of the RK method.

			\begin{algorithm}
				\caption{RKHS regression}
				\begin{algorithmic}[1]
					\State \textbf{Training}
					\Inputs{$\mathcal{D}_n$,$\tau$, $\lambda$, $k_\theta$}
					\Initialize{\strut 
						\text{Compute the $n\times n$ matrix}~~$K_{x,x}^{\theta}$}
					
					\State \textbf{Optimization:} Select $\boldsymbol{\alpha}^*$ as
					\begin{eqnarray}
					\boldsymbol{\alpha}^*=\arg\min_{\boldsymbol{\alpha}\in\mathds{R}^n}&&\dfrac{1}{2}\boldsymbol{\alpha}^T K_{x,x}^{\theta}\boldsymbol{\alpha}-\boldsymbol{\alpha}^T \boldsymbol{y}\nonumber  \\
					s.t&&\dfrac{1}{\lambda n}(\tau-1)\leq \alpha_i\leq\dfrac{1}{\lambda n}\tau,~\forall ~1\leq i\leq n\nonumber\\\text{and}&& \sum_{i=1}^{n}\alpha_i=0\nonumber 
					\end{eqnarray}
					\State Define $b$ as the \text{$\tau$-quantile of $\big(y_i-\sum_{j=1}^{n}\alpha_j k_\theta(x_i,x_j)\big)_{1\leq i \leq n}$}
					\State \textbf{Prediction}
					\Inputs{$\mathcal{X}_{\test}$, $\boldsymbol{\alpha}^*$, $k_\theta$}
					
					\For{each point in $x^*\in\mathcal{X}_{\test}$}
					\State\text{compute}~~$K_{x^*,x}^{\theta}$
					\State $\hat{q}_{\tau}(x_{\test})= K_{x^*,x}^{\theta}\boldsymbol{\alpha}^* +b$
					\EndFor
				\end{algorithmic}\label{Fig:rkhsalg}
			\end{algorithm}
			\subsubsection{Computational complexity}
			Let us notice two things. Firstly, the minimal upper bound complexity for solving (\ref{eq:RKquad}) is $O(n^3)$. Indeed solving (\ref{eq:RKquad}) without the constraints is easier and it needs $O(n^3)$. Secondly the optimization problem (\ref{eq:RKquad}) is convex, thus the optimum is global.
			
			There are two main approaches for solving (\ref{eq:RKquad}), the interior point method \cite{boyd2004convex} and the iterative methods like libSVM \cite{chang2011libsvm}. 
			The interior point method is based on the Newton algorithm, one method is the barrier method (see~\cite{boyd2004convex} page 590). 
			It is shown that the number of iterations $N$ for reaching a solution with precision $\epsilon$ is $O(\sqrt{n}\log(\frac{n}{\epsilon})).$ 
			Moreover each iteration of a Newton type algorithm costs $O(n^3)$ because it needs to inverse a Hessian. 
			Thus, the complexity of an interior point method for finding a global solution with precision $\epsilon$ is $$O\left(n^{7/2}\log(n/\epsilon)\right).$$
			On another hand, iterative methods like libSVM  transform the main problem into a smaller one. At each iteration the algorithm solves a subproblem in $O(n)$. 
			Contrary to the interior point methods, the number of iterations depends explicitly on the matrix $K_{x,x}^{\theta}$. \cite{list2009svm} shows that the number of iterations is $$O\left(n^2\kappa(K_{x,x}^{\theta})\log(1/\epsilon) \right),$$ 
			where $\kappa(K_{x,x}^{\theta})=\lambda_{\max}(K_{x,x}^{\theta})/\lambda_{\min}(K_{x,x}^{\theta})$. Note that $\kappa(K_{x,x}^{\theta})$ depends on the type of the kernel, 
			it evolves in $O(n^{s'})$ with $s'>1$ an increasing value of the regularity of $k_{\theta}$ \cite{chang1999eigenvalues}. For more information about the eigenvalues of $K_{x,x}^{\theta}$ one can consult \cite{braun2006accurate}.
			
			To summarize, it implies that the complexity of the libSVM method has an upper bound higher than the interior point algorithm. 
			However, these algorithms are known to converge pretty fast. In practice, the upper bound is almost never reached, and thus the most important factor is the cost per iteration, rather than the number of iterations needed. 
			This is the reason why libSVM is popular in this setting.
			
			\section{Bayesian approaches }\label{sec:bayesian}
Bayesian formalism has been used for a wide class of problems such as classification and regression \cite{rasmussen2006gaussian}, model averaging \cite{box2011bayesian} and model selection \cite{raftery1995bayesian}.
		The first Bayesian quantile regression model was introduced in  \cite{yu2001bayesian} where the authors worked under linear frameworks combined with improper uniform priors. Nevertheless the linear hypothesis may be too restrictive to treat the stochastic black box setting. 
		As an alternative \cite{taddy2010bayesian} introduced a mixture modeling framework called Dirichlet process to perform nonlinear quantile regression. However the inference is performed with MCMC methods (see \cite{gilks1995markov,gamerman2006markov} for instance), a procedure that is often costly. A possibility to overcome that drawback is the use of Gaussian process (GP). GPs are powerful in a Bayesian context because of their flexibility and their tractability (GPs are only characterized by their mean $m$ and covariance $k_{\theta}$). 
		Using GPs, a possible approach is to use a joint modeling of the mean and variance, assuming that the distribution is Gaussian everywhere, and then to extract the quantiles of interest as it is done in \cite{kersting2007most,lazaro2011variational}. Nevertheless this strategy may introduce a high bias when the true output distribution is not Gaussian.

		In this section we present QK and VB, two approaches that use  GP as a prior for $q_\tau$. Contrary to the joint modelling approach, here the GP prior for $q_\tau$ does not imply any structure on the output distribution, which allows the creation of very flexible quantile models.

			
			\subsection{Kriging}
	Kriging is based on the hypothesis that
		\begin{equation}\label{gpp}
		S(x)\sim\mathcal{GP}\big(m(x),k_\theta(x,x')\big)\;.
		\end{equation} Which means for every finite set $(x_1,\cdots,x_T)$, the output $\big(S(x_1),\cdots,S(x_T)\big)$ is multivariate Gaussian. Here $m$ is the mean of the process and $k_\theta$ is a kernel function also known as the covariance function. Note that in the sequel we take $m=0$ in order to simplify the computations and notations.
		The covariance function conveys many properties of the process, so its choice should depend on the assumptions made on $S$. A first classical assumption is that the GP is stationary, $i.e$ the correlation between two inputs does not depends on the location but only on the distance between the points. 
		Then, different class of stationary kernels produce GPs with different regularities. The class of Mat\'{e}rn kernels is very convenient because it depends on a regularity hyperparameter that enables the end user to adapt its prior to his regularity assumptions (see \cite{rasmussen2006gaussian} for more details). For example, the  Mat\'{e}rn $5/2$ kernel is defined as 
		\begin{equation}\label{mat}
		k_{\theta}(x,x')=\rho^2\big(1+\sqrt{5}\left\| x-x'\right\| _{\theta}+\frac{5}{3}\left\| x-x'\right\| _{\theta}^2\big)\exp(-\sqrt{5}\left\| x-x'\right\| _{\theta}\big),
		\end{equation}
		where $\rho>0$ and $$\left\| x-x'\right\| _{\theta}^2=(x-x')^T\Lambda_\theta(x-x'),$$
		with $\Lambda_\theta$ a diagonal matrix with diagonal terms the inverses of the $d$ squared length scales $\theta_i$, $i=1,\cdots,d$.
		
		In addition to the assumption \ref{gpp}, let us assume $y_i$ is observed with noise such that
		\begin{equation}\label{gaussiannoise}
		y_i=S(x_i)+\varepsilon_i
		\end{equation}
		\text{with}~$\varepsilon_i\sim \mathcal{N}(0,\sigma_i^2)$.
		As a consequence the associated likelihood is Gaussian, $i.e.$
		\begin{equation}
		p\big(\mathcal{Y}_{n}|\mathcal{X}_n)=\mathcal{N}\big(\boldsymbol{0},K^\theta_{x,x}+\diag(\boldsymbol{\sigma}^2)\big),
		\end{equation}
		with $\boldsymbol{\sigma}=(\sigma_1,\cdots,\sigma_n).$
	Because $\big(\mathcal{Y}_{n}, S(x_*)\big)^T$ is a Gaussian vector of zero mean and covariance
		\begin{equation}
		\boldsymbol{K} = 
		\begin{pmatrix}
		K_{x,x}^\theta+\diag(\boldsymbol{\sigma}^2) & K_{x,x_*}^\theta\\
		K_{x_*,x}^\theta &K_{x_*,x_*}^\theta\\
		\end{pmatrix},
		\end{equation}
	the distribution of $S(x_*)$ knowing $\mathcal{Y}_{n}$ is still Gaussian (see \cite{rasmussen2006gaussian} appendix A.2 for more details). Thus, it is possible to provide the distribution a posteriori for Kriging regression as
		
		\begin{equation}
		S(x_*)\sim \mathcal{N}\big(\bar{S}(x_*),\mathds{V}_S(x_*)\big)\label{gp}
		\end{equation} with
		$$\bar{S}(x_*) = K_{x_*,x}^{\theta} (K_{x,x}^{\theta}+\diag(\boldsymbol{\sigma}^2))^{-1}\mathcal{Y}_{n},$$
		$$\mathds{V}_S(x_*)=k_{\theta}(x_*,x_*)-K_{x_*,x}^{\theta}(K_{x,x}^{\theta}+\diag(\boldsymbol{\sigma}^2))^{-1}K_{x,x_*}^\theta.$$
		
		As in Section \ref{sec:RK}, the covariance functions are usually chosen among a set of predefined ones (for example the Mat\'{e}rn $5/2$ see Eq.\ref{mat}), that depend on a set of hyperparameters $\boldsymbol{\theta}\in\mathds{R}^{d+1}$. 
		The best hyperparameter $\boldsymbol{\theta}^*$ can be selected as the maximizer of the marginal likelihood. More precisely it follows (see \cite{rasmussen2006gaussian} for instance) that
		\begin{align}\label{likel}
		p(\mathcal{Y}_{n}|\mathcal{X}_n,\boldsymbol{\theta},\boldsymbol{\sigma})=&-\dfrac{1}{2}\mathcal{Y}_{n}^T(K_{x,x}^{\theta}+B)^{-1}\mathcal{Y}_{n}\\&-\dfrac{1}{2}\log|(K_{x,x}^{\theta}+B)|-\dfrac{n}{2}\log(2\pi),\nonumber
		\end{align}
		where $|K|$ is the determinant of the matrix $K$.
		Maximizing this likelihood with respect to $\boldsymbol{\theta}$ is usually done using derivative-based algorithms,
		although the problem is non-convex and known to have several local maxima.
		
		Different estimators of $S$ may be extracted based on (\ref{gp}). 
		Here $\hat{S}$ is fixed as $\bar{S}_{\theta^*}$. 
		Note that this classical choice is made because the maximum a posteriori of a Gaussian distribution coincides with its mean. 
			
			\subsubsection{Quantile kriging}\label{QK}
			\begin{algorithm}
				\caption{Quantile kriging }
				\begin{algorithmic}[1]
					\State \textbf{Training}
					\Inputs{$\mathcal{D}_{n',r}$,$\tau$, $k_\theta$}
					\Initialize {\strut \text{Compute the $n'\times n'$ matrix}~~$K_{x,x}^{\theta}$\\Define the local estimators of the $\tau$-quantile
						
						{\For{i = 1 to n'}
							\State $\hat{q}_{\tau}(x_i)\gets y_{i,([r\tau])}$
							\State\text{Estimate} $\sigma_i$ by bootstrap
							\EndFor}
					}
					\State Define $B=\text{Diag}(\sigma_1^2,\dots,\sigma_{n'}^2)$ and compute ~~$(K_{x,x}^{\theta}+B)^{-1}$
					\State Define the kernel hyperparameters $\theta^*$ as $\argmax_{\theta}$ of $$ -\dfrac{1}{2}\mathcal{Q}_{n'}^T(K_{x,x}^{\theta}+B)^{-1}\mathcal{Q}_{n'}-\dfrac{1}{2}\log|K_{x,x}^{\theta}+B|-\dfrac{n}{2}\log(2\pi)$$
					
					\Inputs{$\mathcal{X}_{\test}$, $\mathcal{Q}_{n'}$, $\theta^*$, $B$ }
					\For{each point in $x^*\in\mathcal{X}_{\test}$}
					\State{\text{compute}~~$K_{x^*,x}^{\theta^*}$}
					
					\State	$\hat{q}_{\tau}^{\theta^*}(x^*)=K_{x^*,x}^{\theta^*} (K_{x,x}^{\theta^*}+B)^{-1}\hat{q}_{\tau}$
					\EndFor
				\end{algorithmic}\label{krig}
			\end{algorithm} 
		As $q_\tau$ is a latent quantity, the solution proposed in \cite{plumlee2014building} is to consider the sample $\mathcal{D}_{n',r}$ corresponding to a design of experiments with $n'$ different points that are repeated $r$ times in order to obtain quantile observations. For each $x_i\in \mathcal{X}$, $1\leq i\leq n'$, let us define:
		$$y_{i,r}=(y_{i,1},.., y_{i,r})$$
		and
		$$\mathcal{D}_{n',\tau,r}=\Big(\big(x_1,\hat{q}_{\tau}(x_1)\big),...,\big(x_n,\hat{q}_{\tau}(x_n)\big)\Big), \text{with}~~ \hat{q}_{\tau }(x_i)=y_{i,([r\tau])}.$$
		Following \cite{plumlee2014building}, let us assume that
		\begin{equation}
		\hat{q}_{\tau }(x_i)=q_{\tau}(x_i)+\varepsilon_i, \text{ with }\varepsilon_i\sim \mathcal{N}(0,\sigma_i^2).\label{gaus}
		\end{equation}
		Note that from a statistical point of view Assumption (\ref{gaus}) is wrong because the distribution is asymetric around the quantile but asymptotically consistent as illustrated by the central limit theorem for sample quantiles.
		The resulting estimator is
		\begin{equation}\label{qk}
		\hat{q}_{\tau}(x_*)=K_{x_*,x}^{\theta} (K_{x,x}^{\theta}+B)^{-1}\mathcal{Q}_{n'},
		\end{equation}
		with $\mathcal{Q}_{n'}=\big(\hat{q}_{\tau}(x_1),\dotsb,\hat{q}_{\tau}(x_{n'})\big)$ and $B=\text{diag}(\sigma_1^2, \ldots, \sigma_n'^2)$. 
		
		There are several possibilities to evaluate the noise variances $\sigma_i^2$. Here we choose to use a bootstrap technique 
		(that is, generate bootstrapped samples of $y_{i,r}$, compute the corresponding $\hat{q}_{\tau}(x_i)$ values and take the variance over those values as the noise variance) but it is possible to use the central limit theorem as it is presented in \cite{bachoc2013parametric}.
		The hyperparameters are selected based on $(\ref{likel})$ changing $\mathcal{Y}_n$ by $\mathcal{Q}_{n'}$.
		Algorithm \ref{krig} details the implementation of the QK method.
		
			\subsubsection{Computational complexity}
			
			Optimizing (\ref{likel}) with a Newton type algorithm implies to inverse a $(d+1)\times (d+1)$ matrix. In addition, for each $\theta$, obtaining the partial derivatives of (\ref{likel}) requires the computation of $(K_{x,x}^{\theta}+B)^{-1}$ \cite{rasmussen2006gaussian}. Thus, at each step of the algorithm the complexity is $O(n'^3+d^3)$.
			Assuming the starting point $\theta_{\start}$ is close to an optimal $\theta^*$, based on the same analysis as in section \ref{compnn}, the complexity to find $\theta^*$ is $$O\big(L(d^3+n'^3)\big),$$
			with $L$ the Hessian Lipschitz constant.
			
			Finally, obtaining $\hat{q}_\tau^{\theta^*}$ from (\ref{qk}) implies inversing the matrix $K_{x,x}^{\theta}+B$ that is in $O(n'^3)$. So the whole complexity is
			$$O\big(L(d^3+n'^3)+n'^3\big).$$
			
			\subsection{Bayesian variational regression}
		Quantile kriging requires repeated observations to obtain direct observations of the quantile and make the hypothesis of Gaussian errors acceptable. Variational approaches allow us to remove this critical constraint, while setting a more realistic statistical hypothesis on $\varepsilon$. 
		Starting from the decomposition of Eq.\ref{gaussiannoise}, $\varepsilon(x)$ is now assumed to follow a Laplace asymmetric distribution \cite{yu2005three,lum2012spatial}, implying:
		\begin{equation}
		p\big(y|q,\tau,\sigma,x\big)=\dfrac{\tau(1-\tau)}{\sigma}\exp\left(-\dfrac{l_{\tau}(y-q(x))}{\sigma}\right),
		\end{equation}
		with the priors on $q$ and $\sigma$ that has to be fixed. Note that
		such assumption may be justified by the fact that at $\sigma$ fixed, 
		minimizing the empirical risk associated to the pinball loss is equivalent to maximizing the asymmetric Laplace likelihood, which is given by 
		\begin{equation}
		p\big(\mathcal{Y}_n|q_{\tau},\mathcal{X}_n,\theta\big)=\prod_{i=1}^{n}\dfrac{\tau(1-\tau)}{\sigma}\exp\left(-\dfrac{l_{\tau}(y_i-q_{\tau}(x_i))}{\sigma}\right).\label{lap}
		\end{equation}
		According to the Bayes formula, the posterior on $q_\tau$ can be written as
		$$p\big(q_{\tau}|\mathcal{D}_{n}\big)=\dfrac{p\big(\mathcal{Y}_n|\mathcal{X}_n,q_{\tau}\big)\;p(q_{\tau})}{p\big(\mathcal{Y}_{n}|\mathcal{X}_n\big)}.$$
		As the normalizing constant is independent of $q_{\tau}$, considering only the likelihood and the prior is enough. We obtain 
		\begin{equation}\label{prop}
		p(q_{\tau}|\mathcal{D}_{n})\propto p\big(\mathcal{Y}_{n}|\mathcal{X}_n,q_{\tau}\big) \; p(q_{\tau}).
		\end{equation}
		Because of the Laplace asymetric likelihood, contrary to the classical kriging model, here the posterior distribution (\ref{prop}) is not Gaussian anymore. Thus, it is not possible to provide an analytical expression for the regression model.
		To overcome this problem,
		\cite{boukouvalas2012gaussian} used a variational approach with an expectation-propagation (EP) algorithm \cite{minka2001expectation}, while \cite{abeywardana2015variational} 
		used a variational expectation maximization (EM) algorithm which was found to perform slightly better.
			\subsubsection{Variational EM algorithm}
		The EM algorithm was introduced in \cite{dempster1977maximum} to compute maximum-likelihood estimates. Since then, it has been widely used in a large variety of fields (see \cite{mclachlan2007algorithm} for more details). 
		Classically, the purpose of the EM algorithm is to find $\zeta$ a vector of parameters that define the model and that maximizes $p(\mathcal{Y}_n|\zeta)$ thanks to the introduction of the hidden variables $z=(z_1,\cdots,z_M)$. However dealing with this classical formalism implies that $p(z|\mathcal{Y}_n,\zeta)$ is known or some sufficient statistics can be computed (see \cite{tzikas2008variational, robert2013monte} for more details), which is not always possible. Using the variational EM framework is a possibility to bypass this requirement \cite{tzikas2008variational}, by approximating $p(z|\mathcal{Y}_n,\zeta)$  by a probability distribution $\tilde{p}$ that factorizes under the form 
		$$\tilde{p}(z)=\prod_{j=1}^{M}\tilde{p}_j(z_j).$$
		Starting from the log-likelihood $\log p(\mathcal{Y}_n|\zeta)$, thanks to Jensen's inequality, it is possible to show that:
		$$\log p(\mathcal{Y}_n|\zeta) \geq \mathcal{L}(\tilde{p},\zeta)+\KL(\tilde{p}||p),$$
		where 	$$\mathcal{L}(\tilde{p},\zeta)=\int \tilde{p}(z) \log\left( \dfrac{p\big(\mathcal{Y}_n,z|\zeta\big)}{\tilde{p}(z)}\right) dz,$$
		and $\KL$ is the Kullback-Leibler divergence:
		$$\KL(\tilde{p}||p)=-\int\tilde{p}(z)\log(\dfrac{p\big(z|\mathcal{Y}_n,\zeta\big)}{\tilde{p}(z)})dz.$$ 
		As presented on Figure \ref{EM}, the EM algorithm can be viewed as a two-step optimization technique. The lower bound $\mathcal{L}$ is first maximized with respect to $\tilde{p}$ (E-step) so that to minimize $\KL(\tilde{p}||p)$. Next the likelihood is directly maximized with respect to the parameter $\zeta$ (M-step). 
		
		Classically and in the following, the E-step optimization problem is analytically solved (see \cite{tzikas2008variational} for details about the computation). In this particular case, the quantities $\tilde{p}_j$ are limited to conditionally conjugate exponential families. But note that different strategies have been developed to relax this assumption (generally it comes with an higher computational complexity) so that creating more flexible models (see the review \cite{zhang2018advances} for instance).
			\begin{figure*}[!ht]
				\begin{tabular}{ccc}
					\includegraphics[width=.33\textwidth]{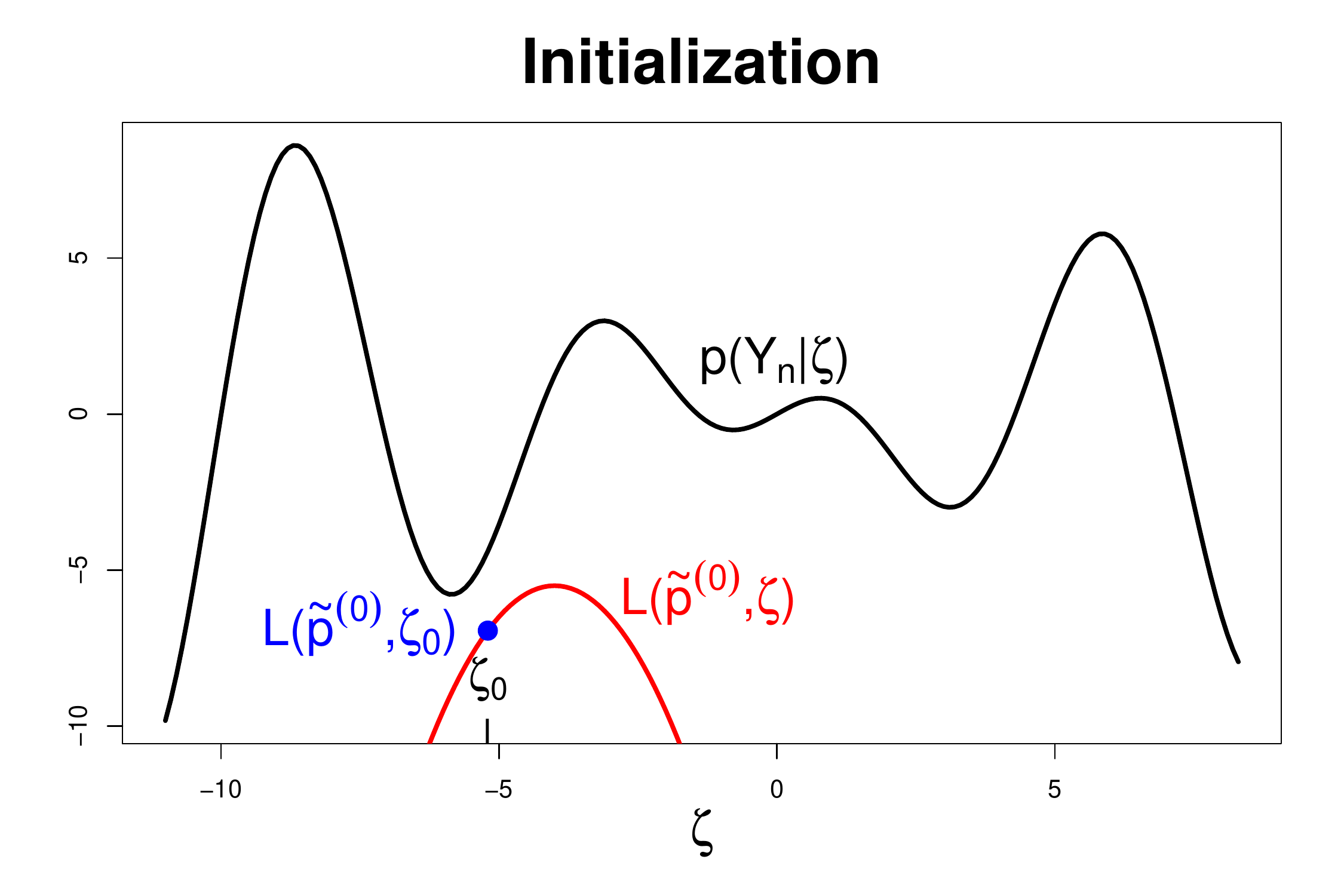}
					\includegraphics[width=.33\textwidth]{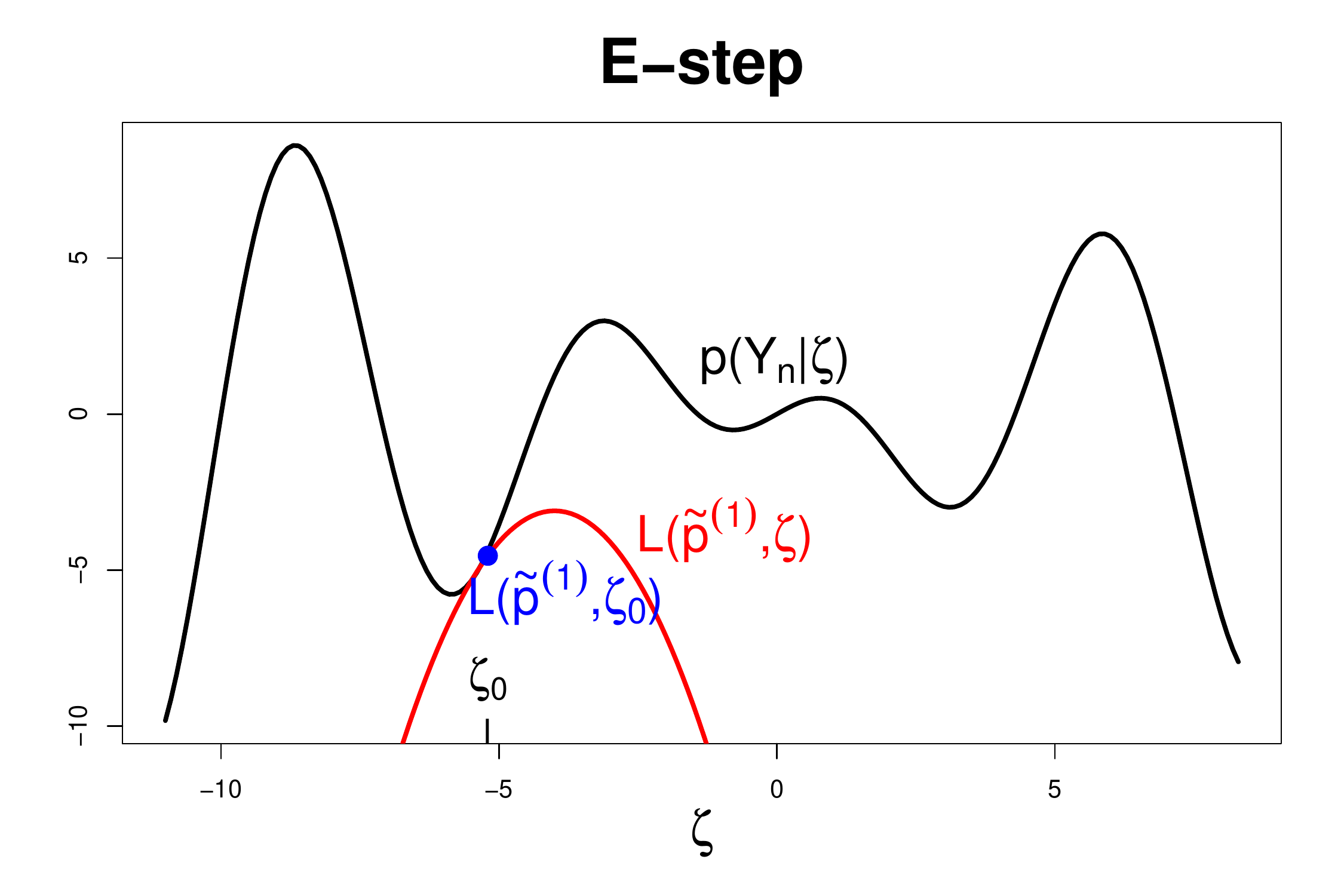}
					\includegraphics[width=.33\textwidth]{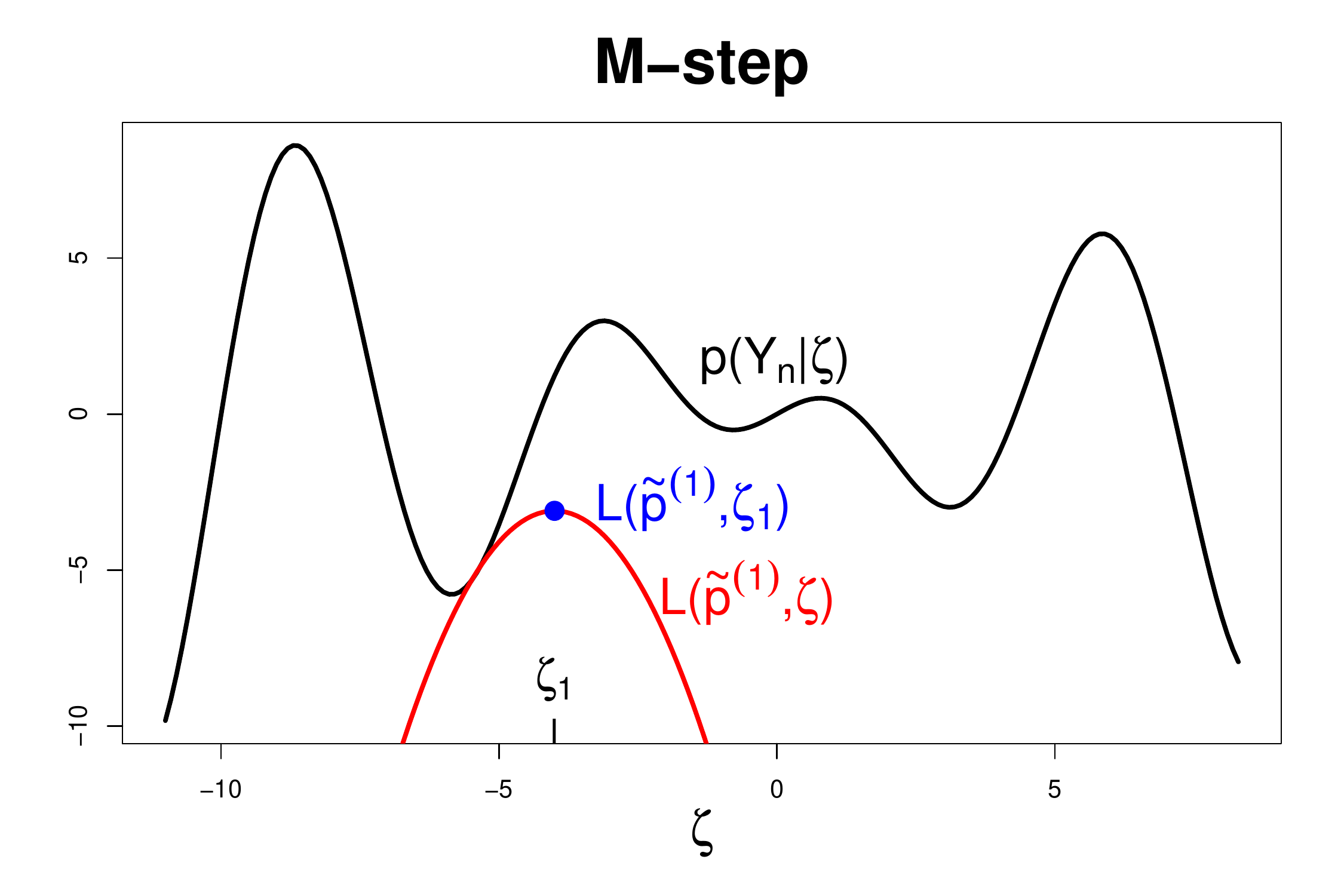}
				\end{tabular}
				\caption{First steps of the variational EM algorithm.}
			\end{figure*}\label{EM}
		
			
			\subsubsection{Variational EM applied to quantile regression}
		Following \cite{abeywardana2015variational}, let us suppose that 
		$$q_{\tau}(x)\sim \mathcal{GP}\big(m(x),k_\theta(x,x')\big)$$
		$$\sigma\sim \IG(10^{-6},10^{-6}),$$ 
		with $\IG$ defining the inverse gamma distribution and for sake of simplicity, $m=0$. Note that contrary to the formalism introduce with QK, here $\sigma$ is taken as a random variable.
		To allow analytical computation, let us introduce an alternative definition of the Laplace distribution \cite{lum2012spatial,kotz2012laplace}:
		
		\begin{equation}
		p(y_i|q_{\tau},x_i,\sigma,\tau)=\int\mathcal{N}(y_i|\mu_i,\sigma_{y_i})\exp(-w_i)dw,\label{modif}
		\end{equation}
		where $\mu_{y_i}=q_{\tau}(x_i)+\frac{1-2\tau}{\tau(1-\tau)}\sigma w_i$, $\sigma_{y_i}=\frac{2}{\tau(1-\tau)}\sigma^2 w_i$ and $w_i$ is distributed according to an exponential distribution of parameter $1$. 
		
		The distribution of $q_{\tau}$ at a new point $x_*$ is given by averaging the output of all Gaussian models with respect to the posterior $p(q_{\tau},\sigma,w|x,Y)$:
		\begin{equation}
		p\big(q_{\tau}(x_*)|,\mathcal{D}_n\big)=\int p\big(q_{\tau}(x_*)|q_{\tau},\sigma,w,\mathcal{D}_n\big)\; p\big(q_{\tau},\sigma,w|\mathcal{D}_n\big)dq\,d\sigma\,dw.\label{post}
		\end{equation}
		Here the crux is to compute the posterior $$p\big(q_{\tau},\sigma,w|\mathcal{D}_n\big)\propto   p\big(\mathcal{Y}_n|q_{\tau},\sigma,w,\mathcal{X}_n\big)   \,p\big(q_{\tau},\sigma,w\big).$$
		To do so, in \cite{abeywardana2015variational} the authors use $z=(q_\tau(\mathcal{X}_n), w, \sigma)$ as hidden variables and $\zeta=\boldsymbol{\theta}\in\mathds{R}^{d+1}$ as parameters and the variational factorization approximation 
		\begin{equation}
		p(q_\tau,\sigma,w|\mathcal{D}_n)\approx\tilde{p}(q_\tau,w,\sigma|\mathcal{D}_n)=\tilde{p}(q_\tau|\mathcal{D}_n)\;\tilde{p}(w|\mathcal{D}_n)\;\tilde{p}(\sigma|\mathcal{D}_n) .\label{approx}
		\end{equation} 
		The EM algorithm provides a nice formalism here. Although the goal is to find $\boldsymbol{\theta}$ such that $p\big(\mathcal{Y}_n|\mathcal{X}_n,\boldsymbol{\theta}\big)$ is maximal, the algorithm estimates the underlying GP (i.e. $p\big(q_{\tau},\sigma,w|\mathcal{D}_n\big)$) that is able to have a likelihood as large as possible. Then the estimated value $p(q_{\tau},\sigma,w|\mathcal{D}_n)$ is plugged into (\ref{post}) to provide the final quantity of interest.\\\\
		\paragraph{E-step.}
		Because the posteriors are conjugated, it is possible to obtain an analytical expression of the optimal distribution $\tilde{p}(q_{\tau})$: 
		$$\tilde{p}(q_{\tau}|\mathcal{D}_n)\sim \mathcal{N}\left(\mu_\theta,\Sigma_\theta \right),$$
		where
		$$	\mu_\theta=\Sigma_\theta\left( \boldsymbol{D} \mathcal{Y}_n - \dfrac{1-2\tau}{2}\mathds{E}\Big( \dfrac{1}{\sigma}\Big)\boldsymbol{1}\right)$$
		\text{ and}
		$$\Sigma_\theta= \left(\boldsymbol{D}+K_{x,x}^{\theta~~-1} \right)^{-1},$$
		\text{with}
		$$\boldsymbol{D}=\dfrac{\tau(1-\tau)}{2} \mathds{E}\Big( \dfrac{1}{\sigma^2}\Big) \diag\left(  \mathds{E}\Big( \dfrac{1}{w_i}\Big)\right)_{i=1,..,n}.$$
		The posterior on $w_i$ is a Generalized Inverse Gaussian \textbf{GIG}$(1/2,\alpha_i,\beta_i)$ with :
		$$\alpha_i=\left( \dfrac{(1-2\tau)^2}{2\tau(1-\tau)}+2\right) $$
		\text{ and } 
		$$
		\beta_i=\dfrac{\tau(1-\tau)}{2}\mathds{E}\Big( \dfrac{1}{\sigma^2}\Big)\Big(y_i^2-2y_i\mathds{E}\big( q_{\tau}(x_i) \big) + \mathds{E}\big( q_{\tau}(x_i)^2 \big)\Big).$$
		Due to numerical problems, in \cite{abeywardana2015variational} the authors  use the restriction $\tilde{p}(\sigma)=IG(a,b)$. 
		Finding the best $a,b$ is done numerically. Then finding the best $a,b$ is equivalent to maximizing:
		\begin{eqnarray}
		J(a,b)&=&(a-N-10^{-6})\log(b-\psi(a))\nonumber\\&+&(b-\gamma)\dfrac{a}{b}-\delta\dfrac{a(a+1)}{b^2}-a\log(b)+\log\Gamma(a),\nonumber
		\end{eqnarray}
		with $$\gamma=-\dfrac{1-2\tau}{2}\sum_{i=1}^{n}y_i-\mathds{E}\big( q_{\tau}(x_i) \big) $$ and 
		$$\delta=\dfrac{\alpha(1-\tau)}{4}\sum_{i=1}^{n}\mathds{E}\Big( \dfrac{1}{w_i}\Big)\Big(y_i^2-2y_i\mathds{E}\big( q_{\tau}(x_i) \big) +\mathds{E}\big( q_{\tau}(x_i)^2 \big) \Big). $$
		\paragraph{M-step.}
		Ignoring terms that do not depend on $\boldsymbol{\theta}$, we obtain the lower bound:
		\begin{eqnarray}
		\mathcal{\tilde{L}}(\boldsymbol{\theta})&=&\int \tilde{p}(q_{\tau}|\boldsymbol{\theta}) \,\tilde{p}(w) \,\tilde{p}(\sigma) \log p(y|q_{\tau},w,\sigma)p(q_{\tau}|\boldsymbol{\theta})d\sigma dw dq_{\tau}\nonumber\\&-& \int \tilde{p}(q_{\tau}|\boldsymbol{\theta})\log\tilde{p}(q_{\tau}|\boldsymbol{\theta})dq_{\tau}\nonumber\\
		&=&\dfrac{1}{2}\Big(\mu_\theta^T\Sigma_\theta^{-1}\mu_\theta-\log| \boldsymbol{D}^{-1} +K_{x,x}^{\theta}| \Big).\label{MVVB}
		\end{eqnarray}
		The optimization of $\mathcal{\tilde{L}}$ with respect to $\boldsymbol{\theta}$ is done using a numerical optimizer.
		
		Recalling the goal is to compute (\ref{post}),
		thanks to (\ref{approx}), we make the approximation:
		$$p(q_{\tau}|x_*,\mathcal{D}_n)\approx\int p\big(q_{\tau}(x_*)|q_{\tau},\sigma,\mathcal{D}_n\big)\, \tilde{p}(q_{\tau})\;\tilde{p}(\sigma)\,\tilde{p}(w)\,dq_{\tau}
		dw d\sigma.$$
		Then we obtain 
		$$p(q_{\tau}|x_*,\mathcal{D}_n)\approx\mathcal{N}\big( \bar{q}_\tau(x_*),\mathds{V}_q(x_*)\big),$$
		where 
		\begin{eqnarray*}
			\bar{q}_\tau(x_*)&=&K_{x_*,x}^{\theta}K_{x,x}^{\theta~~-1}\mu_\theta \text{ and}\nonumber \\
			\mathds{V}_q(x_*)&=&k_{\theta}(x_*,x_*)-K_{x_*,x}^{\theta}K_{x,x}^{\theta~~-1}K_{x_*,x}^{\theta~~T}+K_{x_*,x}^{\theta}K_{x,x}^{\theta~~-1}\Sigma_\theta K_{x,x}^{\theta~~-1}K_{x_*,x}^{\theta~~T}.
		\end{eqnarray*}
		
			\begin{algorithm}
				\caption{Bayesian variational regression}
				\begin{algorithmic}[1]
					\State \textbf{Training}
					\Inputs{$\mathcal{D}_{n}$,$\tau$, $k_{\theta_0}$}
					\Initialize{\strut
						\text{Compute the $n\times n$ matrix}~~$K_{x,x}^{\theta}$ and $K_{x,x}^{\theta~~-1}$\\
						$\boldsymbol{\theta}=\boldsymbol{\theta}_0$
					}
					\For{t = 1 to $n_{\It}$}
					\State \textbf{E-step}
					\State Compute $\Sigma_\theta$, $\mu_\theta$, $\alpha_i$, $\beta_i$, $w_i$,
					 $(a,b)$
					\State \textbf{M-step}
					\State $\boldsymbol{\theta}=\arg\max\dfrac{1}{2}\Big(\mu_\theta^T\Sigma_\theta^{-1}\mu_\theta-\log|\boldsymbol{D}^{-1} +K_{x,x}^{\theta}| \Big)$
					\EndFor
					\State \textbf{Prediction}
					\Inputs{$\mathcal{X}_{\test}$, $\boldsymbol{\theta}^*=\boldsymbol{\theta}$, $\mu_{\theta^*}$}
					\For{each point in $x^*\in\mathcal{X}_{\test}$}
					\State	$\hat{q}_{\tau}(x^*)=K_{x^*,x}^{\theta^*}K_{x,x}^{\theta^*~-1}\mu_{\theta^*}$
					\EndFor
				\end{algorithmic}\label{VB}
			\end{algorithm} 
			Finally, as explained in section \ref{QK}, the quantile estimator $\hat{q}_{\tau}$ is selected as $\bar{q}_{\tau}$.
			Algorithm \ref{VB} details the implementation of the VB method.
			
			\subsubsection{Computational complexity}
			\paragraph{E-step.}
			
			The complexity of this step is in $$O(n^3).$$ In fact the algorithm computes $\Sigma_\theta$ that implies inversing a matrix of size $n\times n$. \\\\
			\paragraph{M-step.}
			
			Optimizing $\mathcal{\tilde{L}}$ with a Newton type algorithm costs $O(n^3+d^3)$ at each iteration (for details refer to the optimization description of (\ref{likel})). Assuming the starting point $\theta_{\start}$ is close to an optimal $\theta^*$, based on the same analysis as in section \ref{compnn}, the whole complexity is in $$O\big(L(d^3+n^3)\big).$$	
			\paragraph{Overall complexity.}
			At each iteration of the EM algorithm, the computation cost is $O(L(d^3+n^3)+n^3)$. The final complexity is obtained by multiplying by the number of iterations $n_{\It}$ of the EM algorithm. 
			Thus, the overall complexity is in $$O\big(n_{\It}(L(d^3+n^3)+n^3)\big).$$

			\section{Metamodel summary and implementation}\label{sec:implementation}
			In this section we detail our implementation procedure. After providing a summary of the six metamodels in Table \ref{tab:summary}, we present the packages we used and the hyperparameters we chose (which hyperparameters we set and which hyperparameters we optimized). We then describe the procedure we used to optimize the hyperparameters (optimization strategies and evaluation metrics).
			\subsection{Summary of the models }\label{sec:summary}
			Table \ref{tab:summary} lists the analytical expressions of the six metamodels, along with the associated underlying quantity.
		
			\begin{table*}
				
				\hspace{-1cm}
				\begin{tabular}{|c|c|c|c|}
					\hline
					Method& Definition of $\hat{q}_\tau(x^*)$  &Related quantity&Complexity\\
					\hline
					&&&\\
					KN& $ \displaystyle{y_{([K\tau])}(x^*)}$ &The $K$-nearest points from $x^*$&$O\big(nN_{new}(d+\log n)\big)$\\
					&&&\\
					\hline
					&&&\\
					RF& $ \displaystyle{\inf\{y_i : \hat{F}(y_i|X=x^*)\geq \tau\}}$&  $ \displaystyle{\hat{F}(y_i|X=x^*)=\sum_{i=1}^{n}\bar{\omega}_i(x^*)\mathds{1}_{\{y_i\leq y\}}}$&$O(N\tilde{d}n\log^2 n)$\\
					&&&\\
					\hline
					& For a $3$ layer NN  &With $w_i^{(h_1)}$ , $w_{j}^{(h_2)}$, $b_i^{(h_1)}$, $b_j^{(h_2)}$, $w^{(h_3)}$, $b^{(h_3)}$,&\\
					
					&$\displaystyle{ g_3(  \sum_{j=1}^{J_2}g_{2}( \sum_{i=1}^{J_1}g_{1}(\left\langle w_i^{(h_1)},x^*\right\rangle }$  &$1\leq i\leq J_1$, $1\leq j\leq J_2$,  minimizing&$O\big(M_{s}L_{\eta^*}n(Jd)^3\big)$\\ NN &$ \displaystyle{+b_i^{(h_1)} )w_{j}^{(h_2)}+b_j^{(h_2)})w^{(h_3)}+b^{(h_3)})}$&$ \displaystyle{
						\dfrac{1}{n}\sum_{t=1}^{n}l_{\tau}(y_i-\hat{q}_{\tau}(x_i))+\sum_{j,i} \dfrac{\lambda}{J_j}\left\| w_i^{(h_j)}\right\| ^2
					}$ &\\
					&&&\\
					\hline
					
					&&With $\boldsymbol{\alpha}=(\alpha_1,\dots,\alpha_n)$ minimizing&\\
					&& $\dfrac{1}{2}\boldsymbol{\alpha}^T K_{x,x}^{\theta}\boldsymbol{\alpha}-\boldsymbol{\alpha}^T \boldsymbol{y}$&\\
					RK&$ \displaystyle{\sum_{i=1}^{n}\alpha_i  k_{\theta}(x^*,x_i) +b}$&\text{s.t}~$\dfrac{\tau-1}{\lambda n}\leq \alpha_i\leq\dfrac{\tau}{\lambda n},~\forall ~1\leq i\leq n$&$O\big(n^{7/2}\log(\frac{n}{\epsilon})\big)$\\
					&& and $\displaystyle{\sum_{i=1}^{n}\alpha_i=0}$ and $b$ the $\tau$-quantile of &\\
					&&$(y_i-\sum_{j=1}^{n}\alpha_j k_\theta(x_i,x_j))_{1\leq i \leq n}$&\\
					\hline

					&&Maximizing the likelihood:&\\
					
					QK& $ K_{x^*,x}^{\theta} (K_{x,x}^{\theta}+B)^{-1}\hat{q}_{\tau}$& $ p(\mathcal{Q}_{n'}|\mathcal{X}_{n'})$&$O\big(L(d^3+n'^3)+n'^3\big)$\\
					&&$\hat{q}_{\tau }(x_i)=q_{\tau}(x_i)+\varepsilon_i~~~~\varepsilon_i\sim\mathcal{N}(0,B_{ii})$&\\
					\hline
					&&Approached solution that maximize:&\\
					VB& $K_{x^*,x}^{\theta}K_{x,x}^{\theta~~-1}\mu_\theta$ &  $p(\mathcal{Y}_{n}|\mathcal{X}_n)$&$O\Big(n_{\It}\big(L(d^3+n^3)+n^3\big)\Big)$\\
					&&$y_i=q_\tau(x_i)+\varepsilon~~~~\varepsilon\sim \text{ALP}(0,\sigma)$&\\
					\hline
				\end{tabular}

				\caption{Summary of the metamodels, included the definition of the estimators, the associated numerical quantity and the related computation complexity. In this table $L$ and $L_{\eta^*}$ are Hessian Lipschitz constants, $M_s$ the number of multistarts (on the weights at hyperparameters fixed), $d$ the input dimension, $N_{new}$ the size of the prediction set, $n_{\It}$ the number of iterations for the EM algorithm, $k_\theta(.,.)$ the kernel function, $B$ a diagonal matrix, $J$ the
total number of neurons, $\tilde{d}$ the number of covariables considered to find the best splitting point.}	\label{tab:summary}
			\end{table*}
		
			\subsection{Packages and hyperparameter choices}
			Each method depends on many parameters that can be tuned to improve performance, for example the choice of the kernel function and the value of its parameters for RK, QK and VB or the penalization factor for RK and NN.  Here, to limit the computational burden, we chose to optimize only the most critical ones. When possible, for the other parameters, we applied the arbitrary choices and values made by the authors of the original papers. Most changes were made to improve robustness. Below, we describe our experimental settings, also summarized in Table \ref{tab:hyperparameters}.\\\\
			\paragraph{Nearest Neighbors.}  
			We set $d()$ as the Euclidean distance and optimized only the size $K$ of the neighborhood.
			\\\\
			\paragraph{Random forest.}  In this case, the only hyperparameter we optimized was the maximum size of the leaves $m_s\in\mathds{N}^*$. Regarding the number of trees, we noticed that the metamodel needs many more trees than are needed for the estimation of the expectation. In some problems, the metamodel needs up to 5,000 trees to stabilize the variance. Thus, in our experiments we set the number of trees at 10,000 in all cases. We set the number of dimensions considered for the split at the default choice $d/3$. The depth of the tree is not constrained and the splitting rule is based on Eq. \ref{cart}. We used the R package \textit{QuantregForest} \cite{meinshausen2007quantregforest}.
			\\\\
			\paragraph{Neural network.} Based on \cite{cannon2011quantile}, we set the number of hidden layers at one and the transfer function as the sigmoid. The optimization algorithm is a Newton method, we set $E_K$ at $1/2^K$ with $K=1,2,5,10,15,20,25,30,35$ and the number of multistarts to optimize the empirical risk at five. We optimized the number of neurons $J_1$ in the hidden unit and the regularization parameter $\lambda \in\mathds{R_+}$. The metamodel is generated using the R package qrnn \cite{cannon2011quantile}.
			\\\\
			\paragraph{Regression in RKHS.} The kernel was set as a Mat\'{e}rn 5/2. We optimized the length scale parameters $\theta\in\mathds{R}_+^d$ and the regularization hyperparameter $\lambda\in\mathds{R_+}$. Optimization (\ref{eq:RKquad}) is done with the quadratic optimizer quadprog \cite{turlach2007quadprog}.
			\\\\
			\paragraph{Quantile Kriging.} The kernel was set as a Mat\'{e}rn 5/2. The number of repetitions was set according to the total number of observations (see Table \ref{tab:datasize}). We optimized the length scale hyperparameter $\theta\in\mathds{R}_+^d$ and variance hyperparameter $\rho\in$ $\mathds{R}_+$. QK is implemented in the R package \textit{DiceKriging} \cite{roustant2012dicekriging}.
			\\\\
			\paragraph{Variational regression.}
			The kernel was set as a Mat\'{e}rn 5/2, the number of EM iterations $n_{\It}$ at $50$. We optimized the length scale hyperparameter $\theta\in\mathds{R}_+^{d}$ and variance hyperparameter $\rho\in$ $\mathds{R}_+$. The implementation is based on the Matlab code provided in \cite{abeywardana2015variational}.
			
			\begin{table}
				\centering
				\begin{tabular}{|c|c|}
					\hline
					Method& Hyperparameters \\
					\hline
					KN& number of neighbors $K\in\mathds{N}^*$ \\
					\hline
					RF& maximum size of the leaves $m_s\in\mathds{N}^*$\\
					\hline
					NN &  regularization $\lambda\in\mathds{R_+}$,  $J_1\in\mathds{N^*}$  \\
					\hline
					RK & regularization $\lambda\in\mathds{R_+}$, lengthscales $\theta\in\mathds{R_+}^d$ \\ 
					\hline
					QK& length scale and variance $\theta\in\mathds{R_+}^{d+1}$\\
					\hline
					VB& length scale and variance $\theta\in\mathds{R_+}^{d+1}$\\
					\hline
				\end{tabular}
				\caption{Hyperparameters optimized on our benchmark. }\label{tab:hyperparameters}
			\end{table}	
			
			\subsection{Tuning the hyperparameters}
			In the previous section, we defined the hyperparameters we wanted to optimize for each method. In fact, once the type of metamodel is chosen, the quantile estimator is given by a function $\hat{q}_{\Theta}:\mathcal{X}\rightarrow \mathds{R}$ where $\Theta\in\mathds{R}^v$ are called hyperparameters and $v$ is metamodel dependent.
			Hyperparameter optimization (also known as model selection) is an essential procedure when dealing with non-parametric estimators. Although $\hat{q}_{\Theta}$ may be very efficient on $\mathcal{D}_n$, the prediction may perform very poorly on an independent dataset $\mathcal{D}'_p$. The goal is to find the ${\Theta}$ that provides the best possible generalized estimator. In the following, we present the validation metric used to optimize the hyperparameter values and detail the hyperparameters optimization procedure associated with each method.
			
			\subsubsection{Metrics}
			In the standard conditional expectation estimation, the validation and performance metrics are both based on $\left\| \hat{m}_\Theta-m\right\| _{L^2}$, where $\hat{m}_\Theta$ is the estimator and $m$ the targeted value. With the quantile estimation procedure the two metrics are no longer the same. The goal is to find $\hat{q}_\Theta$ such that $\left\| \hat{q}_\Theta-q\right\| _{L^2}$ is as small as possible. However, $q$ is unobserved so the validation metric cannot be based on the $L^2$ norm. Here we present two metrics able to measure the generalization capacity of a quantile metamodel.
			
			Bayesian metamodels (QK and VB) have their own validation metric, this is the likelihood function that can be maximized with respect to $\Theta$. For quantile kriging, we use (\ref{likel}) while in the variational approach we use (\ref{MVVB}). The optimal hyperparameters are then:
			
			\begin{equation}
			\Theta_{mv}^*=\argmax_\Theta p\big(\mathcal{Y}_n|\mathcal{X}_n,\Theta\big).
			\end{equation}	
			
			The second metric available for all metamodels is $k$-fold cross-validation associated with the pinball loss. The metric can be computed as follows. First, the data are split into $k$ parts, then the model is trained on $\mathcal{D}_{-j}$, the training set without the $j$-th part. The performance is evaluated on the remaining part $\mathcal{D}_{j}$. As the quantile minimizes the pinball loss (on $\mathcal{D}_{j}$), the evaluation metric is
			\begin{equation}
			E_{cv}(\hat{q}_{\tau}^{\Theta})=\dfrac{1}{k}\sum_{j=1}^{k}\dfrac{1}{n_j'}\sum_{i=1}^{n_j'}l_{\tau}\big((y_i-\hat{q}_{\tau}^{\Theta}(x_i)\big),\label{cv}
			\end{equation}
			where $n_j'$ is the number of observations in each fold. The optimal cross-validation hyperparameters are then:
			$$\Theta_{cv}^* \in \argmin_\Theta E_{cv}(\hat{q}_{\tau}^{\Theta}).$$
			
			In our experiments, we chose $k=5$ to limit the computational cost. However, we observed empirically that choosing a larger $k$ did not substantially modify the performances of the metamodels.
			Although cross-validation is available for QK, we chose to stay in the spirit of the methods and to only use likelihood to select hyperparameters. Our choice is supported by \cite{bachoc2013parametric} that does not show a clear improvement using cross validation instead of maximum likelihood techniques.
			\subsubsection{Hyperparameters optimization procedure}
		Both likelihood functions come with analytical derivatives, enabling the use of gradient-based algorithms. However, since both functions are multi-modal, 
			multi-start techniques are necessary (and generally very efficient, see \cite{hansen2009benchmarking}) to avoid being trapped in local optima. 
			To account for the increasing difficulty of the optimization task with the dimension while limiting the computational cost, 
			we ran $n_{\start}=20d$ optimization procedures from different starting points $\theta_{\start}$ and chose the set of starting points based on a \textit{maximin} Latin hypercube design.

			For QK, the BFGS algorithm is used to optimize (\ref{likel}). For VB, two derivative-based optimizers are used alternately for the E- and M-steps. Since each step may lead the algorithm toward a local minimum, we chose to apply the multi-start strategy in the entire EM procedure.
			
			Optimization of the cross-validation metric (\ref{cv}) is done under the black box framework, since no structural, derivative or even regularity information is available. Hence, all optimizations are carried out using the branch-and-bound algorithm named Simultaneous Optimistic Optimization (SOO) \cite{munos2011optimistic}. SOO is a global optimizer, hence robust to local minima. 
			
			We used \cite{tange2018gnu} to parallelize the computations.
			
			\subsubsection{Oracle metamodels}
			Each method presented in this paper is a trade-off between power and the difficulty of finding good hyperparameters.  A good method should be powerful (i.e. provide flexible fits) but easy to tune. In order to assess the strengths and weaknesses of the hyperparameter tuning methods in addition to standard metamodels, we provide what we call the \textit{oracle metamodels} for each problem. Instead of using the cross-validation or likelihood metric, the oracle tunings are directly based on the evaluation metric 
				\begin{equation}
			E_{L^2}(\hat{q})=\sum_{i=1}^{n_{\text{test}}}\big(\hat{q}^{\Theta}(x_i)-q(x_i)\big)^2,\label{l2}
			\end{equation}
			where $n_{\text{test}}$ is the size of the test set. In a sense, they provide a upper bound on the performance of each method. This allows us to show which metamodels have the potential to tackle the problems and which are intrinsically too rigid or make poor use of information.
			In addition, this allows us to directly assess the quality of the validation procedure. 
			\section{Benchmark design and experimental setting}\label{sec:benchmarkdesign}
			Many factors can affect the efficiency of methods to estimate the right quantile. For our benchmark system, we considered five models or test cases to evaluate the performance of the six metamodels. We decided to focus primarily on the dimensionality of the problem, the number of training points available, the signal-to-noise ratio defined as
			$$\snr=\dfrac{\mathds{V}_{X}\big(\mathds{E}(Y|X)\big)}{\mathds{E}_{X}\big(\mathds{V}(Y|X)\big)},$$
		 and the pdf value at the targeted quantile for test cases in which the distribution shape and the distribution spread (i.e. level of heteroscedasticity) can vary significantly.
			Our two objectives were to:
			\begin{enumerate}
				\item discover if there is a single best method for all factors variations considered or specific choices depending on the configuration at hand, and
				\item assess the performance of the quantile regression, and in particular, the configurations for which the current state-of-the-art is insufficient.
			\end{enumerate}
			
			A full 3D factorial experimental design was used to analyze the efficiency of the metamodels, the 3 factors being the test case (4 test cases), the number of training points (4 levels) and the quantile order (0.1, 0.5 and 0.9). We used part of this complete design to focus our analyses on the characteristics of the test cases (dimension, pdf shape and heteroscedasticity).

			\subsection{Test cases and numerical experiments}
			\paragraph{Test case 1} is a $1$d toy problem on $\left[ -1,1\right] $ defined as
			\begin{equation}
			Y_x = 5\sin(8x) + (0.2+3x^3)\xi, \nonumber
			\end{equation}
			with $\xi=\eta\mathbb{1}_{\eta\leq0}+7\eta\mathbb{1}_{\eta> 0}\text{ where }\eta\sim \mathcal{N}(0,1).$
			
			The signal-to-noise ratio is $\snr\approx 0.5$, it is consider as small. The pdf value $f(x,q_\tau)$ varies substantially according to $x$ for all $q_\tau$. Indeed, on the interval $[-0.5,-0.3]$, for all values of $q_\tau$, the pdf is very large (almost equal to $+\infty$) because the variance of the distribution on this interval is very small. In contrast, for $\tau=0.9$ (resp. $\tau=0.1$) the pdf is very small in the interval $[0.6,1]$ (resp. $[-1,-0.6]$). In $[0.6,1]$ the pdf of the $0.9$-quantile is equal to the pdf of the $0.9$-quantile of a normal distribution of variance $49(0.2+3x^3)^2$ that is, for example, approximately equal to $0.01$ for $x=0.9$. Because of this important variations, we consider the values of the pdf for all quantiles of interest as variable.\\
			
			\paragraph{Test case  2} is a $2$d toy problem on $\left[ -5,5\right]\times\left[ -3,3\right]$ based on the Griewank function \cite{dixon1978global}, defined as
			\begin{equation}
			Y_x =  G(x)\xi, \nonumber
			\end{equation}
			with 
			$$G(x)=\left[\sum_{i=1}^{2}\dfrac{x_i^2}{4000}-\prod_{i=1}^{2}\cos\left(\dfrac{x_i}{\sqrt{i}}\right)+1 \right]$$
			and $\xi=\eta\mathbb{1}_{\eta\leq 0}+5\eta\mathbb{1}_{\eta> 0}\text{ where }\eta\sim \mathcal{N}(0,1).$
			
			The signal-to-noise ratio is $\snr=0$, it is consider as small. The pdf value $f(x,q_\tau)$ varies substantially according to $q_\tau$. Indeed, for $\tau=0.1$ (resp. $\tau=0.9$) the pdf is small (resp. very small), more precisely the pdf of the $0.1$-quantile (resp. $0.9$-quantile) is equivalent to the pdf of the $0.1$-quantile (resp. $0.9$-quantile) of a normal distribution of variance $G^2$ (resp. $25G^2$) with $G$ that takes values in $[0,2]$. That implies $f(x,q_\tau)$ varies with respect to $x$ as well. Note that close to $x=(0,0)$ the pdf is very large because of the very small variance of the associated distribution. The pdf at the $0.5$-quantile is discontinuous. To the left of the $0.5$-quantile the pdf is equal to the pdf at the median of a normal distribution of variance $G^2$ while to the right it is equal to the pdf at the median of a normal distribution of variance $25G^2$. To treat this discontinuity, the value of the pdf is considered as the maximum of both sides. Thus, for this particular case the pdf is considered as large.  \\
		
			\paragraph{Test case  3} is a $1$d toy problem based on the Michalewicz function \cite{dixon1978global} on $\left[ 0,4\right]$, defined as
			\begin{equation}
			Y_x = -2\sin(x)\sin^{30}\left(\dfrac{x^2}{\pi} \right) - \frac{0.1\cos(\pi x/10)^3}{\left|-\sin(x)\sin^{30}\left(\frac{x^2}{\pi}\right)+2\right|}\xi^2,\nonumber
			\end{equation}
			with $\xi=3\eta\mathbb{1}_{\eta\leq 0}+6\eta\mathbb{1}_{\eta> 0}\text{ where }\eta\sim \mathcal{N}(0,1)$.
			
			The signal-to-noise ratio is $\snr\approx0.04$, we consider it as small. The pdf value $f(x,q_\tau)$ varies substantially according to $q_\tau$ and $x$. The conditional distribution of this problem is not a classical one but it is close to the distribution of $-\mathcal{X}^2$ with one degree of freedom. It implies at $x$ fixed, the pdf value increases with $\tau$. For the $0.1$-quantile (resp. $0.9$-quantile) the mean of the pdf in the interval $[0,2.5]$ is approximately $0.05$ (resp. $9.1$). 
			According to $x$ the pdf varies as well. For $x\in[3.5,4]$ the variance of the conditional distribution is very small thus the pdf near the quantiles of interest is very large.
			The value of the pdf at the $0.5$-quantile is between the value of the pdf at the $0.1$ and $0.9$ quantile. Thus, we consider the pdf at $q_{0.1}$ as globally small and at $q_{0.5}$ and at $q_{0.9}$ as globally large.\\
			
				\paragraph{Test case  4} is a $9$d toy problem based on the Ackley function \cite{ackley1987connectionist} on $\left[ -1,-0.7\right]\times\left[ 0,1\right]\times\left[ -0.7,-0.3\right]\times\left[ 0.5,1\right]\times\left[ -1,-0.5\right]\times\left[ -3,-2.6\right]\times\left[ -0.1,0\right]\times\left[ 0,0.1\right]\times\left[ 0,0.8\right]$, defined as a function $$Y_x=30\times A(x)+ R(x)\times \xi$$ with
			\begin{equation}
			A(x) = a\exp\bigg(-b\sqrt{\frac{1}{9}\sum_{i=1}^{9}x_i^2}\bigg)-\exp\bigg(\frac{1}{9}\sum_{i=1}^{9}\cos(c x_i)\bigg)+a+\exp(1),
			\end{equation}
			and
			\begin{equation}
			R(x)=3A(x_2,x_3,\cdots,x_9,x_1),
			\end{equation}
			with $a=10$, $b=2\times10^{-4}$, $c=0.9\pi$ and $\xi$ follows a log-normal distribution of parameters $(0,1)$. 
			
			The signal-to-noise ratio is $\snr\approx 2.3$, it is consider as large. The pdf value $f(x,q_\tau)$ varies substantially according to $q_\tau$ and $x$. Indeed, for $\tau=0.1$ (resp. $\tau=0.9$) the pdf is large (resp. very small). At the conditional distribution of this test case is a log-normal distribution of parameters $\big(\log R(x),1\big)$. The expectations of the pdf value at different values of $\tau$ are $\mathds{E}(f(.,q_{0.1}))=0.1$, $\mathds{E}(f(.,q_{0.5}))=0.04$ and $\mathds{E}(f(.,q_{0.9}))=0.005$. Thus, the pdf at $q_{0.1}$ is consider as large, the pdf at $q_{0.5}$ and $q_{0.9}$ are consider as small and very small.
			
\begin{figure*}[!ht]
				
					\includegraphics[width=1\textwidth]{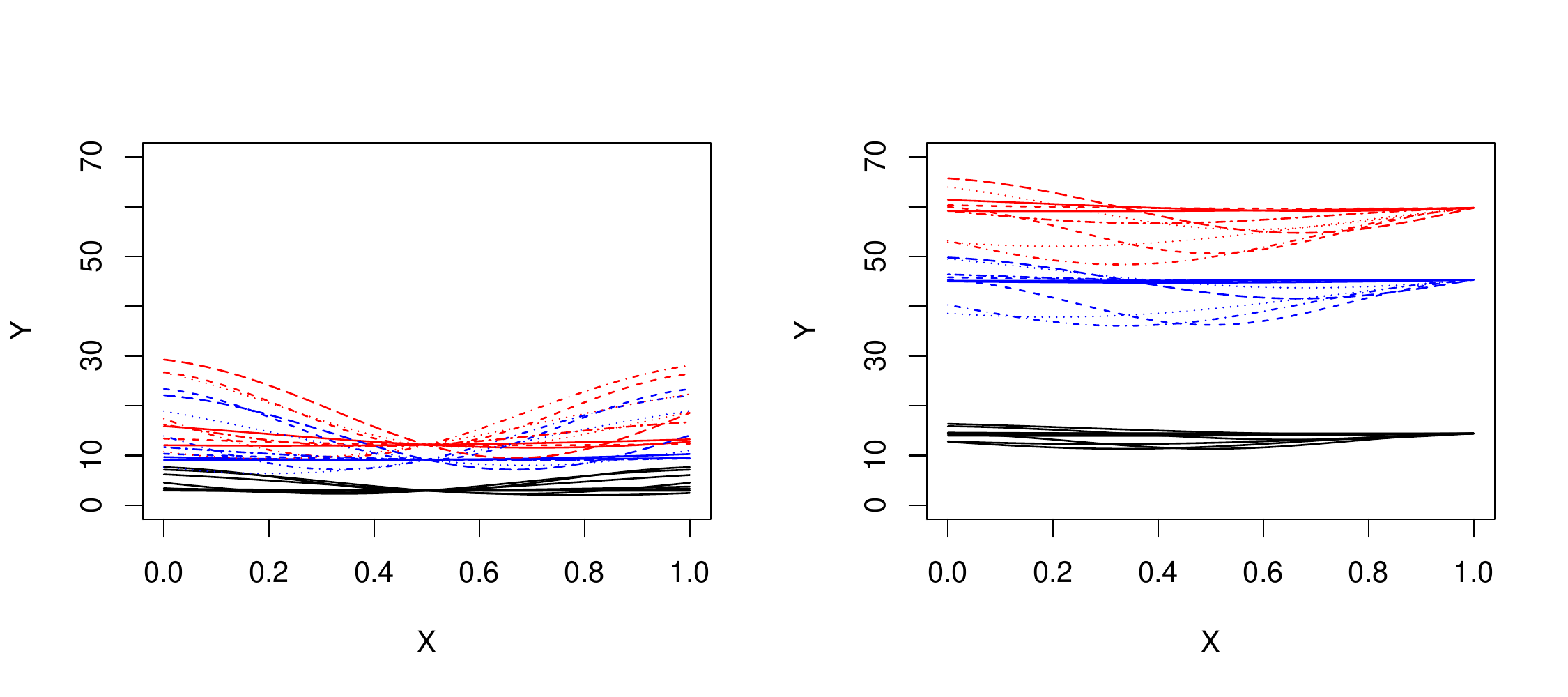}
					
				\caption{Illustration of some marginals of Test case $5$. The conditional quantiles of order $0.9$ (resp. $0.1$) are represented in red (resp. in blue). The black curves represent the difference between the $0.1$-conditional quantiles and the $0.9$-conditional quantiles so that to measure the level of heteroscedasticity. To the right the noise level is low $i.e$ between $0$ and $10$ while to the left the noise level is higher $i.e$ between $10$ and $20$.}\label{toy_prob5}
			\end{figure*}

			To provide a better intuition about this problem, we plotted what we call the marginals. For all dimensions except the $j$-th, the values of the input are fixed to $x_{-j}\in \mathds{R}^8$ and the $j$-th dimension varies. Figure \ref{toy_prob5} represents the evolution of the quantiles w.r.t. the $j$-th dimension for two different $x_{-j}$ and for $j=1,\cdots,9$. In particular it shows that the difference between the $0.1$ and $0.9$-quantile depends significantly on $x$.\\
			
\begin{figure*}[!ht]
				\begin{tabular}{ccc}
					\includegraphics[width=.33\textwidth]{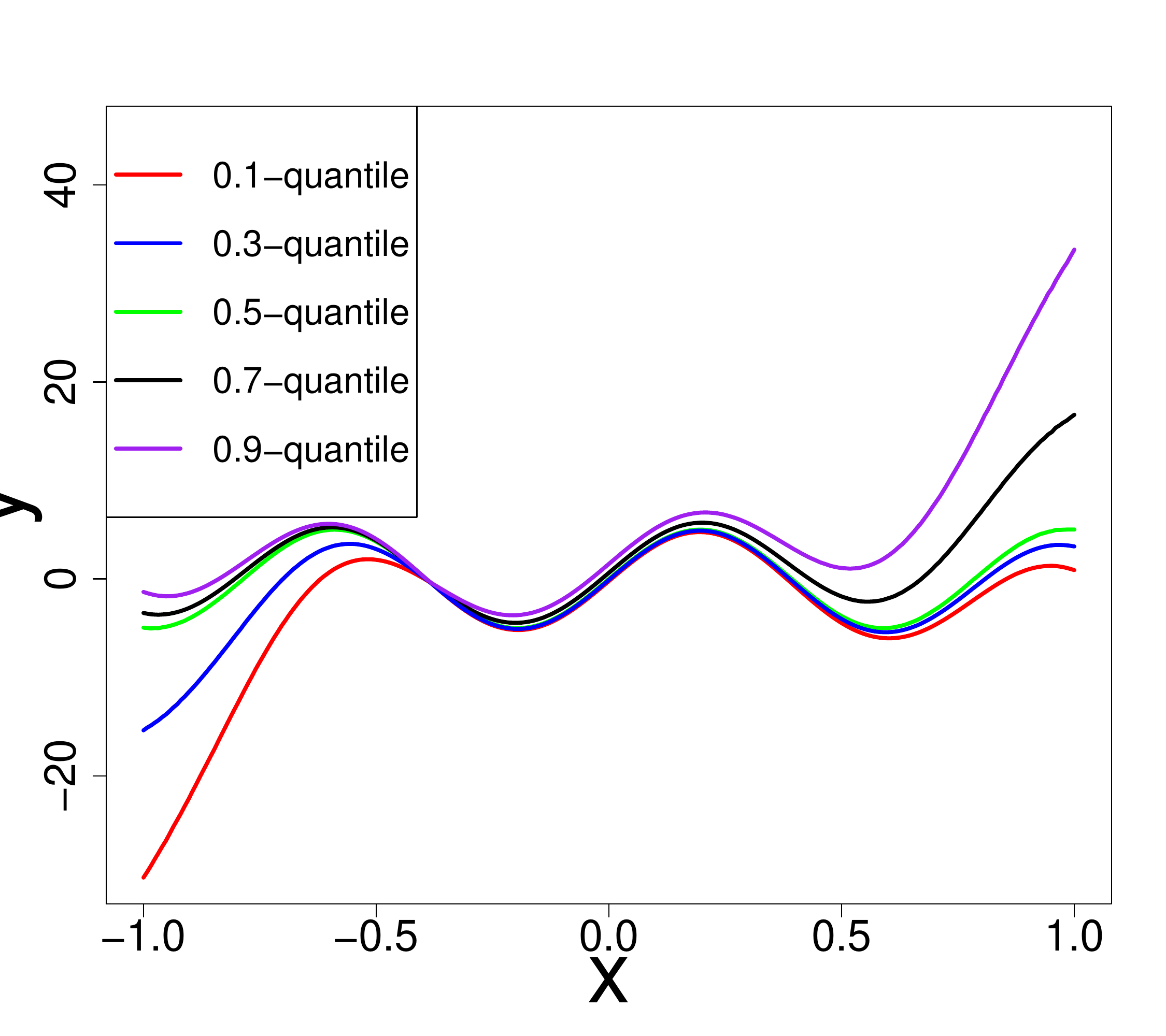}
					\includegraphics[width=.33\textwidth]{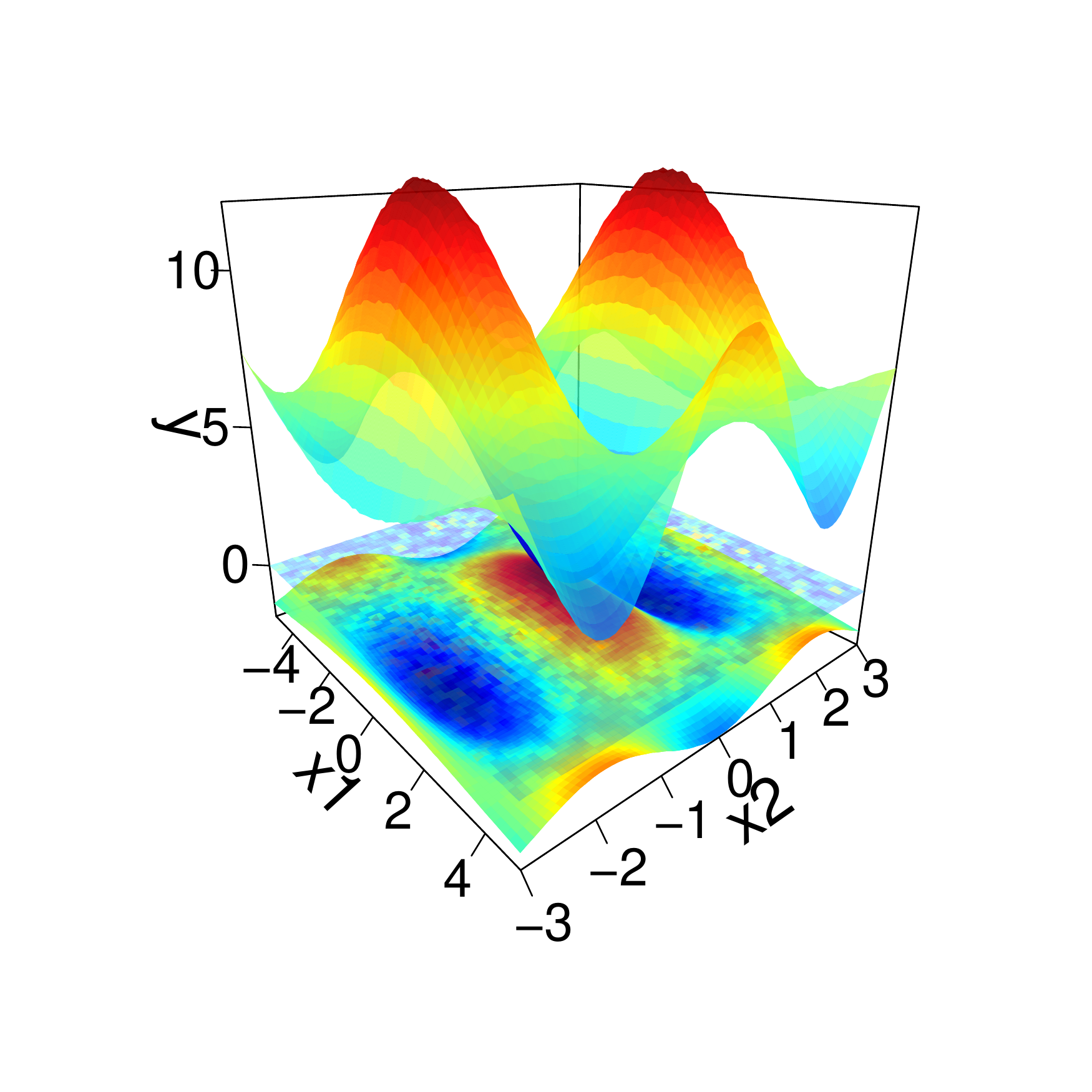}
					\includegraphics[width=.33\textwidth]{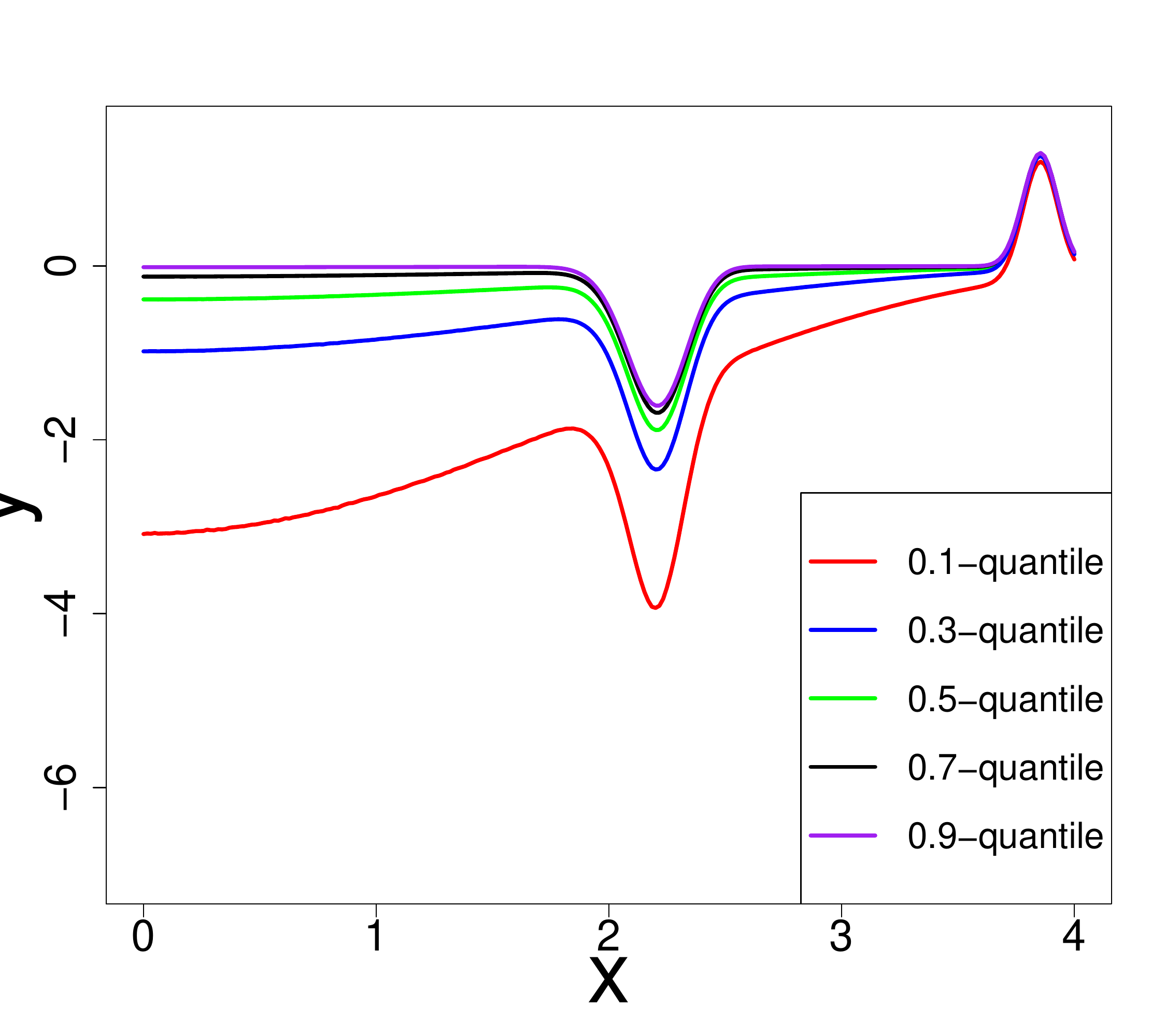}
				\end{tabular}
				\caption{Illustration of the test cases (left: test case 1, center: test case 2, right: test case 3). For test case $1$ and $3$ the $0.1, 0.3, 0.5, 0.7, 0.9$ -quantiles are represented. For test case 2, only the $0.1,0.5,0.9$-quantiles are represented.}\label{toy_prob}
			\end{figure*}

			Note that on those four toy problems, the random term $\xi$ is defined such that the resulting distribution of $Y$ would be strongly asymmetric.
			As $\xi$ is also multiplied by a factor that depends on $x,$ the distribution of $Y$ is also heteroscedastic. 
			The  first three toy problems are represented in Figure~\ref{toy_prob} and some illustrations of the fourth test case are available Figure \ref{toy_prob5}.
			\\
			
			\paragraph{Test case  5} is based on the agronomical model SUNFLO, a process-based model which was developed to simulate sunflower grain yield (in
			tons per hectare) as a function of climatic time series, environment (soil and climate), management practices and genetic diversity.
			The full description of the model is available in \cite{casadebaig2011sunflo}.
			In the regression model we consider $\mathcal{X}$ corresponding to nine macroscopic traits that characterize the sunflower variety. 
			Although SUNFLO is a deterministic model, for each simulation the climatic time series are randomly chosen within a database containing $190$ years of weather records,
			which makes the output stochastic (see also \cite{picheny2017optimization} for more details). 
			
				The signal-to-noise ratio is $\snr\approx 0.1$. The pdf value $f(x,q_\tau)$ varies substantially according to $q_\tau$, more precisely $\mathds{E}(f(.,q_{0.1}))=0.17$, $\mathds{E}(f(.,q_{0.5}))=0.15$ and $\mathds{E}(f(.,q_{0.9}))=0.05$. Thus, the pdf at $q_{0.1}$ and at $q_{0.5}$ are consider as large, and the pdf at $q_{0.9}$ is consider as small.
				
\begin{figure*}[!ht]
				\begin{tabular}{cc}
					\includegraphics[trim=0mm 15mm 0mm 16mm, clip, width=.45\linewidth]{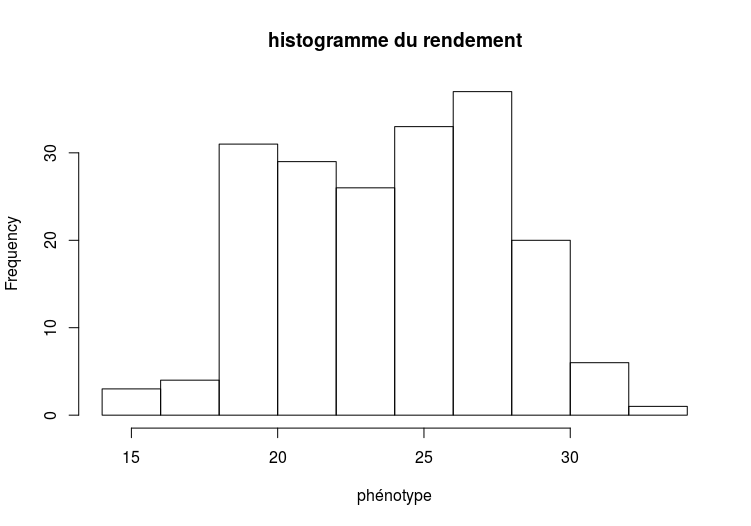}
					\includegraphics[trim=0mm 15mm 0mm 16mm, clip, width=.45\linewidth]{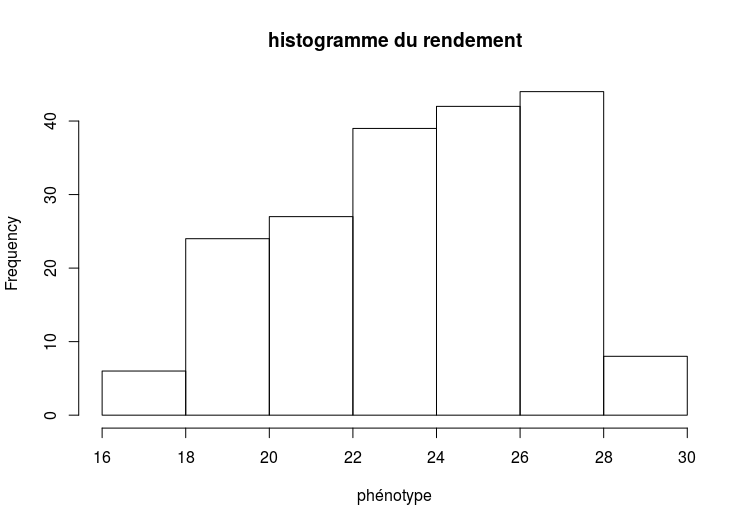}
				\end{tabular}
				\caption{Output of the model SUNFLO with two different inputs $x_1,~x_2\in\mathds{R}^9$ evaluated over $190$ different climatic time series. }\label{heterosunflo}
			\end{figure*}

				In addition, the shape of the distribution varies significantly with respect to the input as illustrated Figure \ref{heterosunflo}. \\
			
			\paragraph{Numerical experiments.}
			On all problems, we consider four sample sizes. Those sizes depend on the dimension and are empirically chosen so that the smallest size corresponds to the minimal information required by the metamodels to work and the largest size is chosen keeping in mind the potentially high cost of simulators. Besides, our focus is on computer experiments, where data sizes rarely exceed thousands of points. 
			For the $1d$ problems, the points are generated on a uniform grid. For the $2d$ and $9d$ problems, the observations are taken on a Latin hypercube design 
			optimized for a \textit{maximin} criterion to ensure space-filling \cite{fang2005design}. The same samples are used by all methods except QK, as it requires repetitions. 
			For QK, the number of distinct points and number of repetitions depends on the budget. The different sample sizes are reported in Table~\ref{tab:datasize}.
			Finally, for each sample size and problem, 10 samples are drawn in order to assess robustness with respect to the data.
			
			\begin{table*}
			
				\begin{center}
					\begin{tabular}{|c|c|c|c|c|c|c|}
						\hline
						\multicolumn{2}{|c|} {}& Test case 1  &Test case 2&Test case 3&Test case 4&Sunflo\\
						\hline
						\multicolumn{2}{|c|} {Dimension} & $1$ & 2 & 1 & 9 &9 \\
						\hline
						\multicolumn{2}{|c|} {Heteroscedasticity} &very strong &very strong&very strong&strong&weak\\
						\hline
						\multicolumn{2}{|c|} {Shape variation}& very strong&weak&weak&weak& strong\\
						\hline
						pdf value  &$\tau=0.1$ &variable  & small  &globally very small &large&large\\ \cline{3-7}
						near the & $\tau=0.5$ &variable &large &small & small&large  \\ \cline{3-7}
						$\tau$-quantile & $\tau=0.9$&variable &very small &very large & very small&small  \\ \hline
					\end{tabular}
				\end{center}
				\caption{Summary of the characteristic of the problems.  }\label{tablegroup1}
			\end{table*}
			
			\begin{table}
				\begin{center}
					\begin{tabular}[b]{|c|c|c|c|c|c|c|c|c|}
						\hline
						Dimension & \multicolumn{4}{c|}{Data size (no repetitions)} & \multicolumn{4}{c|}{Data size (with repetitions)}\\
						\hline
						1 & 40 & 80 & 160 & 320 & 5 (8) & 10 (8) & 10 (16) & 16 (20)\\ 
						2 & 100 & 200 &400 & 800 & 10 (10) & 20 (10) & 25 (16) & 40 (20)\\ 
						9 & 250 & 500 & 1000 & 2000 & 25 (10) & 50 (10) & 100 (10) & 100 (20)\\
						\hline 
					\end{tabular}
					\caption{Data sizes for the different problems. The number in parentheses are the number of repetitions for QK.}\label{tab:datasize}
				\end{center}
			\end{table}
			
			\subsection{Structuration between the questions and the numerical setting}\label{gr}
			
\begin{figure*}[!ht]
				\begin{tabular}{cc}
					\includegraphics[width=.45\textwidth]{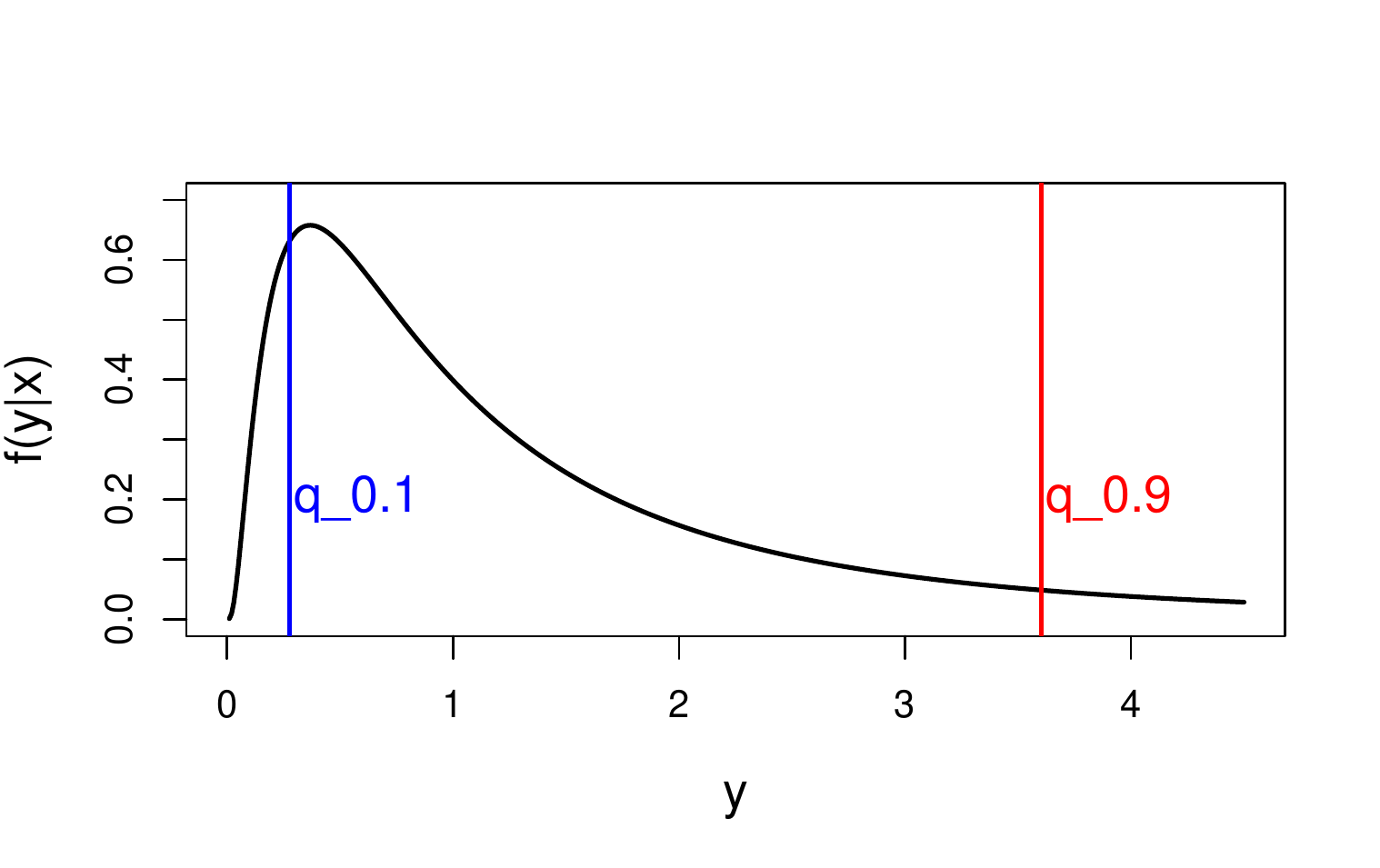}
					\includegraphics[width=.45\textwidth]{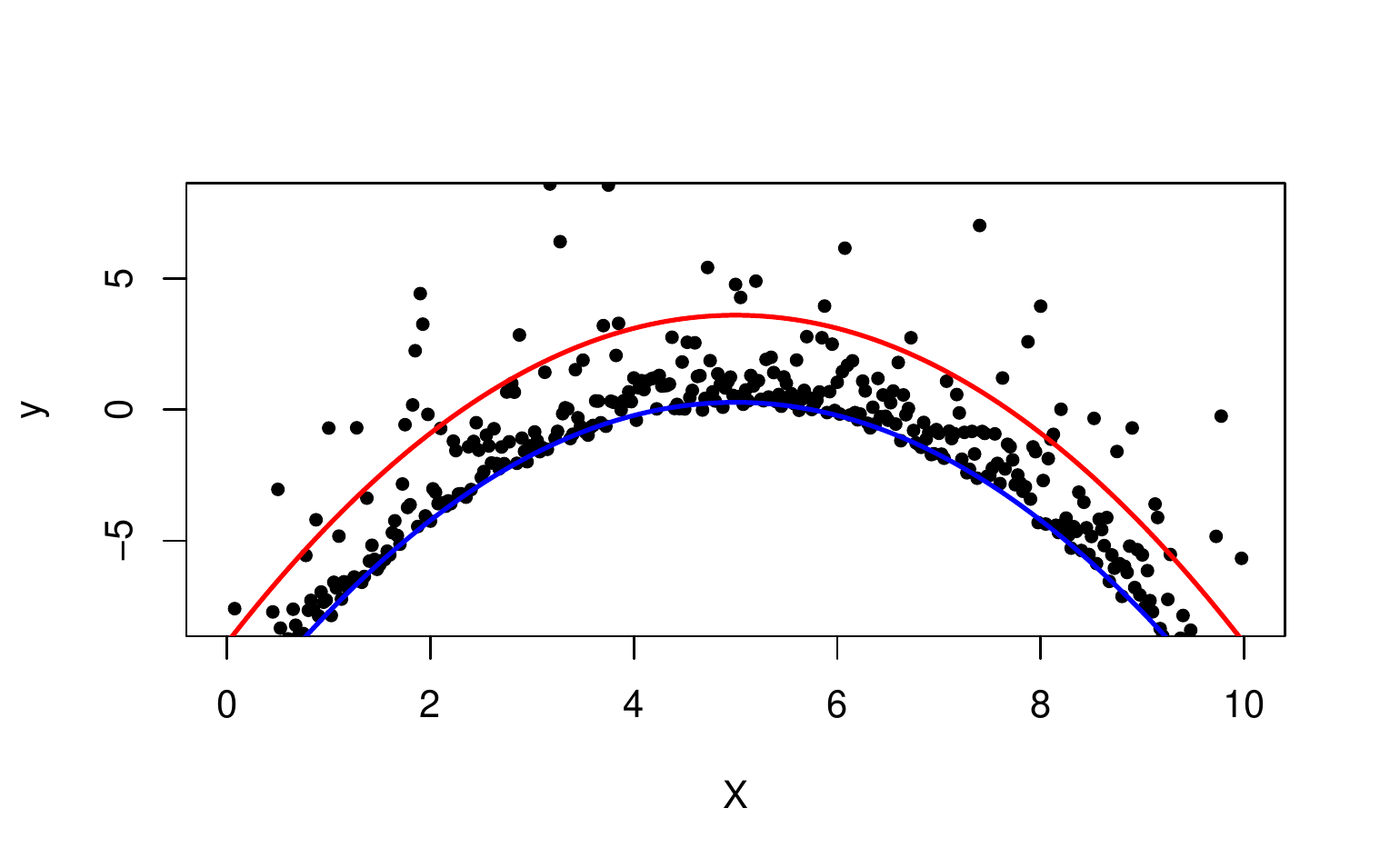}
				\end{tabular}
				\caption{Left: log-normal density fonction with $\mu=0$ and $\sigma=1$. Right: a sample generated by the function $\Psi(x,\cdot)=\xi(\cdot)-(x-5)^2/2$, with $\xi$ a random variable following a log-normal distribution of parameters $\mu=0$ and $\sigma=1$. 
					The $0.9$- (resp. the $0.1$-) quantile is represented in red (resp. in blue). One can notice that more information is available in areas with large pdf (i.e. for the $0.1$-quantile) than areas with small pdf.}\label{fig:pdf}
			\end{figure*}

			\paragraph{Factors.}
			Three factors are explicit in our benchmark system: the number of training points, the problem dimension and the quantile level. The other factors depend on the characteristics of the problem concerned: shape variation, pdf value at the quantile, level of heteroscedasticity, signal-to-noise ratio. For all four test cases, we consider three quantile levels: $0.1$, $0.5$ and $0.9$. Note that due to the asymmetry of the problems, learning for the $0.1$ and $0.9$ quantiles is not equivalent in terms of difficulty. Indeed, when the response is heteroscedastic (a variance/spread depending on $x$) and/or when the shape of $\mathds{P}_{x}$ varies in $x$, the pdf $f(x,q_\tau)$ may vary in $x$ as well. Intuitively, quantiles with large pdf values are easier to learn, as the data points may be closer to them. Figure \ref{fig:pdf} illustrates this effect.
			Table~\ref{tablegroup1} summarizes the characteristics of our design concerning the number of training points with respect to the dimension of the test case. To make our results easier to analyze, we divided the problems into groups that allow us to focus on subsets of factors.

			\paragraph{Focus 1: is there a universal winner?}
			To provide a universal ranking of the methods, we use all test cases, training points and quantile levels. As highlighted in Table~\ref{tablegroup1}, we created a set of different problems representative of a large number of characteristics that could be met dealing with any quantile regression problem. Note that our benchmark system is slightly biased towards small-dimensional problems, since only three-fifths of the cases have a dimension higher than two.
			
			\paragraph{Focus 2: what behavior with respect to the dimension, the number of training points, signal-to-noise ratio and pdf value?}
			To analyze the effects of these factors on the performance of the methods, we combine toy problems $1$, $2$, $3$, $4$ and the SUNFLO model. Note that once the pdf value is taking as a explanation variable, toy problem $1$ is excluded from the group because the pdf value near all the studied quantiles cannot be classified as large or small.
			
			\subsection{Performance evaluation and comparison metrics}
			Assessing the performance of quantile regression is not an easy task when only limited data are available. Here, since we are considering toy problems (exept for SUNFLO, in that case the true quantile values are taken as the quantiles of the $190$ years of weather records), the true quantile values can be approximated with precision, so we can evaluate the $L^2$ error for each emulator. The value of the criterion is privided by (\ref{l2}).
			
		 We chose $n_{\text{test}}=250$ for the $1d$ problems and $n_{\text{test}}=4000$ for the others.
			Now, since the problems vary in difficulty and in their response range (Figure \ref{toy_prob}), $E_{L^2}(\hat{q})$ cannot be aggregated directly over several problems or configurations. To do so, we normalize this error by the error obtained by a constant model (the constant being taken as the quantile of the sample):
			\begin{equation}
			E_{cq}(\hat{q})=\sqrt{\dfrac{E_{L2}(\hat{q})}{E_{L2}(\CQ)}}\times100,\label{eq:normL2err}
			\end{equation}
			where $\CQ$ stands for constant quantile.
			
			As an alternative criterion, we consider the ranks of the metamodels based on their $L^2$ error. Although ranks do not provide information regarding the range of errors, they are insensitive to scaling issues, which makes aggregation between configurations more sensible. They allow us to assess whereas any method consistently outperforms others, regardless of overall performance.

			\section{Results}\label{sec:results}	
			\subsection{Focus 1: overall performance and ranks}
			
\begin{figure*}[!ht]
				\begin{tabular}{cc}
					\includegraphics[width=.48\textwidth]{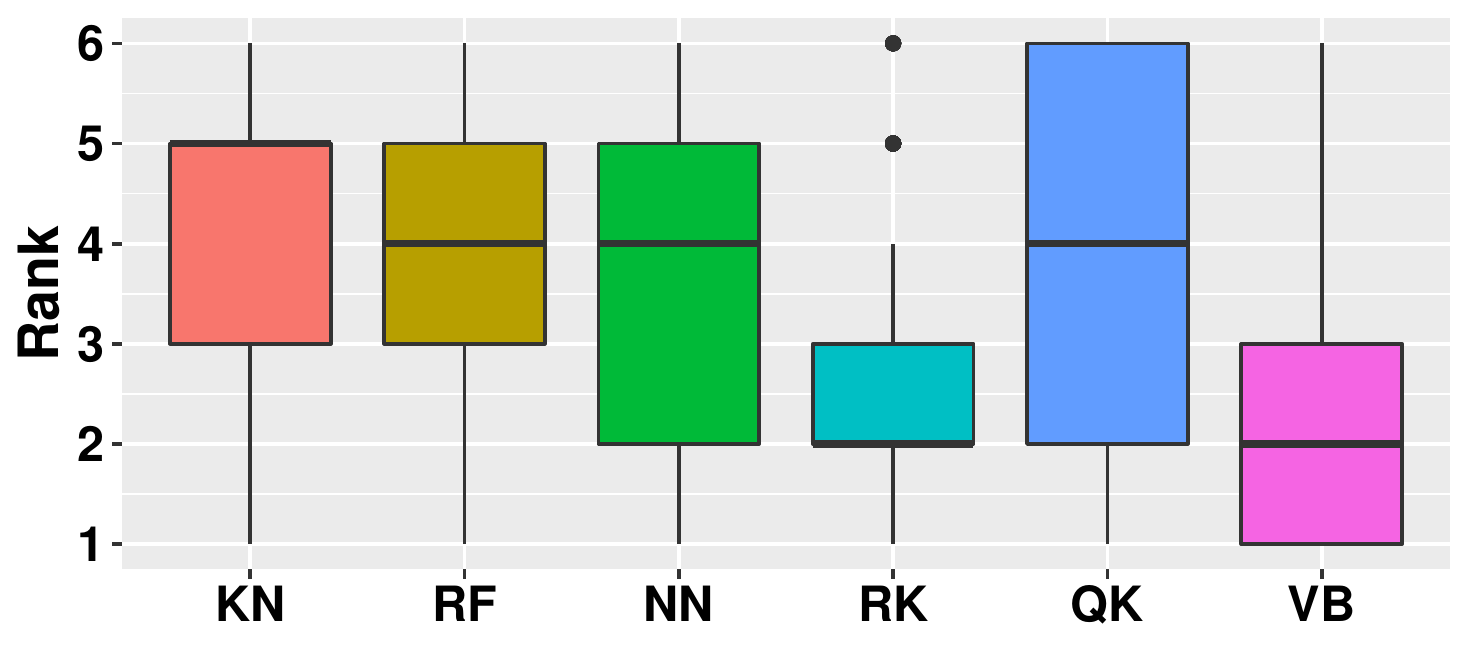}
					\includegraphics[width=.48\textwidth]{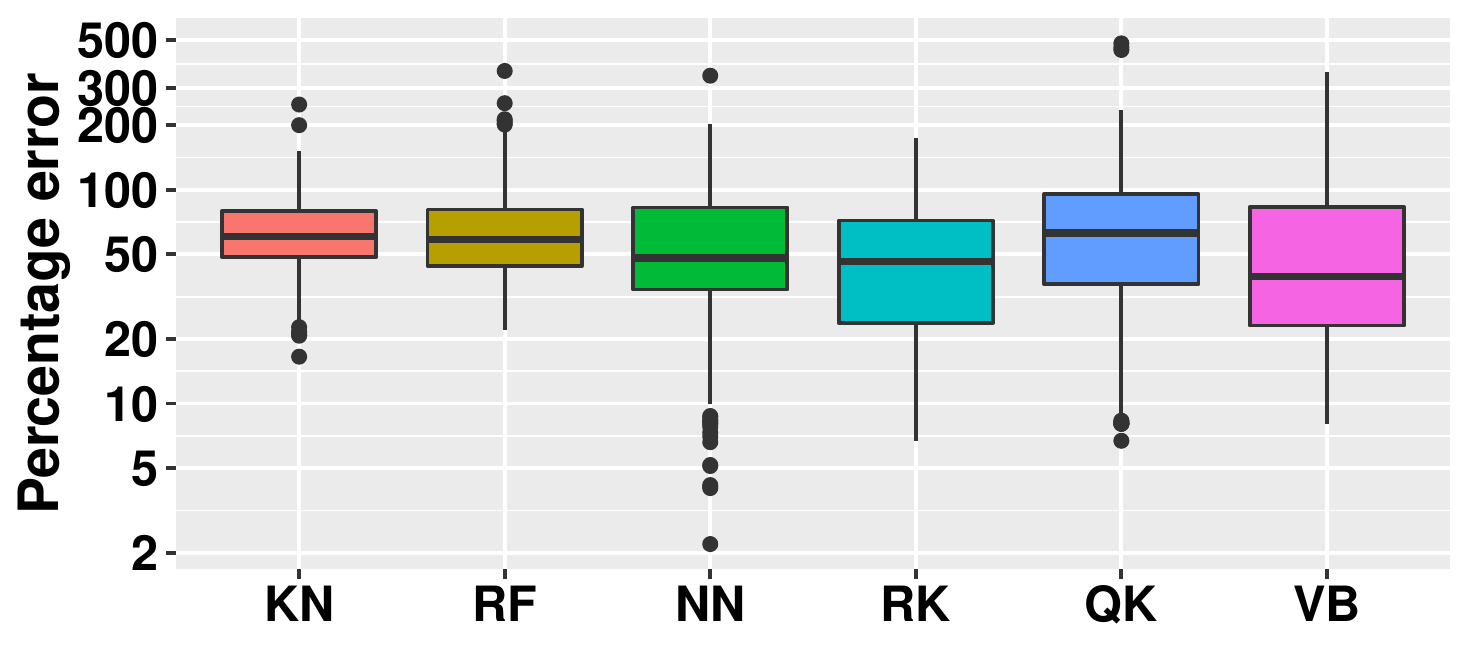}
				\end{tabular}
				\caption{Boxplots of ranks and $E_{cq}$ error over the entire benchmark. Note that for clarity, the right boxplots do not contain the error of the median for the toy problem $2$.}\label{hist}
			\end{figure*}

			First, we consider the overall performance and ranks, integrated over all runs. We have considered $5$ test cases, for each test case we have generated $4$ sizes of training sets, for each test case and sample size 10 samples are drawn and for each of this occurrences we have estimated $3$ different conditional quantiles. Thus, we have $5\times 4 \times 10 \times 3 =600$ experiments. They are shown as boxplots in Figure \ref{hist}. Based on these ranks (Figure \ref{hist}, left), VB appears to be the best solution since it is ranked either 1st or 2nd in 50\% of the problems. RK is in second position, its median is the same as VB but it is generally ranked between 2nd an 3rd. In addition RK seems slightly less risky than VB in the sens that it is almost never rank 5th or 6th. KN is the worst since its median rank is equal to five. 
			However, all boxplots range between 1 and 6, indicating that no method is outperformed by another on all problems. This finding is reinforced by the performance boxplots (Figure \ref{hist}, right), where all median performances are similar (VB and RK being the best and QK the worst which means QK may be very bad sometimes), and the variance is very large. Indeed, the errors range from 2\% for NN (of the error achieved by a constant metamodel) to 500\% for QK, all methods experiencing cases with more than 100\% error (i.e. situations where they are worse than the constant metamodel).
			\subsection{Focus 2: dimension, number of training points and pdf value}
			
			\subsubsection{Performance according to the constant quantile}
			In this section, we analyze the performance of the methods with respect to the pdf value and the number of points.
			
\begin{figure*}[!ht]
				\centering
				\includegraphics[width=1\textwidth]{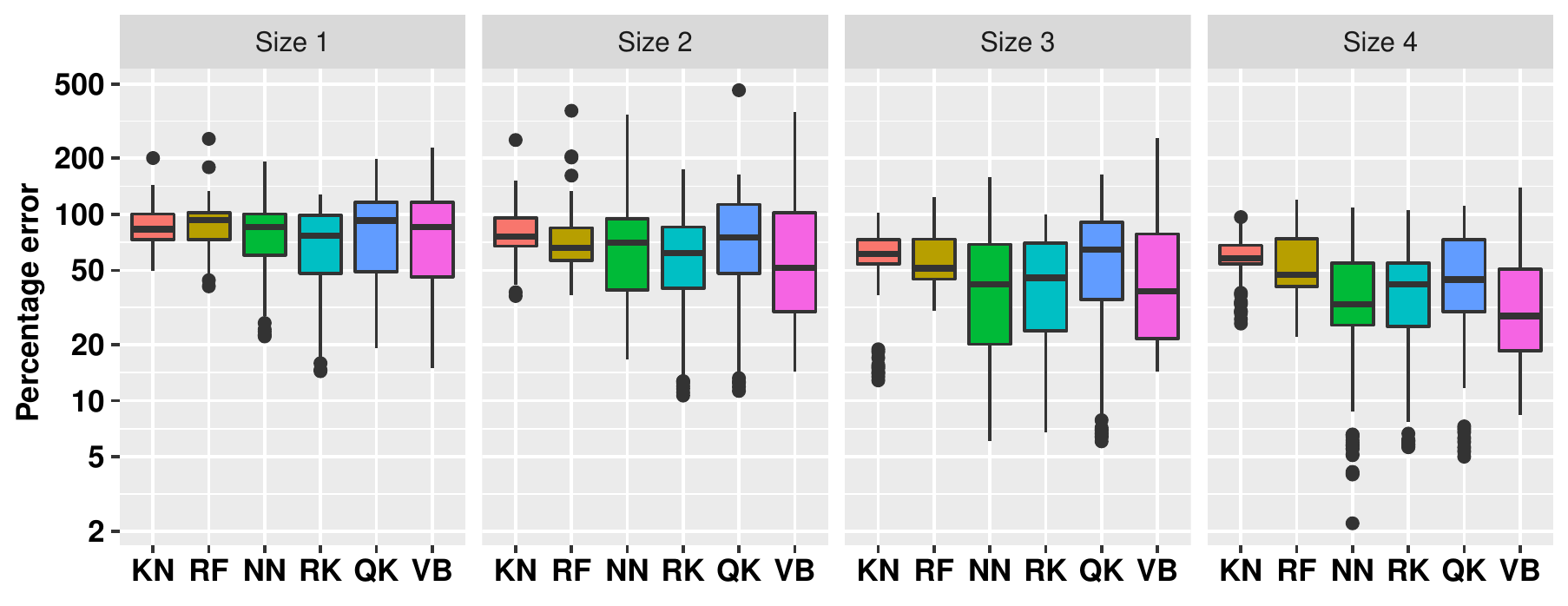}
				\caption{Error according to the sample size}\label{sample_size}
			\end{figure*}

			\paragraph{Sample size:}
			Figure \ref{sample_size} shows the performances of the methods grouped according to the size of the sample. As expected, the performances increase with the size of the sample. For size $1$ ($n\approx50d$), the distribution of $E_{\CQ}$ of all the metamodels is almost centered around 100\%, implying that these correspond to limit cases for quantile regression since the metamodels do not outperform the constant metamodel (although in some cases the error is as small as 40\%). For size $4$ ($n\approx300d$), the median performance is roughly 50\% (twice as accurate as the constant metamodel). BV, RK and especially NN experience situations with very accurate models. However, all the methods also experience bad performances (error greater than 100\%) in the large sample regime. Unfortunately, from Figure \ref{sample_size} we can conclude that no method is sufficiently robust in all cases.\\
			
\begin{figure*}[!ht]
				\centering
				\includegraphics[width=0.9\textwidth]{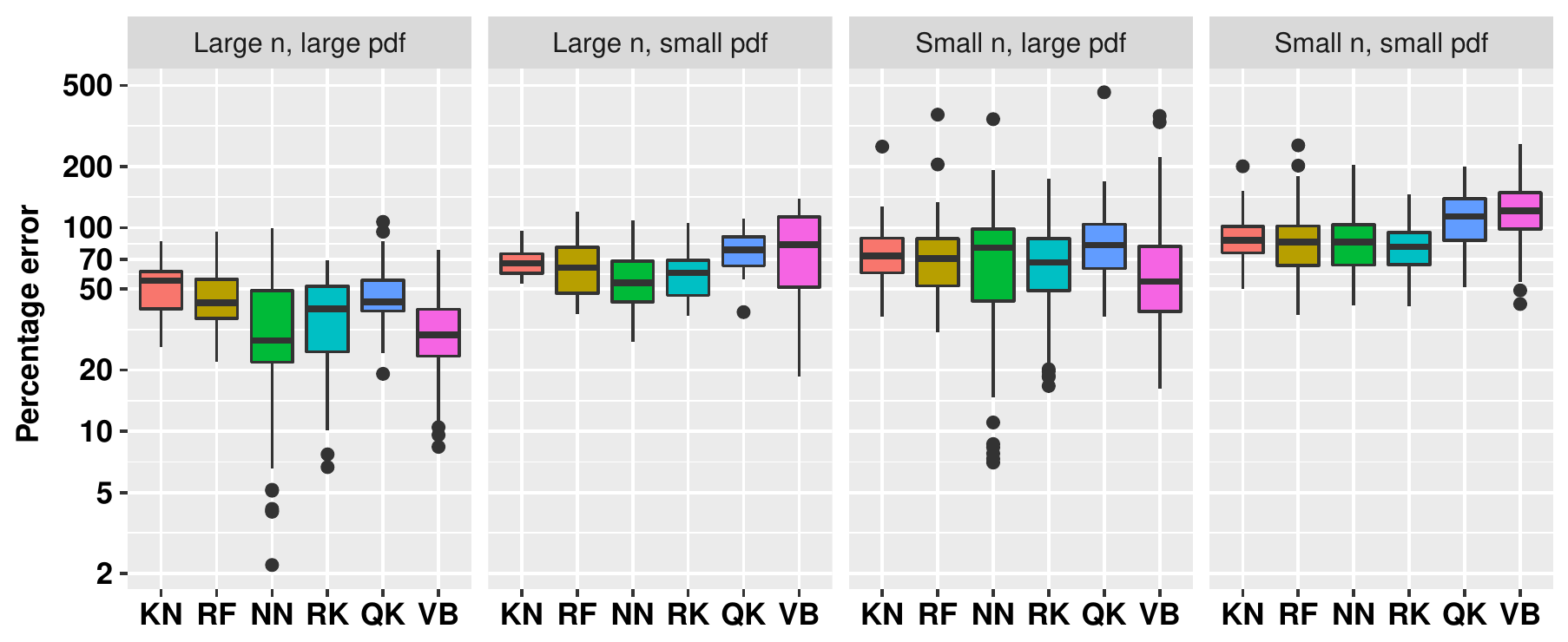}
				\caption{Error according to the size of the training set and the pdf value.}\label{pdf_s}
			\end{figure*}

			\paragraph{Signal-to-noise ratio.} Figure \ref{pdf_s} groups performance with respect to the signal-to-noise ratio. According to the figure the performance depends to a great extent on the signal-to-noise ratio. Dealing with a high signal and considering the performance, there is a clear difference between the statistical methods (KN, RF) and the four others (RK, NN, VB, QK) while this difference is not visible in the small signal setting. The impact of the signal-to-noise ratio is strong on the performance. Dealing with a high signal, the median of the performance for NN, RK, QK, VB is close to 20\% with a small variance while if the signal is small, the error is larger (the median is above 50\% for all methods) and the variance is larger. 
			
			In the following we focus our interest only on cases with small signal-to-noise ratio. Indeed as the performance of RK, NN, QK and VB are close to each others, we think we do not have enough experiments to extract patterns.\\

\begin{figure}[!ht]
				\centering
				\includegraphics[width=0.6\textwidth]{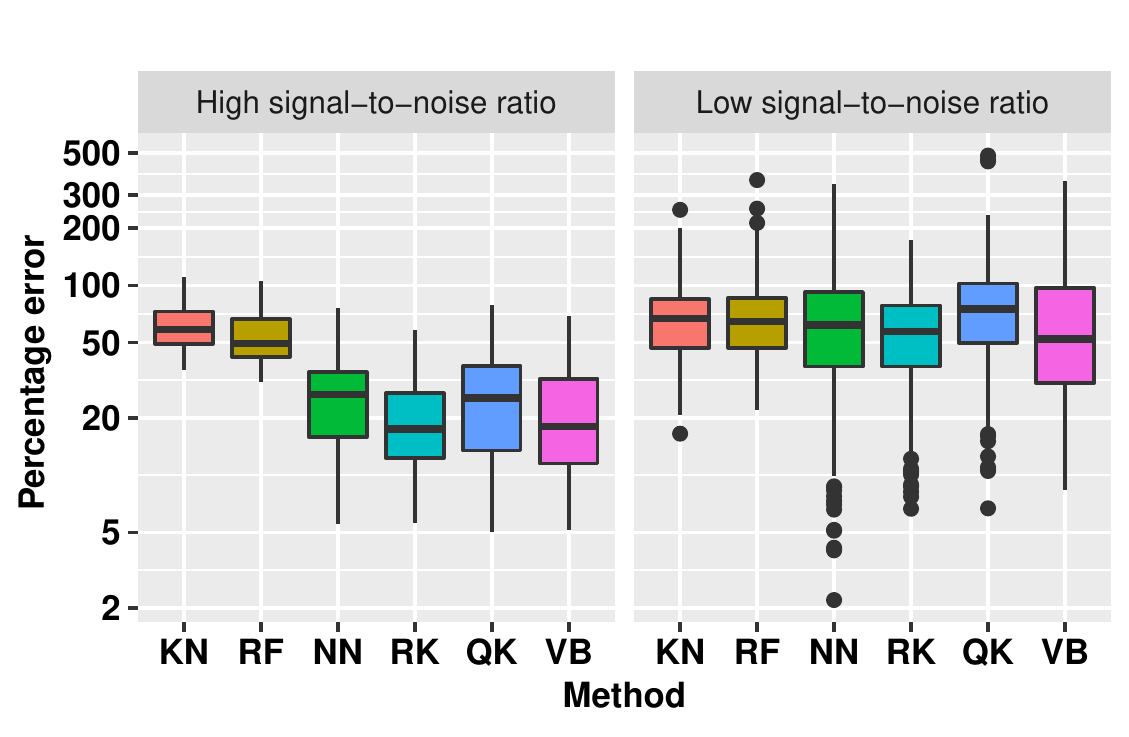}
				\caption{Error according to the signal-to-noise ratio.}\label{signal}
			\end{figure}

			\paragraph{Pdf value with small signal-to-noise ratio.} Figure \ref{pdf_s} groups performance with respect to sample size (either small, i.e. level 1 and 2 or large, i.e. level 3 and 4) and pdf value (according to Table \ref{tablegroup1}). According to the Figure \ref{signal}, the performance depends to a great extent on the pdf value in the neighborhood of the targeted quantile.  With a small $n$, the pdf value has no significant impact on the median of the performance (except for VB) but it does have an impact on the lower bound of the error. More precisely, the median of the error does not depend on the pdf value in the case of a small pdf but sometimes the metamodel errors are sensibly smaller when the pdf is large. With large samples, both the median and the lower bound of the error depend on the pdf value. Metamodels may be very good when the pdf is large, for example $20$ times better than the constant metamodel for NN whereas the error appears to have a lower bound when the pdf is small even with large  $n$. In addition, for a problem with a small $n$ and a large pdf, the performance is similar to the performance for problems with a large $n$ and a small pdf (Figure \ref{pdf_s}, see the two columns in the center).
			
			\subsubsection{Rank in the context of a low signal-to-noise ratio}	
\begin{figure}[!ht]
				\centering
				\includegraphics[width=.6\textwidth]{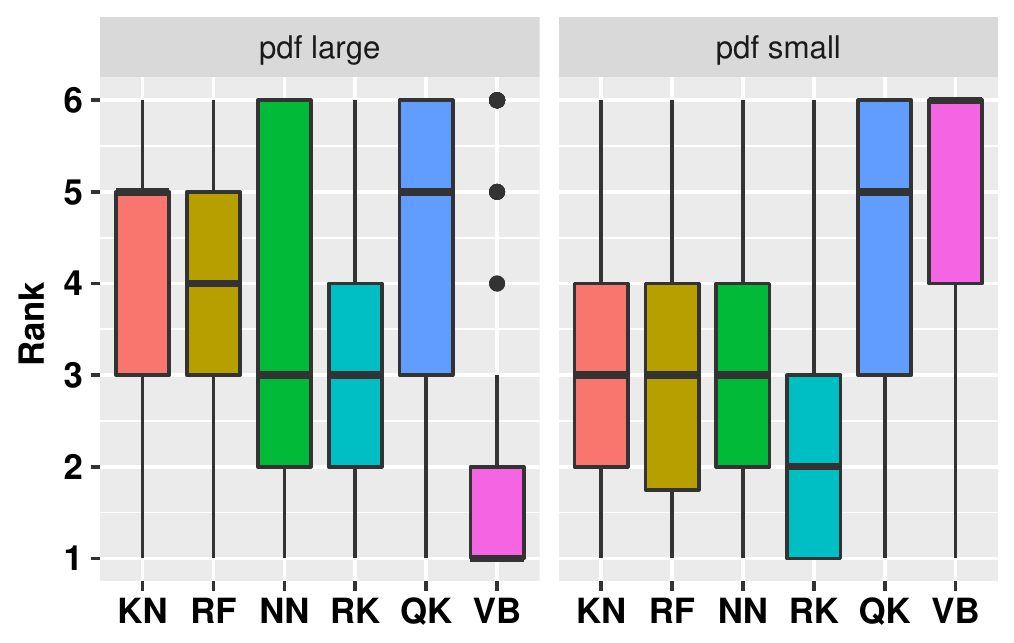}
				\caption{Rank according to the pdf value}\label{pdf}
			\end{figure}

			\paragraph{Pdf value.} Figure \ref{pdf} shows clearly that when the pdf is large, VB is the best model while when the pdf is small (and the problem is heteroscedastic), VB is less good than RN, RK and KN. This observation is supported by Figure \ref{pdf_s} which reveals a strong contrast between the performance of the VB method. QK is poor in both cases, whereas RK performs comparatively better with small pdf.\\
			
\begin{figure*}[!ht]
				\centering
				\includegraphics[width=0.6\textwidth]{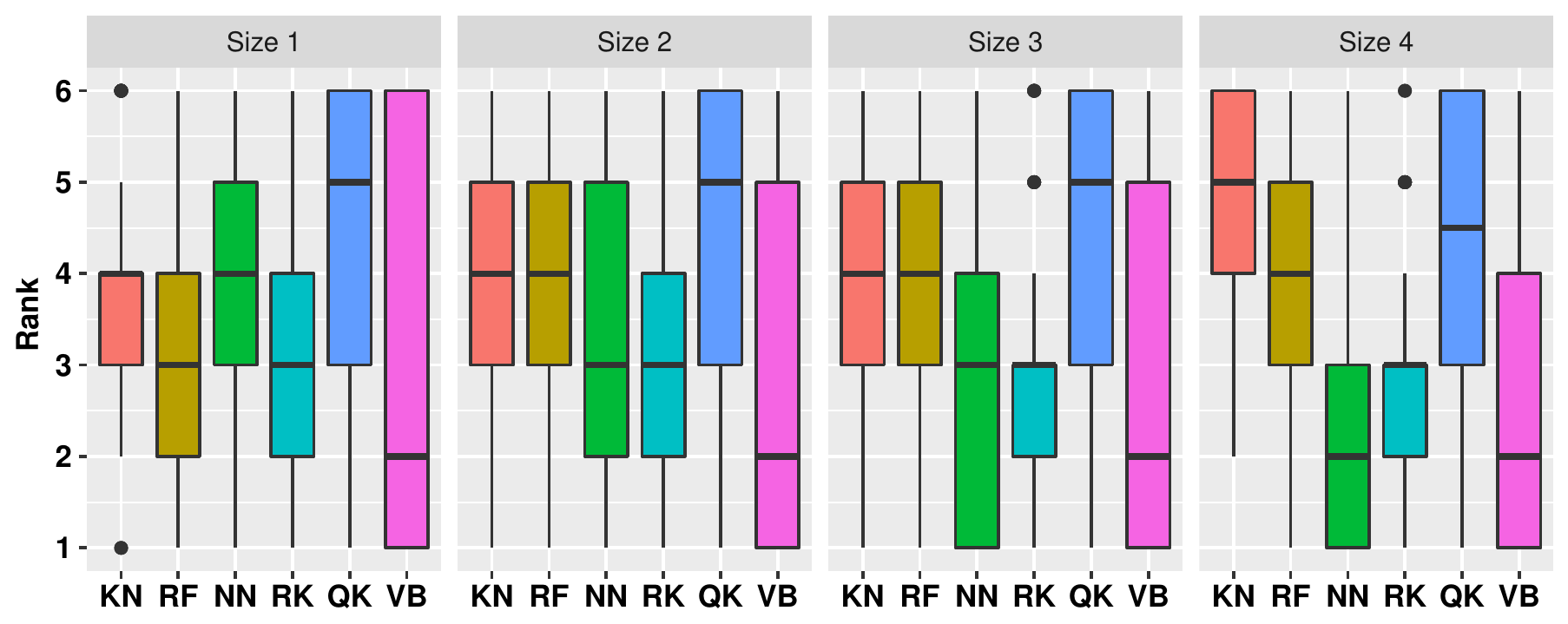}
				\caption{Rank according to the size of the sample}\label{number}
			\end{figure*}

			\paragraph{Sample size.} Figure \ref{number} shows that the number of points has a major impact on the ranking of some methods. The ranking of QK and VB is relatively insensitive to the size of the sample. The other methods are less distinguishable when the sample size is small than when it is large. With a small sample KN, RF, NN and RK are comparable, whereas when the sample size increases, NN and RK clearly outperform KN and RF. For the largest size, NN is slightly better than VB.\\
			
\begin{figure}[!ht]
				\centering
				\includegraphics[width=.4\textwidth]{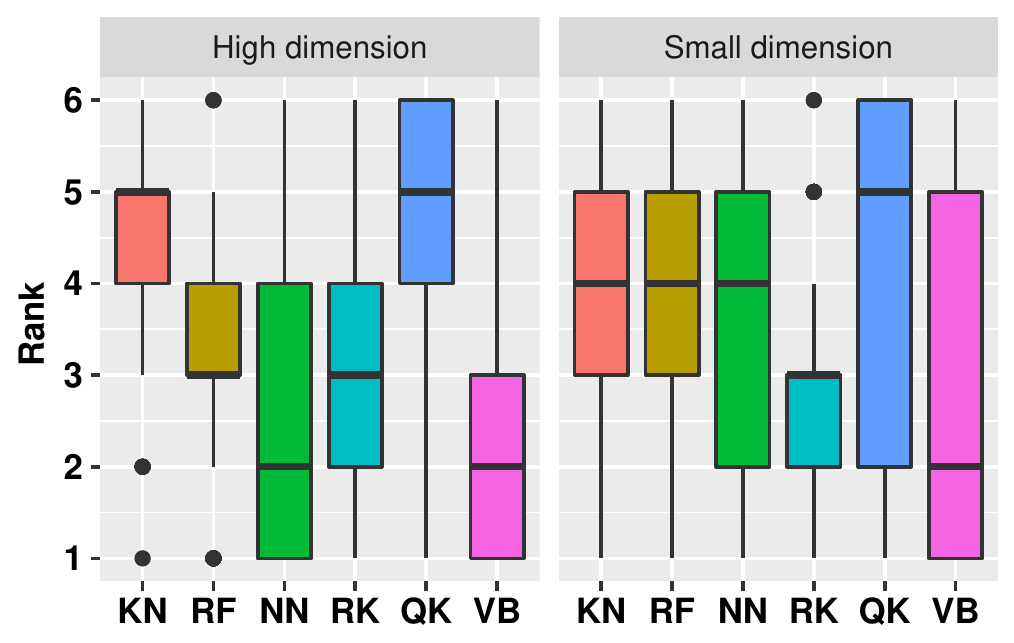}
				\caption{Rank according to the dimension}\label{dim}
			\end{figure}

			\paragraph{Dimension.} Figure \ref{dim} groups performance based on dimension. The first contrast is the permutation between RK and NN. With a small dimension, RK is better than NN but the relative performance of NN increases w.r.t. the dimension. With small dimensions, RF and KN are comparable, but with high dimensions, RF outperforms KN.\\
			
\begin{figure}[!ht]
				\centering
				\includegraphics[width=.4\textwidth]{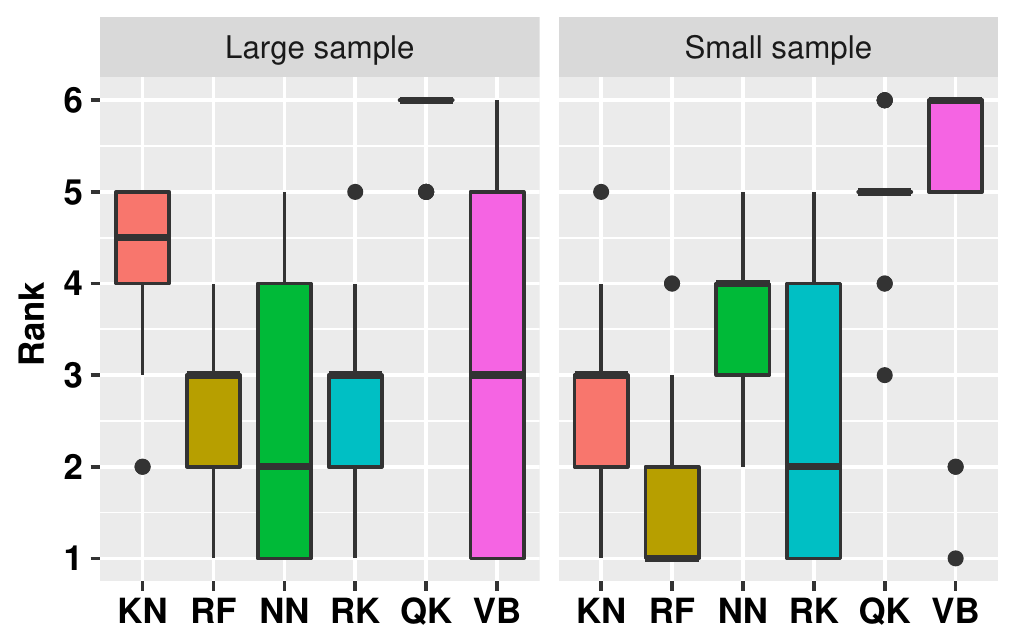}
				\caption{Rank associated to the case where little information is available, $i.e$ high dimension and small pdf}\label{extreme}
			\end{figure}

			\paragraph{High dimension, small pdf.} Figure \ref{extreme} shows an extreme case in which the pdf is small but the dimension is high. With a small $n$, the best method is clearly RF followed by RK and KN. VB and QK are not well ranked. With a larger $n$, as mentioned above, NN and VB are better but with large variance, while RF, RK and KN rank less well.
			
			
			
	
			\section{Extensions and open questions}\label{sec:discussion}

			\subsection{Effect of hyperparameter tuning}
\begin{figure*}[!ht]
				\begin{tabular}{cc}
					\includegraphics[trim={0 0.7cm 2cm 0},clip,width=0.4\textwidth]{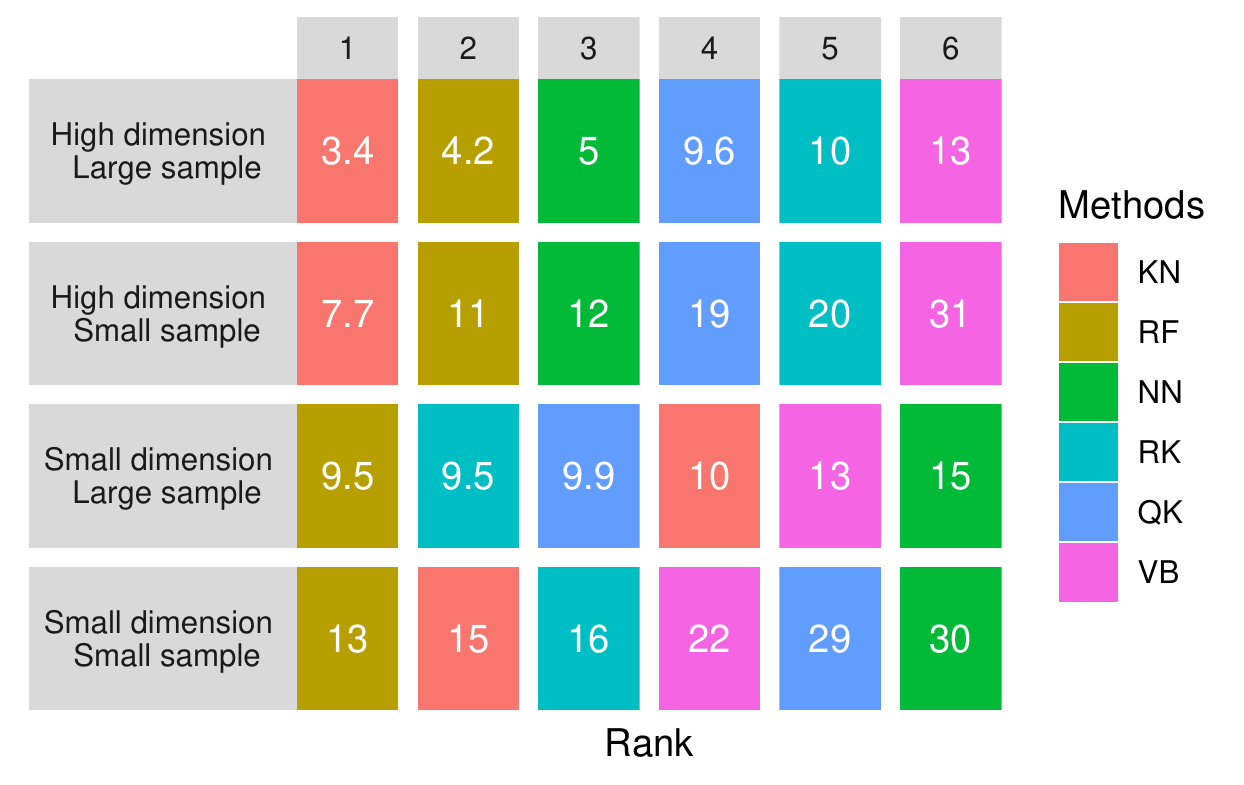}
					\includegraphics[trim={0 0.7cm 2cm 0},clip,width=.4\textwidth]{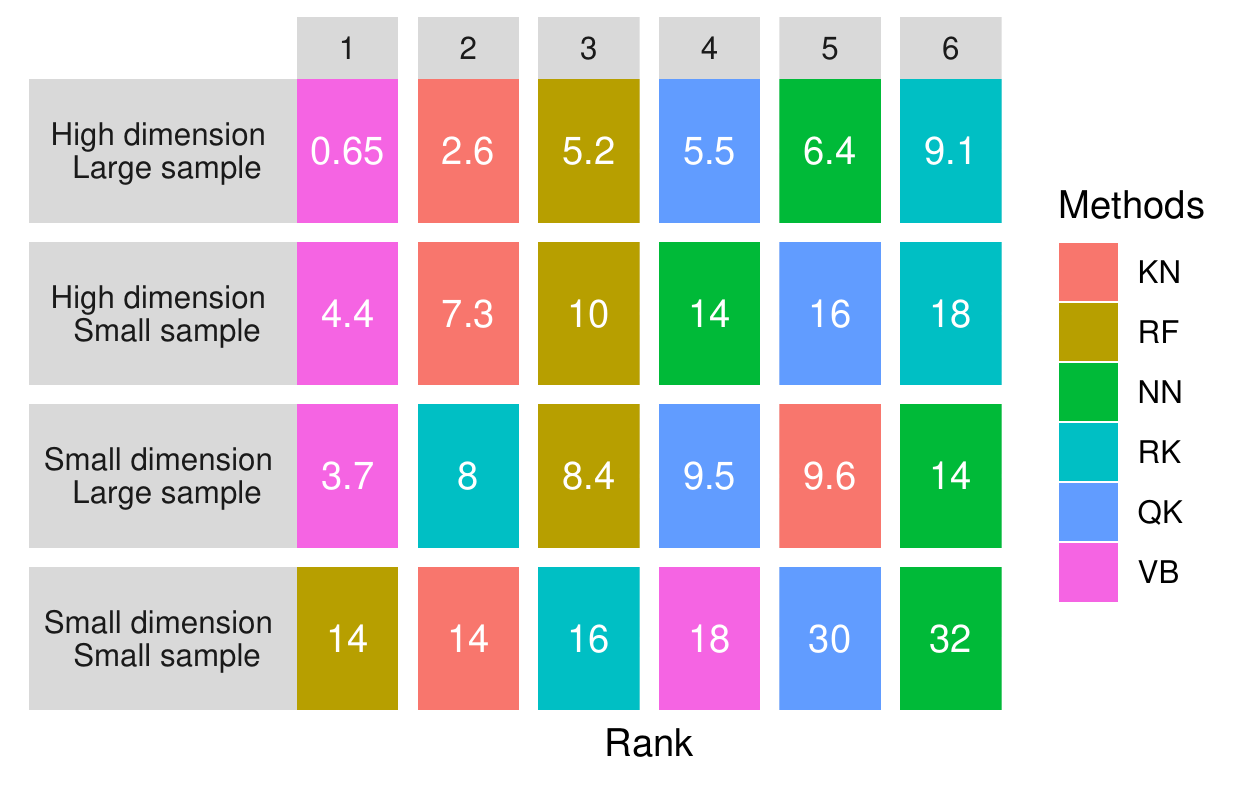}
					\includegraphics[trim={10.5cm 0 0 1.5cm},clip,width=.09\textwidth]{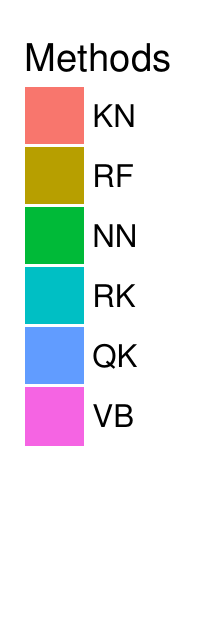}
				\end{tabular}
				\caption{Average $\Delta E$ aggregated over all problems (left) and over the problems with a large pdf only (right),
					arranged in increasing order. For each method the rank is provide on the top of each figure.}\label{delta}
			\end{figure*}

			In the following we define
			$$\Delta E(\hat{q}_{\tau}^{\Theta})= E_{cq}(\hat{q}_{\tau}^{\Theta})-E_{cq}(\hat{q}_{\tau}^{\Theta^*}),$$
			the performance gap between the regular metamodel and its oracle performance (the loss in performance between actual hyperparameter tuning and oracle tuning).
			Figure \ref{delta} gives the average values of $\Delta E$  aggregated respectively over all problems and only aggregated over the problems with a large pdf, and considering the effect of dimension and sample size.  In high dimension, the easiest methods to tune are KN, NN, and RF. In our study, KN and RF have a single hyperparameter to tune regardless of the dimension, and NN has two. This is clearly an advantage in terms of robustness in high dimension. The other methods are kernel-based and require the tuning of at least $d+1$ hyperparameters. This consistently affects RK and QK, but affects VB only in the case of small pdf, while it is the most stable method in the other cases.
			
			With a small dimension, all the methods have roughly the same number of hyperparameters. The most noticeable change compared to the case of a high dimension is the good performance of RK, while NN becomes comparatively the most difficult method to train.
			
			\subsection{On the methods’ behavior}
			
			\paragraph{Statistical order methods.}
			As presented on Figure \ref{signal} this methods are not relevant on high signal-to-noise regime. A possible explanation is the difficulty to fit smooth variations with high amplitudes with a piecewise constant model. 
			
			Concerning the low signal-to-noise regime, it is clear from Figure \ref{dim} that KN performs poorly in a high dimension. This may be due to the irrelevance of the Euclidean distance when there is a significant increase in dimension. RF clearly outperforms KN in this situation, as it is able to produce better neighborhoods than the Euclidean distance. Overall, (compared with the other methods), RF performance increases with dimension. This may be due to the fact that it has fewer hyperparameters to tune.\\
			
			\paragraph{Functional methods.}
			As presented on Figure \ref{signal} this methods are relevant on high signal-to-noise regime.
			
		Concerning the low signal-to-noise regime, Figure \ref{number} shows that NN works poorly in a small sample setting, but it is one of the best methods when the sample is large. This result reflects the high flexibility of NN. Too much flexibility leads to overfitting when the sample is small. In contrast, when the number of points is large, NNs are able to fit the data very well (e.g. Figure \ref{sample_size}, Size $4$). According to Figures \ref{sample_size} and \ref{number}, RK is a robust method. Its robustness in both small and large data settings can be attributed in part to the selected kernel. If the selected kernel is sufficiently smooth (here continuous and derivable), the resulting metamodel cannot produce instable results. However, it seems (Figure \ref{sample_size}, Size $3$ and $4$) that this lack of flexibility may affect the performance with an increase in the size of the data set. In this case, more flexible methods like NN may outperform RK. The contrast between RK and NN shown in Figure \ref{dim} can be explained by the level of difficulty  associated with each method involved in finding good hyperparameters (as explained above).\\
			
			\paragraph{Bayesian models.}
			As presented on Figure \ref{signal} this methods are relevant on high signal-to-noise regime.
			
			Dealing with low signal-to-noise ratio the QK method under-performs comparing to others. One possible explanation is the erroneous assumption in Equation (\ref{gaus}) that the noise is centered, which is more critical for extreme quantiles. Another possible explanation lies in the small number of replica. The local inference (that uses statistical orders) is biased and in a low signal-to-noise regime it has high variance. 
			In addition, the increasingly bad performance of QK with an increase in dimension (Figure \ref{sample_size}) is likely a consequence of the fact that empty areas become larger in high dimensions.
			
			VB is one of the best methods presented in our paper. Figures \ref{pdf} and \ref{delta} show that VB is also the most dependent on the pdf value. When the pdf is large, it is the best method whereas when the pdf is small (and the shape of the distribution and/or the variance depend on the input), it may be the worst. The explanation lies in the philosophy of the model.
			In the  case of RN and RK, the model complexity (i.e. smoothness) is almost entirely related to the regularity of parameter $\lambda$ that is selected by cross-validation. Hence, the model cannot excessively overfit and cannot perform very poorly. With Bayesian methods, the regularization is included in the model hypothesis: in our setting, the quantile is assumed to be a Gaussian process with covariance function $k_{\theta}(.,.)$, so $\theta$ performs the regularization. We observed that if the local quantity of information (roughly the product of the number of points times the pdf value in the neighborhood of the quantile) is too small comparing to the information needed to fit well the quantile, the metamodel tends to interpolate the available data.
			When sufficient information is available, the optimization of the marginal likelihood provides $\theta$ values that allow a good trade-off between flexibility and smoothness. This is likely the reason why VB is easily beaten by RN and RK when the pdf is small.\\
			
			\subsection{Varying shape and heteroscedasticity.}
\begin{figure*}[!ht]
				\begin{tabular}{ccc}
					\includegraphics[width=.33\textwidth]{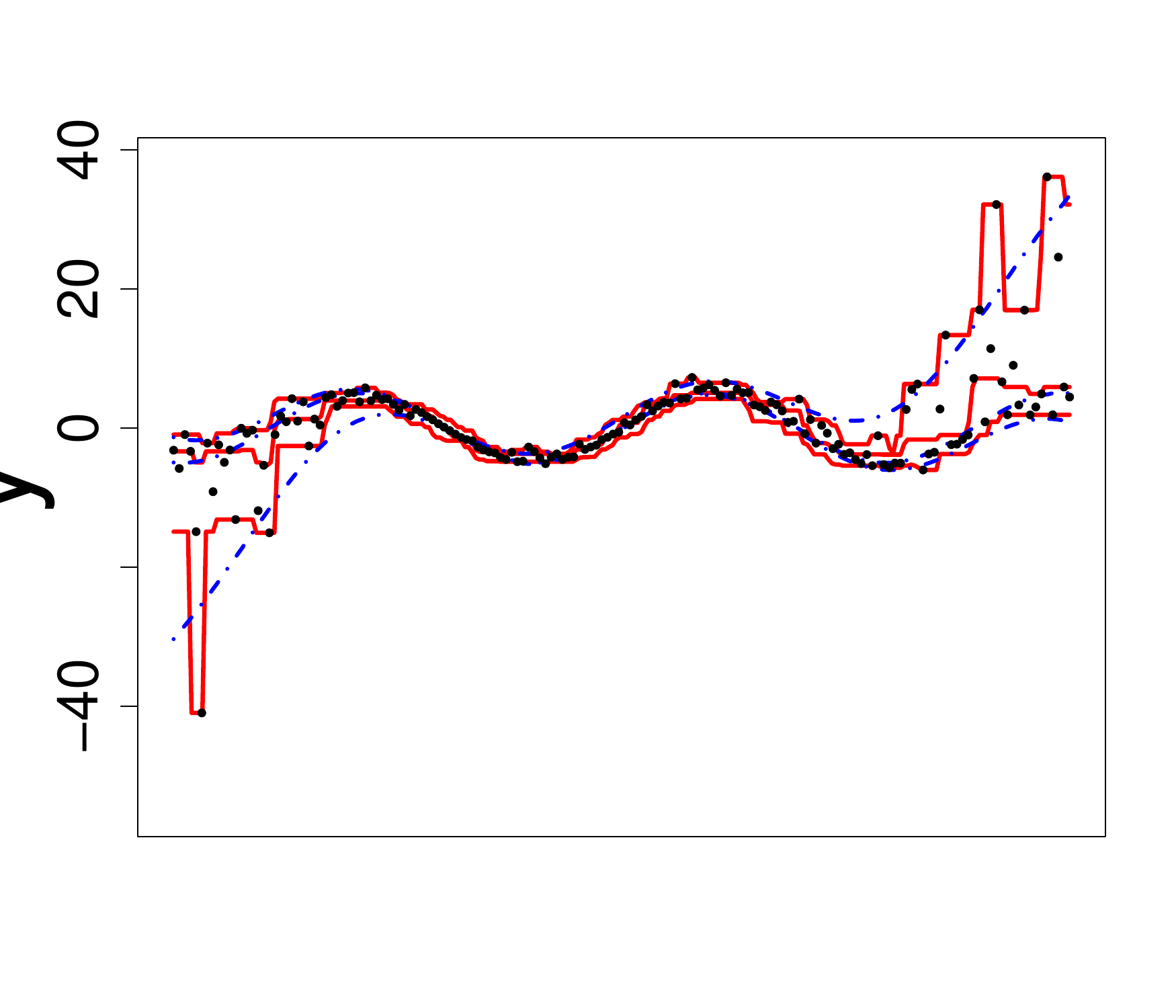}
					\includegraphics[width=.33\textwidth]{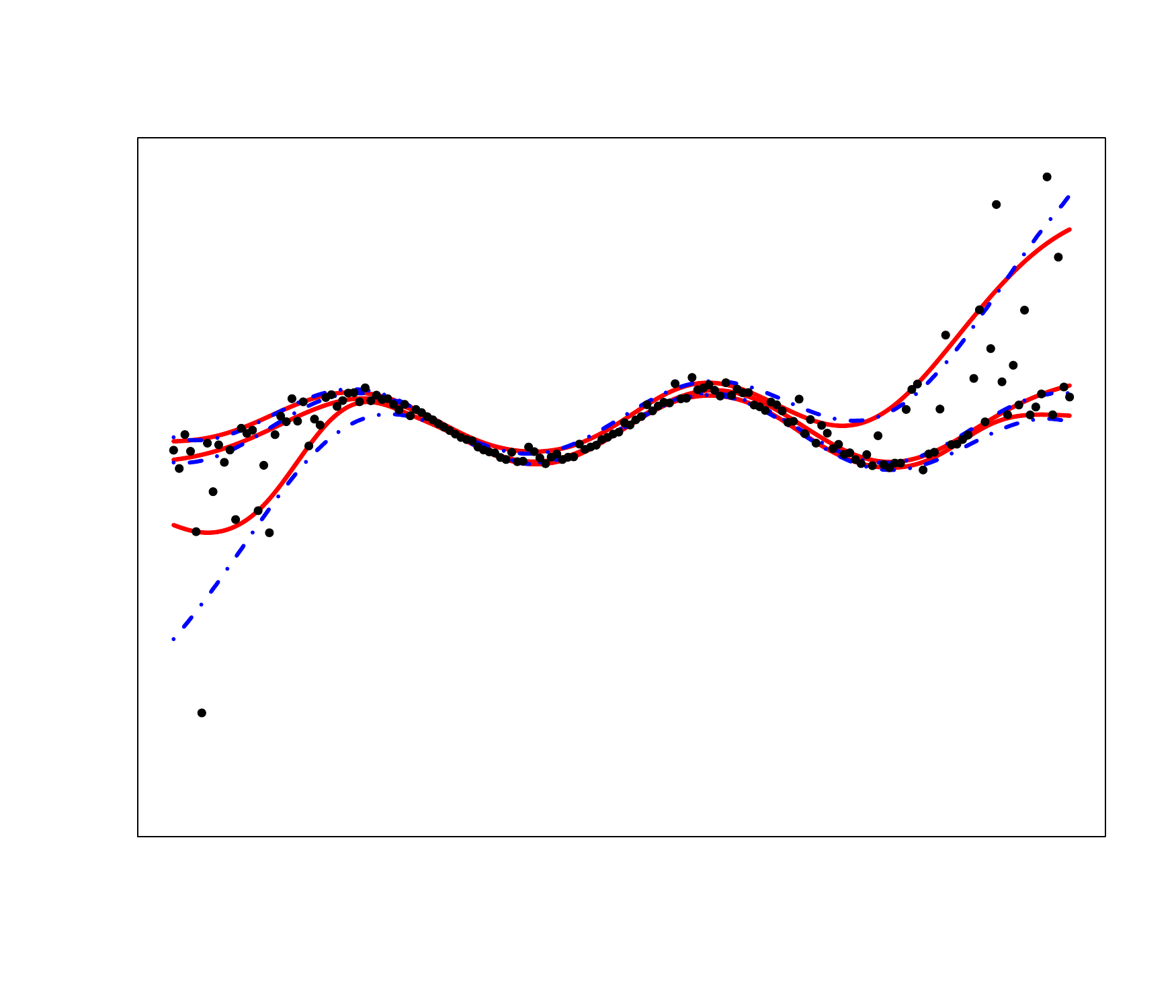}
					\includegraphics[width=.33\textwidth]{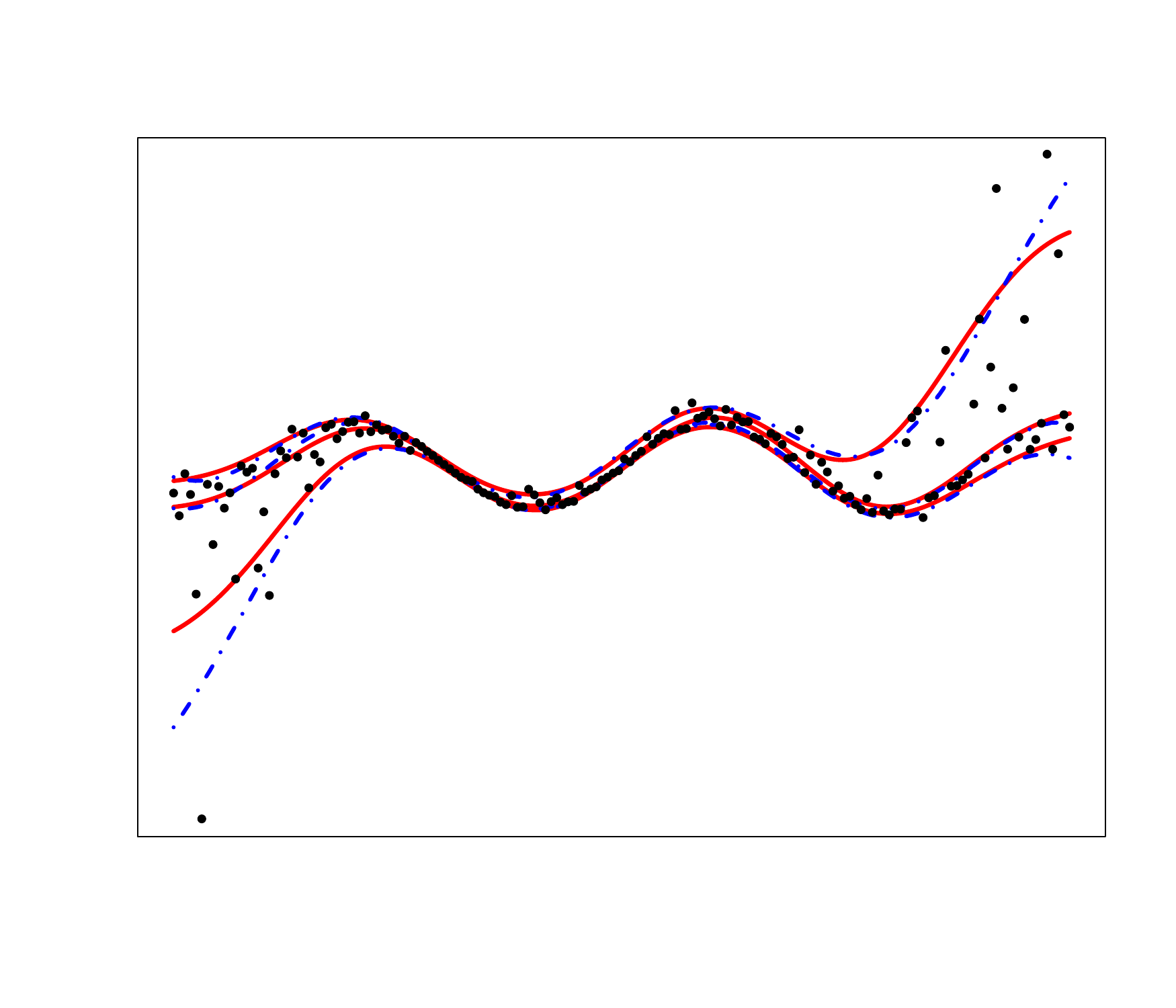}\\
					\includegraphics[width=.33\textwidth]{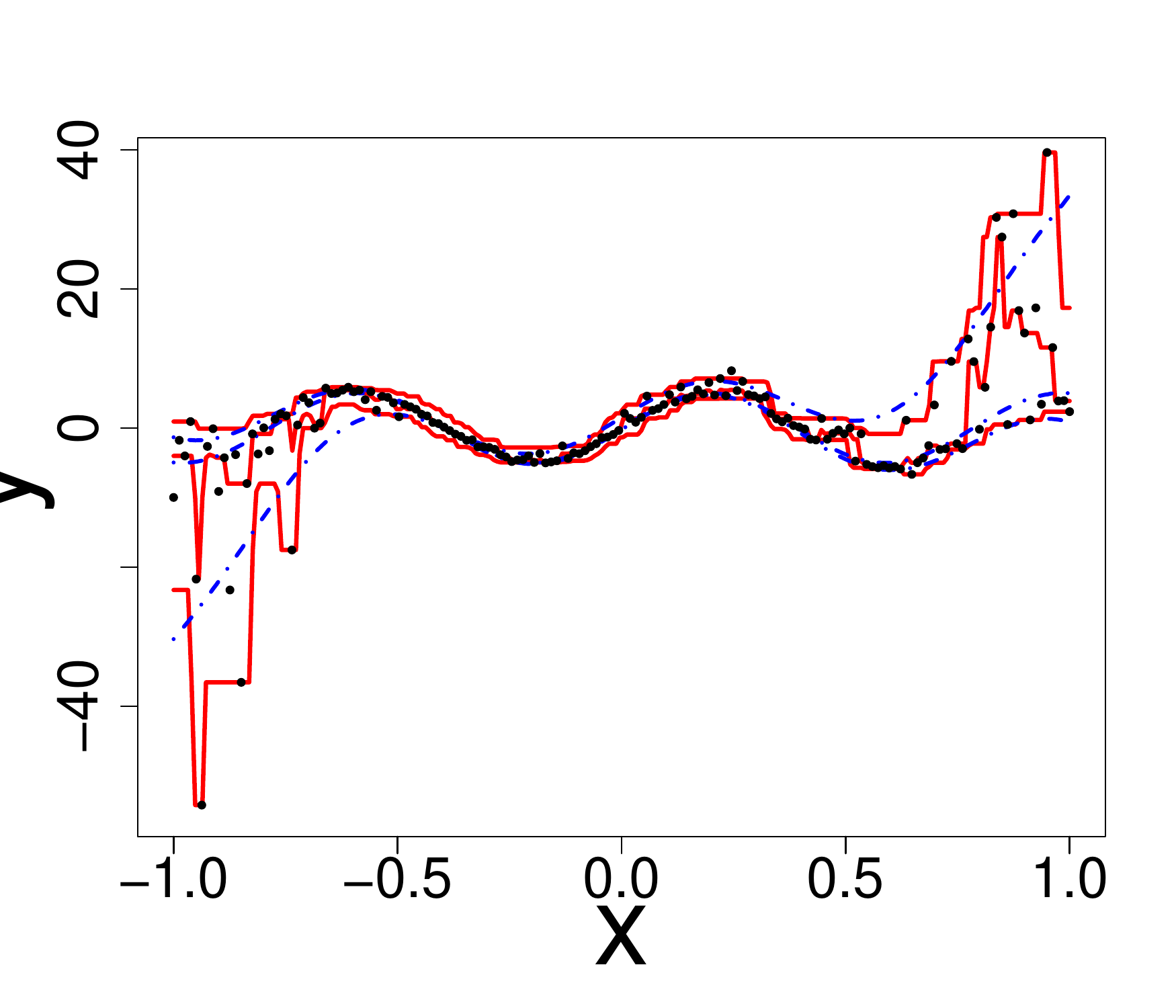}
					\includegraphics[width=.33\textwidth]{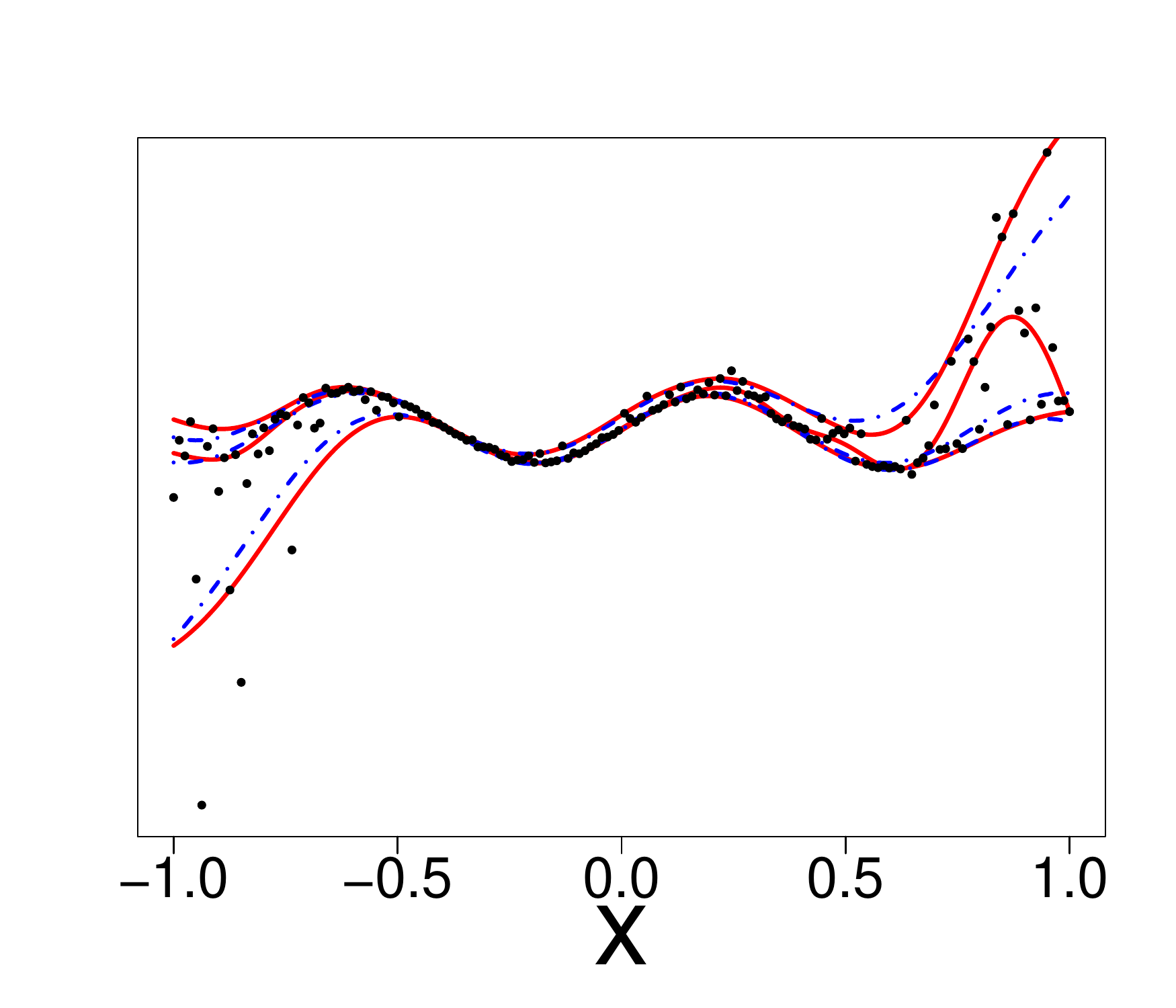}
					\includegraphics[width=.33\textwidth]{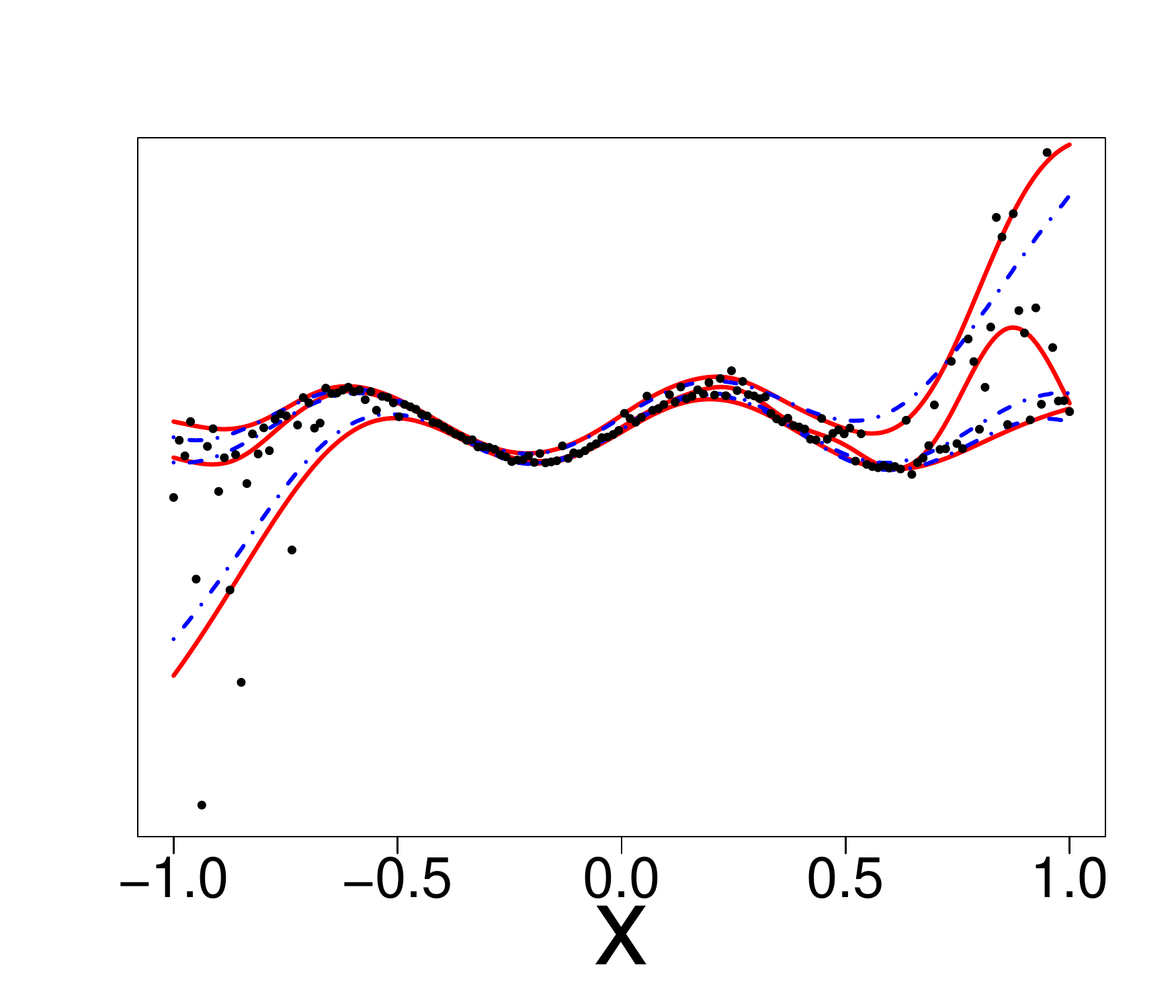}
				\end{tabular}
				\caption{Quantiles estimates using RF (left), RK (middle), VB (right) for two $160$-point samples (top and bottom rows, resp.) of the toy problem $1$.
					Dots: observations; plain red lines: metamodels for the $0.1,0.5,0.9$ quantile estimates; dotted blue lines: actual quantiles.}\label{Shape}
			\end{figure*}

			If the shape of $\mathds{P}_x(Y)$ or the variance of $Y_x$ (heteroscedasticity) vary w.r.t. $x$, then $f(x,q_\tau)$ may vary in $x$.
			Figure \ref{Shape} illustrates the ability of RF, RK and VB to estimate quantiles of a distribution with a strongly varying shape. In this problem, as depicted in Figure \ref{Shape} (top row), the quantiles are not perfectly estimated but the metamodels provide good indications about the shape of the true distribution. However, as can be seen in Figure \ref{Shape} (bottom row), the methods can present strong instabilities. Here, for a sample virtually indistinguishable from the one leading to accurate estimates, the median estimates largely overestimate the true values for large $x$ values. Such instabilities can be partly imputed to the difficulty of the task. However, this is also because no method is actually designed to deal with strongly varying pdf, as we explain below.
			
\begin{figure*}[!ht]
				\begin{tabular}{ccc}
					\includegraphics[width=.33\textwidth]{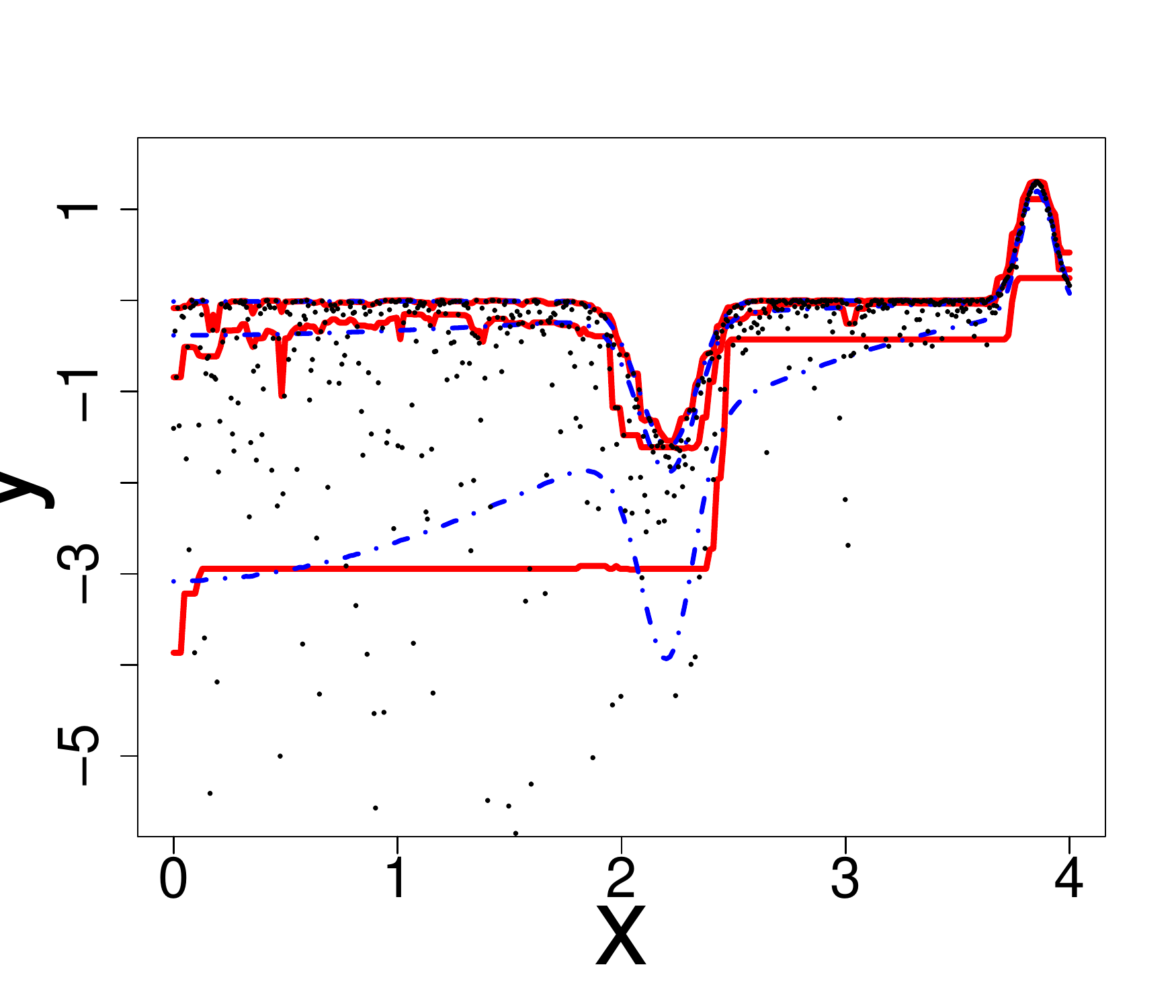}
					\includegraphics[width=.33\textwidth]{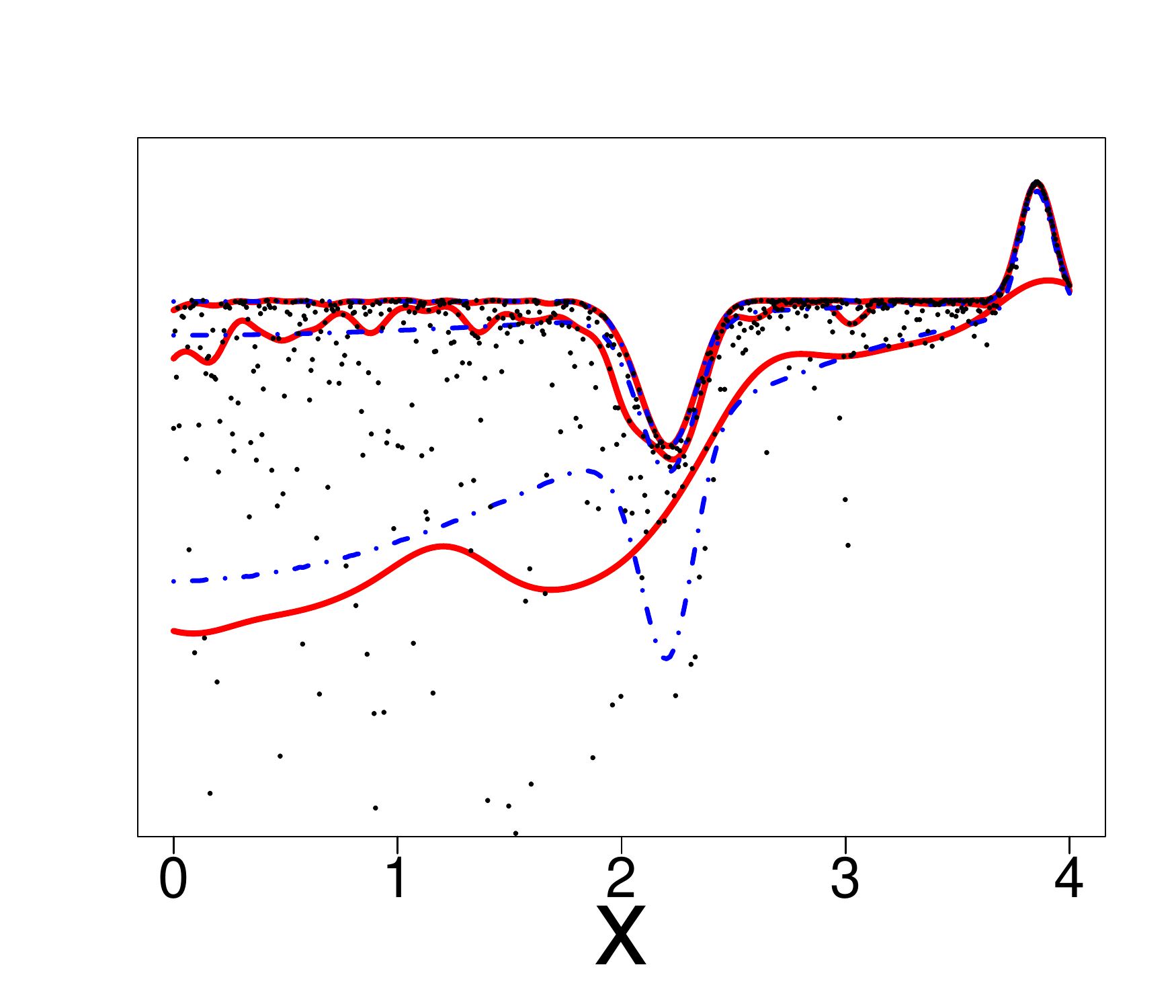}
					\includegraphics[width=.33\textwidth]{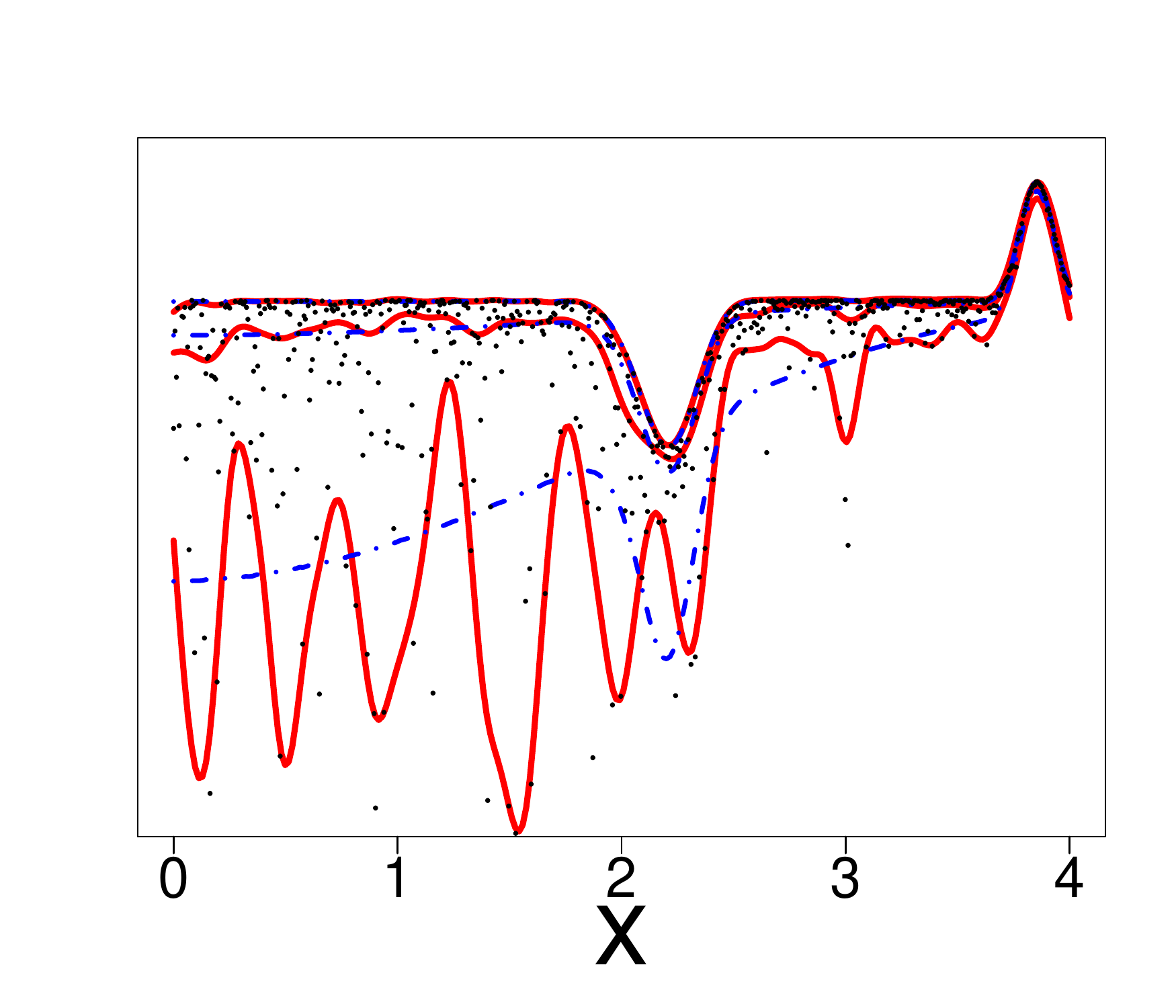}
				\end{tabular}
				\caption{Quantiles estimates using RF (left), RK (middle), VB (right) for a $640$-point sample of the toy problem $3$.
					Dots: observations; plain red lines: metamodels for the $0.1,0.5,0.9$ quantile estimates; dotted blue lines: actual quantiles.}\label{Hete}
			\end{figure*}

			An ideal method would be almost interpolant for a very large pdf but only loosely fit the data when the pdf is small. Indeed, if the pdf is very large then the output is almost deterministic, thus the metamodel should be as closer as possible to the data. In the small pdf case, a point does not provide a lot of information. Information should be extracted from a group of points, that means the metamodel must not interpolate the data. However, most of the methods presented here rely on a single hyperparameter to tune the trade-off between data fitting and generalization: the number of neighbors for KN, the maximum size of the leafs for RF and the penalization factor for NN and RK. As a result, the selected hyperparameters are the ones that are best on average. Theoretically, this is not the case for the Bayesian approaches: QK accounts for it \textit{via} the error variance $\sigma_i^2$ computed by bootstrap, and the weights $w_i$ (Eq. \ref{modif}) allow VB to attribute different ''confidence levels'' to the observations. However in practice, both methods fail to tune the values accurately, as we illustrate below.	Figure \ref{Hete} shows the three quantiles of toy problem 3 and their corresponding RF, RK and VB estimates. For $\tau=0.1$ in particular, the pdf ranges from very small ($x$ close to 0) to very large ($x$ close to 4). Here, RF and RK use a trade-off that globally captures the trend of the quantile, but cannot capture the small hill in the case of large $x$. Inversely, VB perfectly fits this region but dramatic overfitting occurs on the rest of the domain.
			
\begin{figure*}[!ht]
				\begin{tabular}{ccc}
					\includegraphics[width=.33\textwidth]{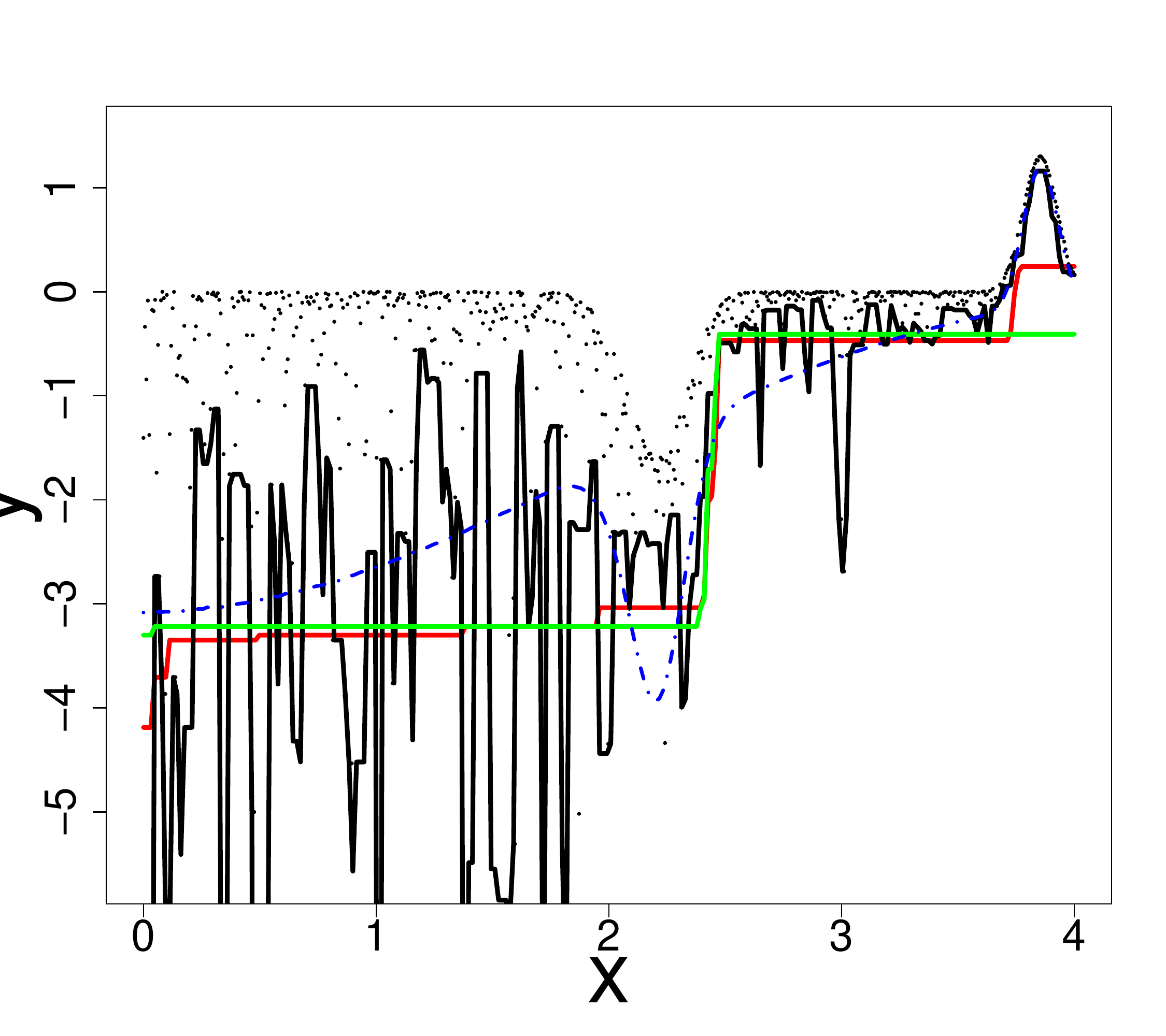}
					\includegraphics[width=.33\textwidth]{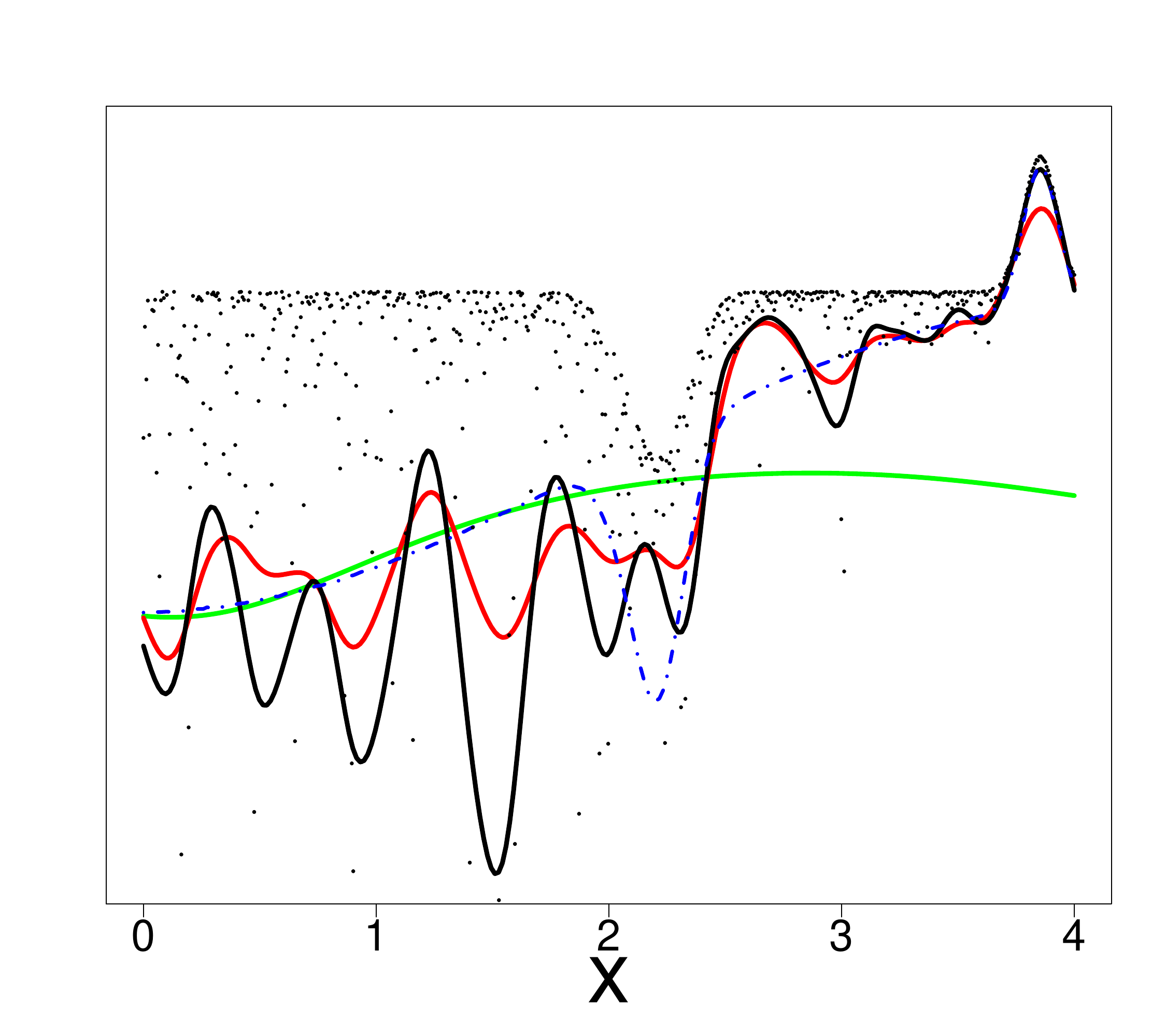}
					\includegraphics[width=.33\textwidth]{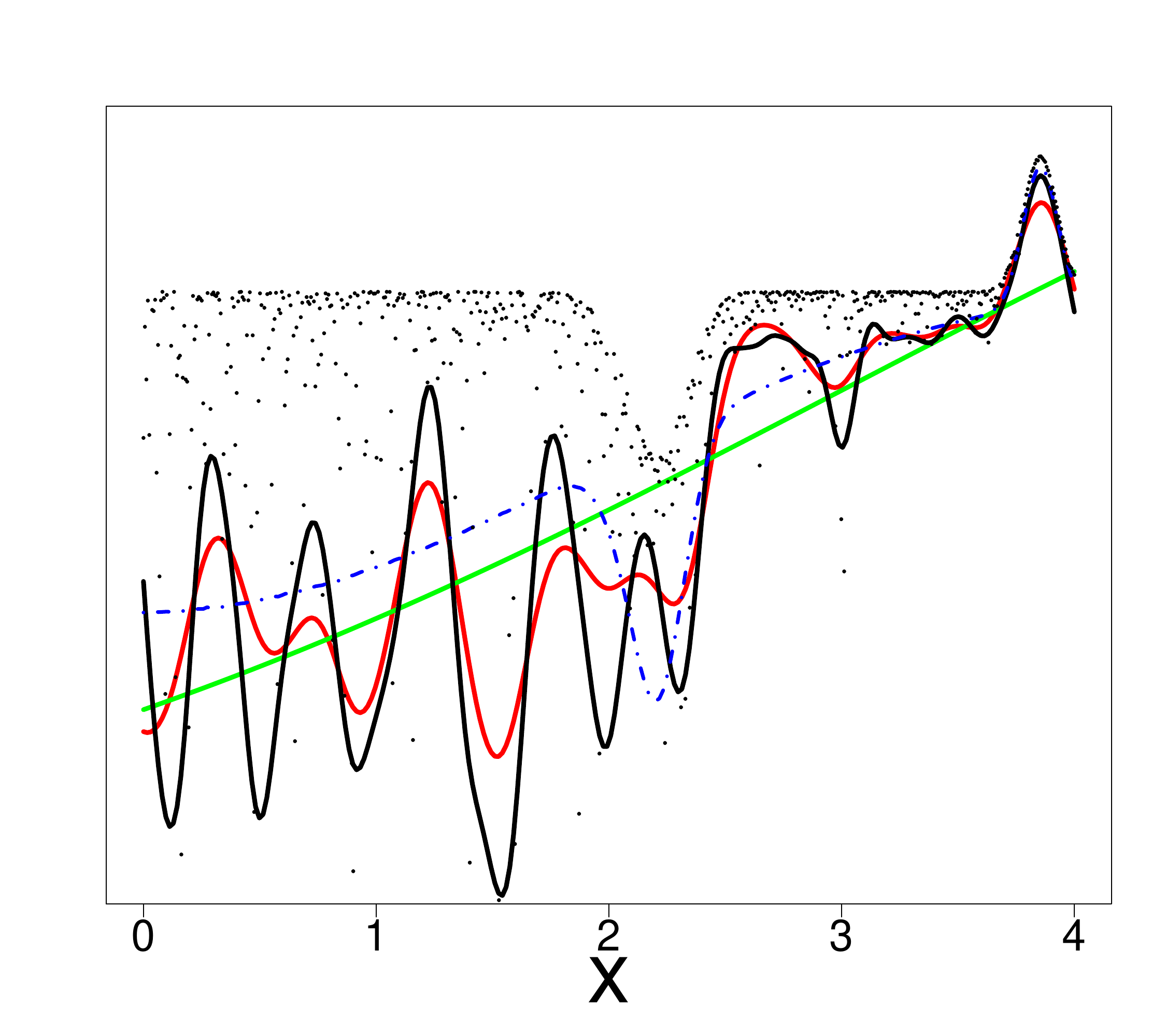}
				\end{tabular}
				\caption{Metamodel responses for toy problem $3$ and $\tau=0.1$ with $640$ training sample (left: RF, center: RK, right: VB) for different values of hyperparameters. The true $0.1$-quantile is presented in dotted blue lines. In green and black two extreme metamodels associated to two extreme hyperparameter values, while in red the oracle metamodels are represented.}\label{fit}
			\end{figure*}

			Finally, Figure \ref{fit} illustrates that this is not an issue of hyperparameter tuning. For each method, we show the oracle estimate, a tuning that tends to underfit and another that tends to overfit. One can see that no tuning is entirely satisfactory, since capturing the region with high pdf leads to overfitting on the rest of the domain and vice-versa.
			
			We believe that further research is necessary to obtain estimators that intrinsically account for strong heteroscedasticity and varying shape. One possible direction is the use of stacking, in the spirit of \cite{sill2009feature}. Under the stacking framework the final estimator could be
			$$\hat{q}(x)=\sum_{i=1}^{N}g_i(x)\hat{q}_{\theta_i}(x),$$
			where $\{\hat{q}_{\theta_i}\}_{1 \leq i \leq N}$ is a set of metamodel and $\{g_i(x)\}_{1 \leq i \leq N}$ is a set of weight functions. Choosing $\{\hat{q}_{\theta_i}\}_{1 \leq i \leq N}$ such that they correspond to different pdf values might provide more flexible estimates.
			
			\subsection{On the non-crossing of the quantile functions}
			While the quantile functions (for different quantile levels) may obviously never cross, unfortunately, their estimators may not always satisfy this property.  This is a well-known issue against which none of the methods presented here is immune.
			
			The neighborhood approaches first estimate the CDF, then extract the quantiles. If the hyperparameters are the same for all quantiles, crossing is impossible. However in our setting, different neighborhood sizes were used for different quantiles.
			
			With functional analysis approaches, crossing may happen even if each quantile is built with the same hyperparameters. In the literature, authors have produced methods to address this issue. It could be reduced by the introduction of additional constraints in the model \cite{takeuchi2006nonparametric} or by the construction of a new model that intrinsically produces non-crossing curves \cite{sangnier2016joint}. However in both cases, the dimension of the optimization problem then increases significantly.
			
			Finally the stochastic process approaches estimate each quantile in independent Gaussian processes, so crossing may occur. While the number of training points is larger than what we consider here ($10^4$ repetitions for each input point), \cite{browne2016stochastic} use GP and takes into account all the quantile orders at the same time and thus ensures non-crossing.
			
			Another approach available for all the methods presented here is related to the rearrangement of curves or isotonic regression \cite{belloni2011conditional,abrevaya2005isotonic}. The idea is to perform many quantile regressions with a large number of different values of $\tau$ or a large set of bootstrapped versions of $\mathcal{D}_n$ and then to rearrange the curves, thereby obtaining the whole distribution and then extracting the quantiles that by definition do not cross.
			
			In theory, adding non-crossing constraints and predicting several quantiles simultaneously could improve the quality of the estimates (in particular as it might add some robustness). However, in practice, it also makes the model more rigid (i.e. a single regularization hyperparameter for all quantiles), and preliminary experiments have shown no gain in accuracy compared to independent predictions, despite a considerably higher computational cost. Hence, multi-quantile predictors were not considered in our study.
			
			\subsection{Assessment of prediction accuracy}
\begin{figure*}[!ht]
		\begin{center}
				\begin{tabular}{cc}
					\includegraphics[width=.45\textwidth]{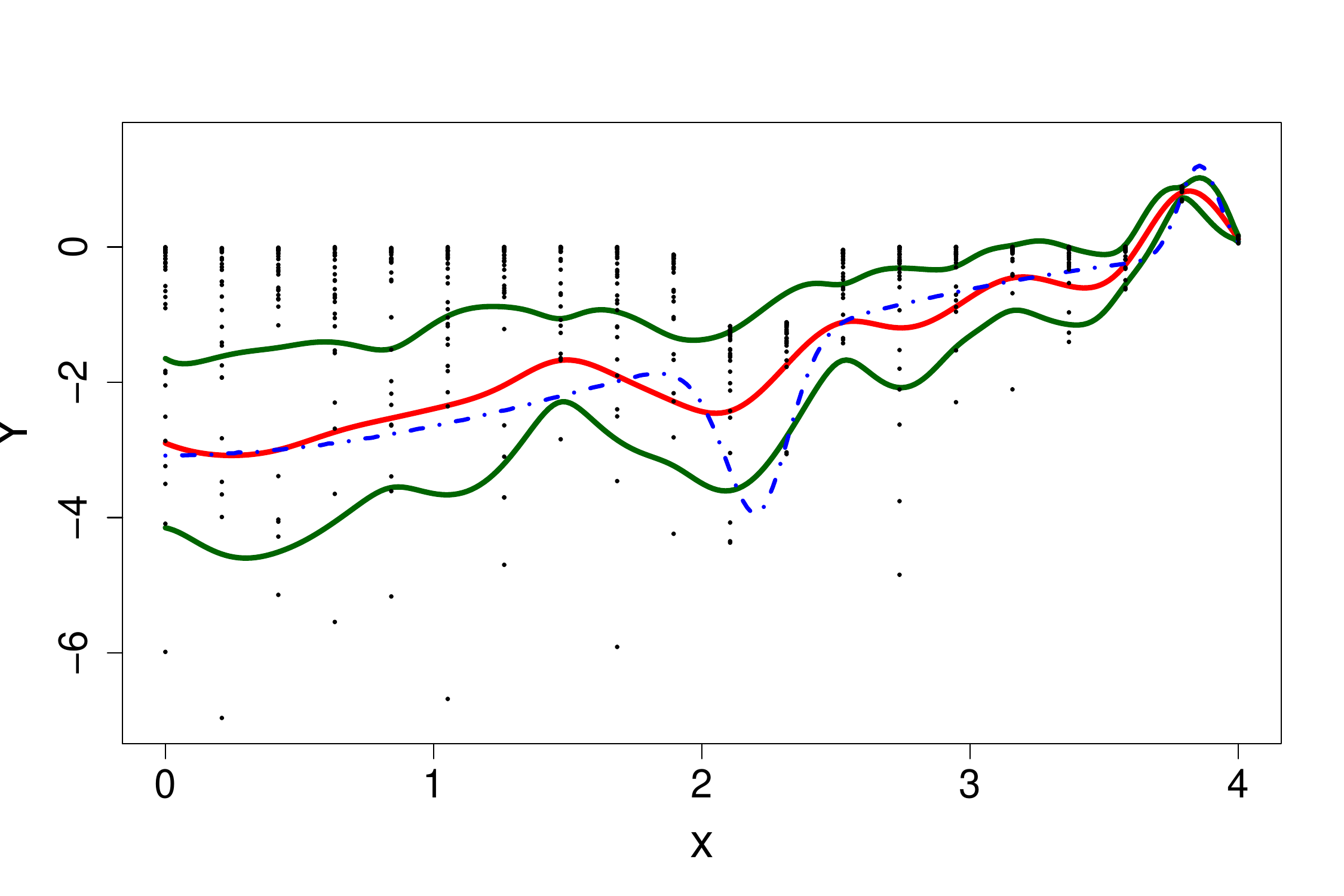}
					\includegraphics[width=.45\textwidth]{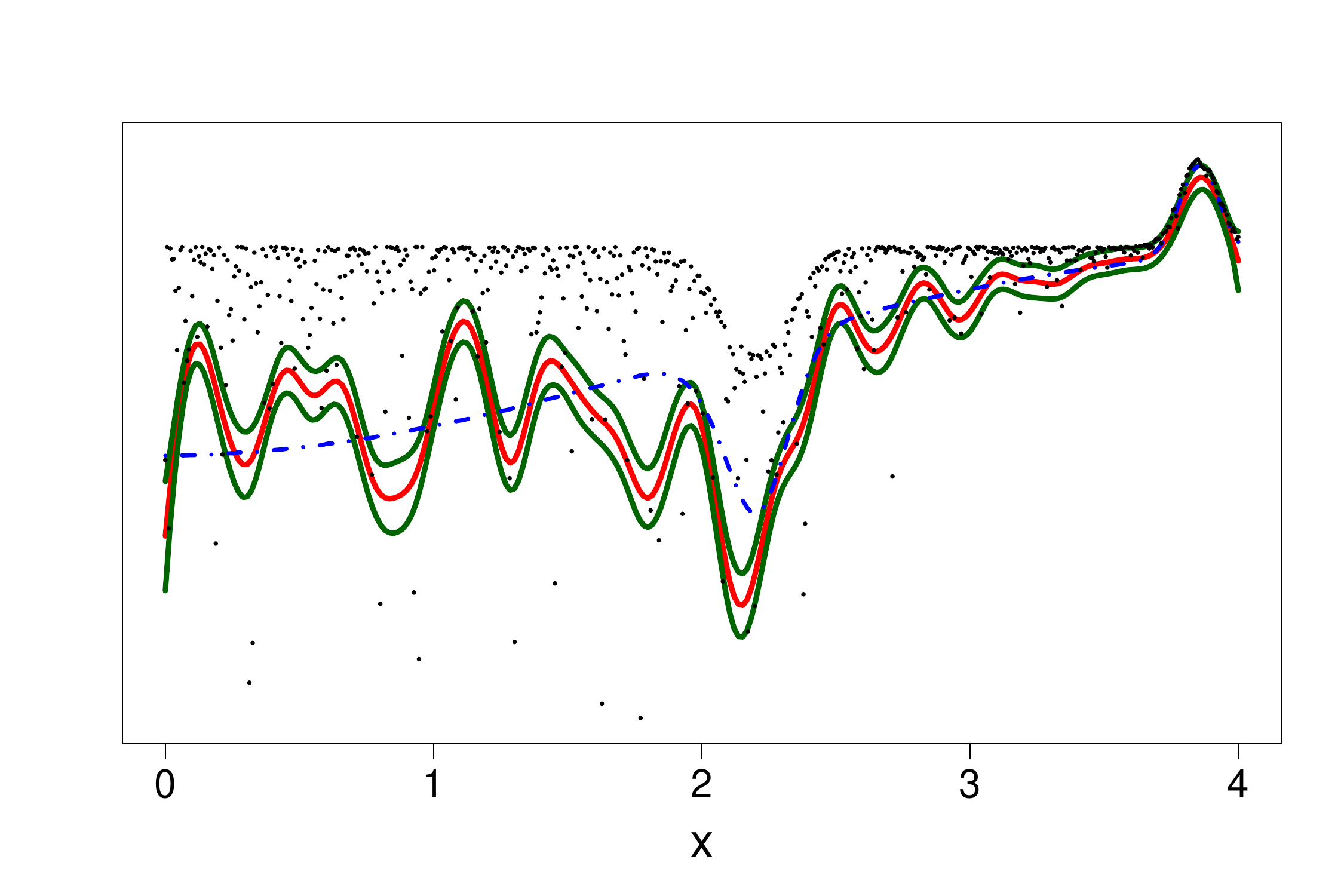}\\
					\includegraphics[width=.45\textwidth]{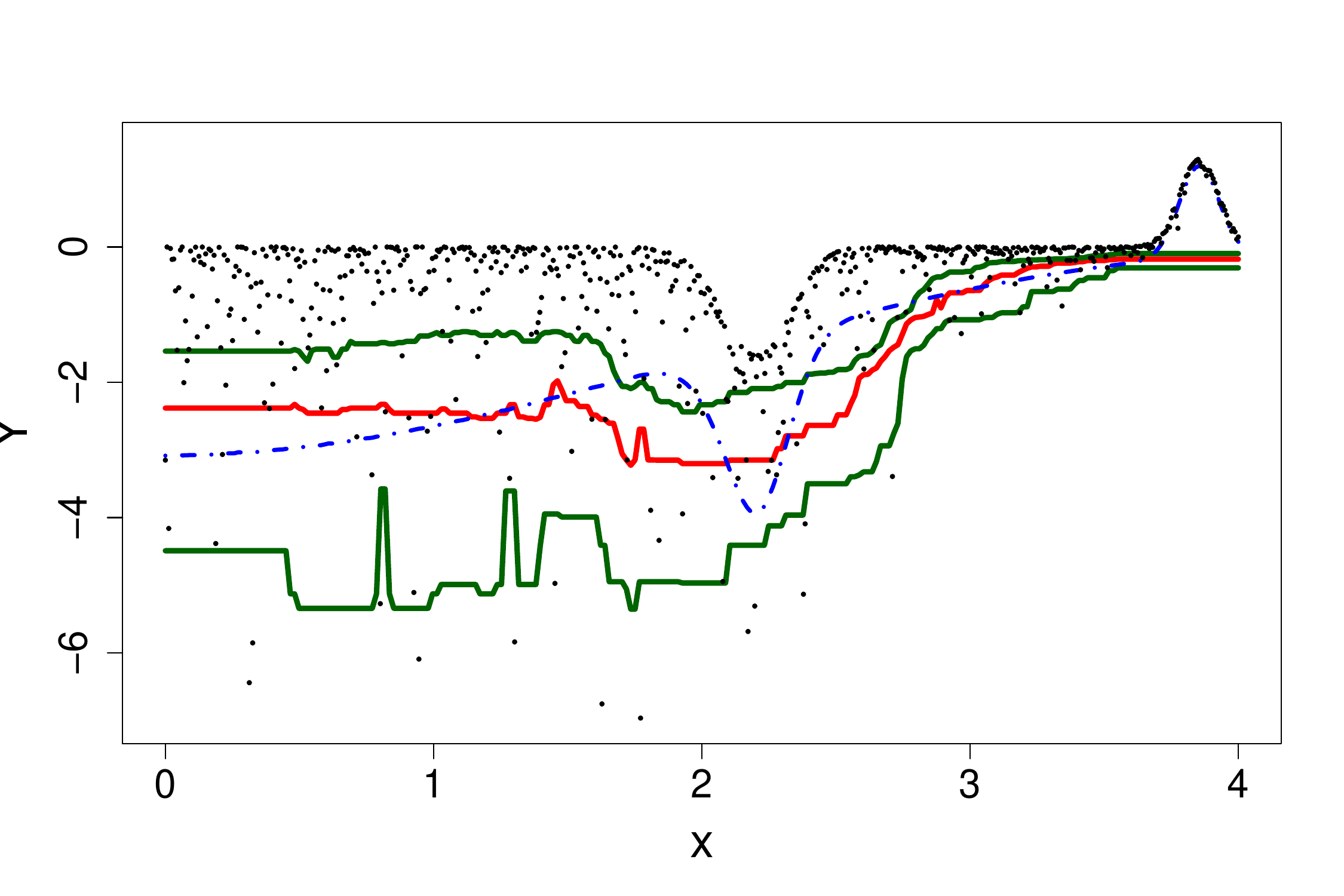}
					\includegraphics[width=.45\textwidth]{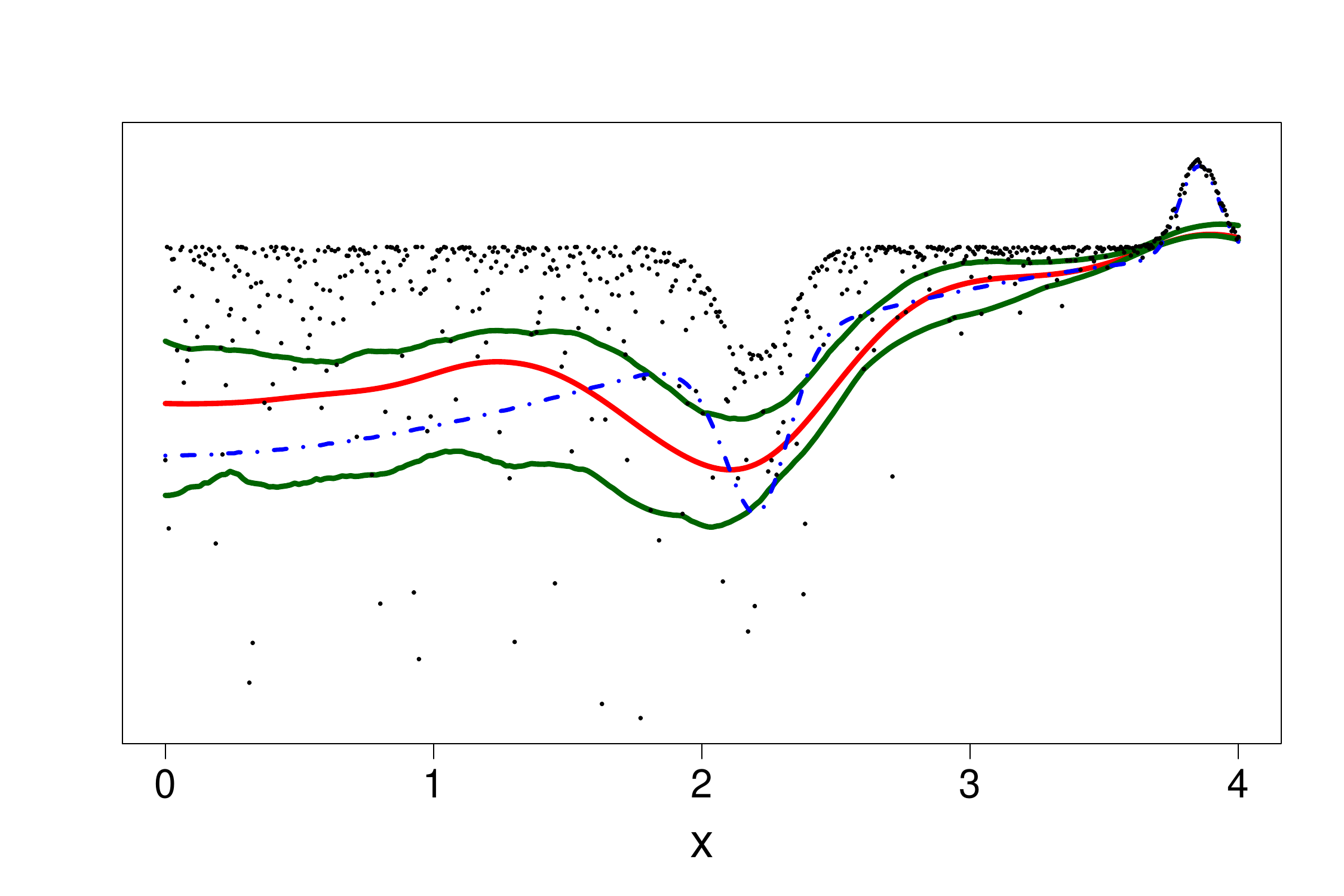}
				\end{tabular}
				\caption{In red quantile metamodel, in blue the true $0.1$-quantile and in green the $0.9$-confidence intervals. Top left QK, top right VB, bottom left KN, bottom right RK.}\label{confidence_intervals}
					\end{center}
			\end{figure*}

			To assess the accuracy of the results and the confidence that we can have in the estimation it could be useful to provide confidence intervals for the predictor. From this point of view the methods are not equal. With Bayesian approaches, theoretical confidence intervals are provided with the models. More precisely as the output model is Gaussian and as it returns the mean and the variance, confidence intervals can be created. For example the $0.9$-confidence interval is provided by
			$$\CI(x)=\hat{q}_\tau(x)\pm1.96\sqrt{\mathds{V}_q(x)}.$$
			However as presented on Figure \ref{confidence_intervals}, while the confidence intervals obtained from QK seem useful, the VB model is clearly overconfident.

			The statistical order methods consider that inside each neighborhood the samples are i.i.d. Based on that, it is possible to use Wilks' formula (see \cite{reiss1976wilks} for instance) or deviation inequality as presented in \cite{torossian2019x} to extract confidence intervals. For example, using Chernoff's inequality, for any $\eta>0$, the confidence interval of order $1-\eta$ is as $\CI(x)=[L_K(x),U_K(x)]$ with 
			
		$$U_K(x)=\min\big\{q,~~\hat{F}^K(q|X=x)\geq\tau~\text{and}~n \KL(\hat{F}^K(q|X=x),\tau)\geq \log(2/\eta)\big\},$$ and
		$$L_K(x)=\max\big\{q,~~\hat{F}^K(q|X=x)\leq\tau~\text{and}~n \KL(\hat{F}^K(q|X=x),\tau)\geq \log(2/\eta)\big\}.$$
		Using this method enables the confidence intervals to be data dependent. For example in Figure \ref{confidence_intervals} the confidence intervals are not symmetric.
		Note that while the confidence intervals obtained with this technique come with theoretical guarantees, they are very conservative and they depend a lot on the size of the training set.
			
			Finally for all methods it is possible to use a bootstrap technique to create different regression models and then to aggregate them in order to create empirical confidence intervals. For example Figure \ref{confidence_intervals} shows confidence intervals using a bootstrap technique for RK. 
	  The main drawback of such method is its computational cost.

	To the best of our knowledge there are no other methods that sensibly  improve this results. Thus, there is still a room for improvement concerning quantile regression model assessment. 
	
			\section{Summary and perspectives}
			\subsection{General recommendations}
			In this presentation we have introduced six metamodels for quantile regression. In the first part of the paper we have provided a full description of the six metamodels, first focusing on their theoretical basis, then discussing their implementation procedure. This part of the paper have enabled us to highlight the similarities and differences of the methods so that providing critical perspectives on the state of the art. 
			The second part of the paper focused on performance comparison according to the dimension of the problem, the size of the learning set, the signal-to-noise ratio and the value of the pdf at the targeted quantile. We have compared the presented methods on $4$ toy problems in dimension 1, 2, 9 and on an agronomic model in dimension 9. 
			
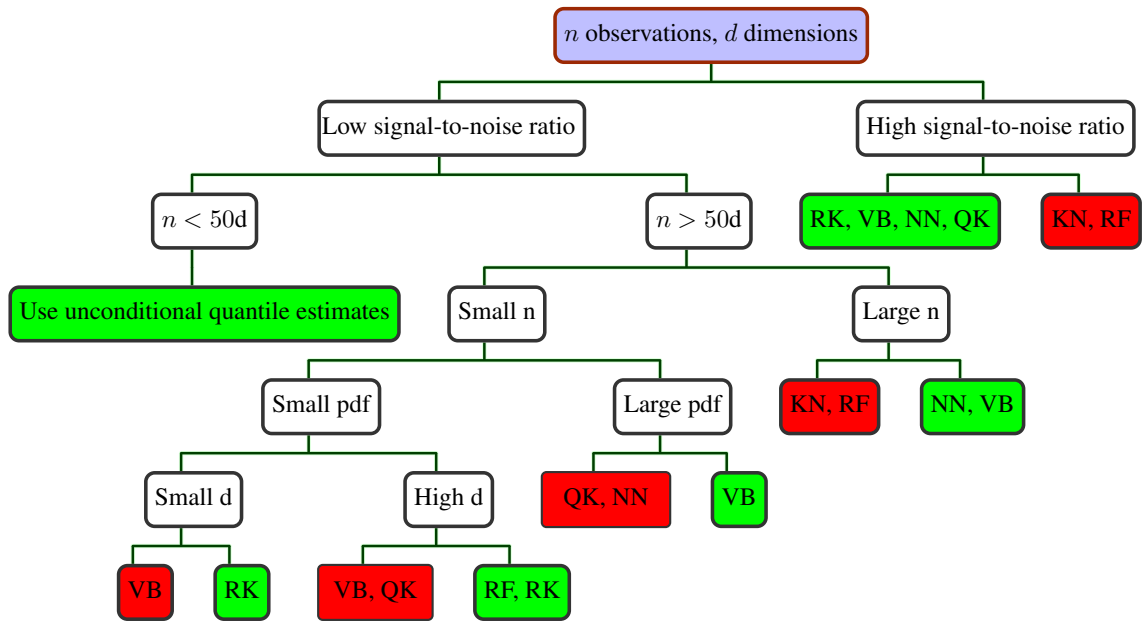
\begin{figure*}[!ht]
			\scalebox{0.95}{
				\begin{tikzpicture}
				\genealogytree
				[
				template=formal graph,
				box={code={
						\gtrifroot
						{\tcbset{colback=blue!25}}{
							\gtrifleafparent
							{\tcbset{colback=blue!50}}{
								\gtrifleafchild
								{\tcbset{colback=green!50}}{}
							}
						}
					}}
					]{
						child{
							g{\text{ $n$ observations, $d$ dimensions }}
						
						child{g[box={colframe=black!80!white,colback=white!5}]{\text{Low signal-to-noise ratio }}	child{g[box={colframe=black!80!white,colback=white!5}]{\text{ $n$ $<50$d }}c[female,box={colframe=black!80!white,colback=green}]{\text{ Use unconditional quantile estimates }}
							}
							child{g[box={colframe=black!80!white,colback=white!5}]{\text{ $n$ $>50$d }}
								
								child{g[box={colframe=black!80!white,colback=white!5}]{\text{ Small n }}child{g[box={colframe=black!80!white,colback=white!5}]{\text{ Small pdf }}
								child{g[box={colframe=black!80!white,colback=white!5}]{\text{ Small d }}c[female,box={colframe=black!80!white,colframe=black!80!white,colback=red}]{\text{ VB }}c[box={colframe=black!80!white,colback=green}]{\text{ RK }}
											
										}child{g[box={colframe=black!80!white,colback=white!5}]{\text{ High d }}c[female,box={colframe=black!80!white,colback=red,size=small}]{\text{VB, QK}}c[box={colframe=black!80!white,colback=green}]{\text{ RF, RK }}
											
										}

									}

									child{g[box={colframe=black!80!white,colback=white!5}]{\text{ Large pdf }}

										c[female,box={colframe=black!80!white,colback=red,size=small}]{\text{ QK, NN }}c[box={colframe=black!80!white,colback=green}]{\text{ VB }}

									}
									
								}
								child{g[box={colframe=black!80!white,colback=white!5}]{\text{ Large n }}c[female,box={colframe=black!80!white,colback=red}]{\text{ KN, RF }}c[box={colframe=black!80!white,colback=green}]{\text{ NN, VB }}
									
								}
							
							}
							
						}
						
						child{g[box={colframe=black!80!white,colback=white!5}]{\text{ High signal-to-noise ratio }}c[female,box={colframe=black!80!white,colback=green}]{\text{ RK, VB, NN, QK }}
							c[female,box={colframe=black!80!white,colback=red}]{\text{ KN, RF }}
							}
						
					}
					}
					\end{tikzpicture}
				}
				\caption{Method recommendation depending on the problem at hand (green: recommended methods, red: methods to avoid). KN: nearest-neighbors, RF: random forests, NN: neural networks, RK: RKHS regression, QK: quantile kriging, VB: variational Bayesian, unconditional quantile estimates: KN with $K=n$.}\label{fig:tree}
		\end{figure*}

				Figure \ref{fig:tree} summarizes our findings. In a nutshell, when the signal-to-noise ratio is high RK, VB, QK and NN shows good results in our experimental setting but as soon as the signal-to-noise ratio decreases, quantile regression requires larger budgets and comparing the methods seems to be more complicated. 
				Indeed, while the rule-of-thumb for computer experiments is a budget (i.e. number of experiments) $10$ times the dimension (see \cite{loeppky2009choosing} for instance), as we work on problems with low signal-to-noise ratio, we found that no method was able to provide a relevant quantile estimate with a number of observations less than 50 times the dimension. For larger budgets, no method works uniformly better than any other. NN and VB are best when the budget is large. When the budget is smaller, RF, RK, KN are best when the pdf is small in the neighborhood of the quantile, in other words, when little information is available. However, VB outperforms all the other methods when  more information is available, that is, when the pdf is large in the neighborhood of the quantile.

        \subsection{Possible ways of improvement}
		In our benchmark, we generally followed the approaches as presented by their authors. However, most of them could be improved. The optimization scheme of NN is the computational bottleneck of the method, which makes it the most expensive method in our benchmark system. One possible improvement would be using the BFGS algorithm (see \cite{lewis2009nonsmooth} for details about the BFGS algorithm applied to non-smooth functions) or the ADAM algorithm \cite{kingma2014adam} to optimize directly the empirical risk associated to the pinball loss. A faster scheme would allow more restarts, and hence improve robustness.
			
			Another improvement concerns the splitting criterion (\ref{cart}) of RF, which is not designed for the quantile but for the expectation. This could lead to poor estimates for problems where quantiles are weakly correlated with expectations. Defining an appropriate splitting criterion could significantly improve the performance of this method.
			
			In our experiments, QK used a predefined number of sampling points that were heuristically defined as a trade-off between space-filling and pointwise quantile estimation accuracy.  The performance of QK could be significantly improved by optimally tuning the ratio between the number of points and repetitions, in the spirit of \cite{binois2018practical}.
			
			The KN method can naturally be extended to a variant that uses the whole sample instead of the $K$ nearest points. The weights associated to each point of the sample could be based on Gaussian or triangular kernel for example. This idea has been developed in \cite{yu2002kernel} to estimate the conditional expectation but we think that it could be possible to adapt this approach to the estimation of the conditional quantile.
			
			Finally, in practice, finding the best hyperparameters was the most difficult part of the proposed benchmark system. While this aspect is often toned down by authors, we believe hyperparameters tuning is a key practical aspect that remains a challenging problem in quantile regression.
			
			\section*{Acknowledgments}
			This work is part of a Ph.D. of L. Torossian funded by INRA and R\'{e}gion Occitanie. The authors would like to thank Edouard Pauwels for the discussions on the topic of convex and non-convex optimization and Pierre Casadebaig for its suggestions and advices concerning the crop model sunflo.
			\bibliographystyle{plain} 
			\bibliography{biblioarticle}
		\end{document}